\documentclass[11pt,a4paper]{article}
\usepackage[pdftex]{graphicx}
\usepackage[cmex10]{amsmath}
\usepackage{amsfonts}
\usepackage{amssymb}
\usepackage{amsthm}
\usepackage{siunitx}

\usepackage{algorithm}
\usepackage{algpseudocode}

\usepackage{xcolor}

\usepackage[hidelinks=true]{hyperref}

\usepackage{xfrac}
\usepackage{enumerate}

\usepackage{subfigure}
\usepackage[font=footnotesize ]{caption}

\usepackage{multirow}
\usepackage{rotating}
\usepackage{makecell}
\usepackage{booktabs}
\usepackage{tabularx,ragged2e}

\usepackage[comma,authoryear]{natbib}

\usepackage{anysize}
\marginsize{2.5cm}{2.5cm}{2.5cm}{2.5cm}

\usepackage{setspace}
\usepackage{parskip}
\usepackage[pagewise]{lineno}

\usepackage{lipsum}

\usepackage{authblk}
\title{\textbf{
Assessing Ranking and Effectiveness of Evolutionary Algorithm Hyperparameters Using Global Sensitivity Analysis Methodologies
}}
\author[1]{Varun~Ojha\thanks{Cite as \\Ojha V, Timmis J, Nicosia G (2022) \textit{Swarm and Evolutionary Computation}, Elsevier.
}}
\author[2]{Jon~Timmis}
\author[3,4]{Giuseppe~Nicosia}

\affil[1]{University of Reading, Reading, United Kingdom}
\affil[2]{The University of Sunderland, Sunderland, United Kingdom}
\affil[3]{University of Catania, Catania, Italy}
\affil[4]{University of Cambridge, Cambridge, United Kingdom}

\date{}
\begin{document}
\onehalfspacing
\maketitle

\definecolor{correct}{rgb}{0.82, 0.1, 0.26} 
\definecolor{insert}{rgb}{0.06, 0.2, 0.65} 
\definecolor{revise}{rgb}{0.0, 0.26, 0.15} 

\definecolor{correct}{rgb}{0.0, 0.0, 0.0} %
\definecolor{revise}{rgb}{0.0, 0.0, 0.0} %
\definecolor{insert}{rgb}{0.0, 0.0, 0.0} %
\definecolor{blue}{rgb}{0.0, 0.0, 0.0} %

\begin{abstract}
    We present a comprehensive global sensitivity analysis of {\color{revise}two single-objective and two multi-objective state-of-the-art global optimization evolutionary algorithms as an \textit{algorithm configuration problem}. That is, we investigate the \textit{quality of influence} hyperparameters have on the performance of algorithms in terms of their \textit{direct effect} and \textit{interaction effect} with other hyperparameters. Using three sensitivity analysis methods, Morris LHS, Morris, and Sobol, to systematically analyze tunable hyperparameters of covariance matrix adaptation evolutionary strategy, differential evolution, non-dominated sorting genetic algorithm III, and multi-objective evolutionary algorithm based on decomposition, the framework reveals the behaviors of hyperparameters to sampling methods and performance metrics. That is, it answers questions like what hyperparameters influence patterns, how they interact, how much they interact, and how much their direct influence is. Consequently, the ranking of hyperparameters suggests their order of tuning, and the pattern of influence reveals the \textit{stability of the algorithms}.}

\end{abstract}

\section{Introduction}
\label{sec:intro_saea}
{\color{revise}Optimization is at the core of advancement in machine learning and problem-solving. Effective optimization plays a vital role in solving problems, whether single-objective or multi-objective problems. For example, be it a simple neural network or deep learning, or a simple linear or nonlinear function, optimizing the coefficients (e.g., weights of neural networks
) is the most crucial aspect, which requires effective optimization algorithms. Evolutionary algorithms (EAs) are global optimization algorithms that iteratively guide a population towards a final population,  solving various problems. EAs are widely used because of their agnostic nature to problems being solved~\citep{de2016evolutionary}. However, their effectiveness relies on hyperparameters like population size and genetic operators~\citep{de2007parameter}. Understanding the sensitivity of hyperparameters to an algorithm's performance can be formulated as an \textit{algorithm configuration problem} (ACP)~\citep{lopez2016irace,iommazzo2019algorithmic}, where informing optimal hyperparameter selection is essential for solving various tasks such as  
neural networks~\citep{crossley2013quantifying}, deep learning~\citep{taylor2021sensitivity}, and bio-inspired algorithms~\citep{das2009differential,ojha2014aco}.} {\color{blue} More specifically, ACP can be described as a process or a framework that aims to find a particular configuration of parameters for a target algorithm. And it minimizes a cost metric incurred by the algorithm on a given problem~\citep{eggensperger2019pitfalls}.
}

Since hyperparameters tuning is crucial in achieving high-quality performance in solving optimization problems, methods such as manual tuning, grid search, and Bayesian search optimization are used. \cite{Bergstra} have shown the importance of \textit{random search} instead of a \textit{grid search} in sampling hyperparameter values. In addition, \textit{manual tuning} without proper knowledge of hyperparameters can lead to too many trial-and-errors, and grid search and Bayesian search optimization are computationally expensive approaches that are often infeasible for such population-based optimization algorithms. {\color{revise}Thus, \cite{Bergstra} suggest that tuning some hyperparameters is more necessary than the others.} Hence, our objective in this research is to assess the ranking and effectiveness {\color{correct} of hyperparameters of} four well-known EAs: covariance matrix adaptation evolutionary strategy (CMA-ES)~\citep{CMAES}, differential evolution (DE)~\citep{DEOriginal}, non-dominated sorting genetic algorithm III (NSGA-III)~\citep{NSGA_III}, and multi-objective evolutionary algorithm based on decomposition (MOEA/D)~\citep{zhang2007moea}. 

{\color{blue}We select these algorithms as they are state of the art algorithms in single-objective and multi-objective optimization. They are the highly cited algorithms not only within the scientific community of bio-inspired computation but also in other scientific disciplinary areas such as operations research, applied mathematics, electrical engineering, civil engineering and many other research areas~\citep{yazdani2021survey}. These algorithms are widely used in multiple multidisciplinary/interdisciplinary problems and are widely used to address real-world problems and open problems in a wide variety of research areas. 

Moreover, researchers have massively investigated these algorithms to improve their performance. For example, a number of improvements to DE have been provided, including success-history based adaptive DE versions~\citep{viktorin2019distance,piotrowski2018step},  mutation operator improvement~\citep{cheng2021differential,das2010differential,islam2011adaptive,das2009differential,biswas2009design,das2007automatic} and scaling factor in mutation for accelerating convergence~\citep{das2005two}. Similarly, an improved step size (mutation) strategy for the CMA-ES algorithm is investigated by \cite{voss2010improved}, improved decomposition strategy like normal boundary intersection-style Tchebycheff approach, adaptive replacement strategies to assign a new solution to a sub-problem, and adaptive weight vector adjustment strategy for sub-problems, respectively proposed by~\cite{zhang2010moea,wang2015adaptive}, and~\cite{qi2014moea} for MOEA/D algorithm. Similarly, for NSGA-III performance enhancement, \cite{cui2019improved} designed an operator to balance the convergence and diversity of the population.

Such usefulness makes these algorithms suitable candidates to be the example of algorithms that can be used as test-beds for the sensitivity analysis methodology presented in our research work. Obviously, our methodology applies to all optimization algorithms with parameters  (e.g., evolutionary, randomized, hybrid, constrained~\citep{yuan2022constrained}, dynamic~\citep{yazdani2021survey}, etc.). The algorithms used in the paper are only a good sample of single- and multi-objective optimization algorithms. We think the optimization research community and interdisciplinary research community mentioned above will benefit as many more optimization algorithms, including many-objective optimization algorithms ~\citep{liang2021clustering,han2022surrogate,rivera2022preference}, that can be studied using the methodology proposed in this paper. 
}

In this work, we develop a framework for comprehensive sensitivity analysis of {\color{revise} hyperparameters of these algorithms} using global sensitivity analysis methodologies: elementary effects~\citep{MorrisOriginal} and variance-based sensitivity analysis~\citep{SOBOL}. Using these methodologies, we assess the effectiveness of EA hyperparameters. {\color{revise} Such an analysis investigates a model's parameters (or an algorithm's hyperparameters) influence on its output~\citep{Sensitivity,Simplifying}, leading to the minimization of} the number of critical tunable hyperparameters to improve a model's performance~\citep{SensitivityTuning,SAUse}.

In our {\color{insert}ACP} framework, the performance of single-objective EAs {\color{correct} was} assessed as per the \textit{best solution}, while the performance of multi-objective EAs {\color{correct} was} assessed using three metrics: \textit{generational distance}~\citep{GDFirstAppearance,GDOtherAppearance}, \textit{inverse generational distance}~\citep{NSGA_III}, and \textit{hyper-volume indicator index}~\citep{HVFirstAppearance}. To evaluate EAs, we use state-of-the-art optimization problems belonging to diverse families: for single-objective optimization, we use a set of {\color{revise}33 problems~\citep{Tesbench,liang2013problem,liang2014problem}}, and for multi-objective optimization, we use a set of 10 problems~\citep{NSGA_III}.

Our {\color{insert}ACP} framework assesses each algorithm on three sensitivity analysis methods: Morris Latin Hypercube sampling~\citep{MorrisOriginal}, Morris sampling~\citep{MorrisOriginal}, and Sobol~\citep{SOBOL}. For each sample drawn from a hyperparameter search space, we ran each algorithm on 30 independent runs {\color{insert}(for some, it was 10 times)} and presented results using elementary effects and Sobol indices. These indices {\color{revise} inform about (i) the direct effect and (ii) the interaction effect of a hyperparameter with other hyperparameters}. Moreover, these two effects form a comparative matrix of low effect to high effect, where the diagonal from low direct and low interaction effects to high direct and high interaction effects shows the \textit{order and ranking} of the hyperparameters. We ran algorithms on a sufficiently large sample set. These experiments were computationally expensive as they, in {\color{correct} total, had \num{19 014 600 000} function evaluations}. Computation of these sensitivity analysis indices is expensive, but they are a one-time effort, and once the ranking is determined, {\color{revise}results are informative to researchers for further analysis and solving optimization problems.} {\color{insert}The source code and results are available at~\url{https://github.com/vojha-code/SAofEAs}.}    

{\color{revise}
Our results reveal the pattern and behavior of hyperparameters to different sampling methods and matrices used to evaluate the performance of the algorithm. These patterns show how hyperparameters interact with one another or how the influence of one hyperparameter overwhelms the other. Moreover, results reveal how an algorithm is susceptible to its various hyperparameters and sampling methods, highlighting the stability of an algorithm. Consequently, these experiments rank the hyperparameter importance for an algorithm. For example, mutation type was found to have the strongest influence on the performance of DE, and results suggest the high importance of population size followed by the initial step size, crossover probability, and mode of decomposition, respectively, in CMA-ES, NSGA-III, and MOEA/D.               

Later in Section~\ref{sec:background}, we present related work. Then, Sections~\ref{sec:ea_algos_theory}, \ref{sec:sensitivity_theory}, and~\ref{sec:experimentalSetup} respectively describe algorithms, methodology, and experiments.} The results are discussed in Section~\ref{sec:results}, followed by conclusions in Section~\ref{sec:conclusions}: {\color{insert}supplementary A and B offer statistical tests and clustering analysis.}


\section{Related Work}
\label{sec:background}
Hyperparameter tuning is a crucial subject that has continuously been reported in the literature over the past decades ~\citep{de2007parameter}. This is because an appropriate hyperparameter setting is challenging since EA hyperparameters exhibit linear and nonlinear effects~\citep{lima2004parameter}, meaning that they show various interactions among them~\citep{de2007parameter,de2004choice,hansen2001completely}. Abundant literature is available on EA hyperparameters tuning~\citep{de2004choice,lima2004parameter,AltroArticolo}. The majority of which focus on the \textit{static}  or \textit{dynamic} setting of the hyperparameters~\citep{eiben2007parameter,kramer2010evolutionary,iglesias2007study}. However, a systematic study of the EA hyperparameters influence is rare~\citep{TuningSensitivity}, and it is largely attributed to the computationally expansive nature of EAs and the empirical evaluation requirement for the tuning of their hyperparameters~\citep{maturana2010autonomous}. For example, a package \textit{Irace} experimentally evaluates optimal hyperparameters for an optimization algorithm~\citep{lopez2016irace}. {\color{revise}Therefore,} \cite{de2007parameter} posed questions like (i) what EA hyperparameters are useful for improving performance, and (ii) how do changes in a  hyperparameter affect the performance of an EA? 

Sensitivity analysis answers questions like \textit{how uncertainty in each of the hyperparameters influences the uncertainty in the output of a model}~\citep{saltelli2004sensitivity}. Hence, sensitivity analysis is useful in answering the questions of~\cite{de2007parameter}. However, sensitivity analysis is a computationally expansive method since hyperparameters are sampled from a vast hyperparameter search space. Therefore, the sensitivity analysis of EAs has very high computational (time) as well as memory (space) overhead.  This has resulted in very few reported works available in the literature, despite its advantages in suggesting a ranking of hyperparameter importance.

The dynamic tuning of hyperparameters requires hyperparameters to adapt during an EA run~\citep{lou2021non}, while static tuning informs which hyperparameters to tune before an EA run~\citep{kramer2010evolutionary}. A systematic approach, like sensitivity analysis, is a static hyperparameter tuning approach. \cite{paul2011sensitivity} offered an introductory work on the usage of \textit{local} and \textit{global} sensitivity analysis. However, they used a simple test case, and they mainly performed a sensitivity analysis of EAs from a theoretical perspective. \cite{TuningSensitivity} performed a comprehensive sensitivity analysis of a parallel asynchronous cellular genetic algorithm on a scheduling problem. They comprehensively evaluated EAs population size, mutation probability,  crossover probability, and other cellular genetic algorithm-related hyperparameters using the Fourier amplitude sensitivity test (Fast99)~\citep{saltelli1999quantitative}. \cite{TuningSensitivity} reported a ranking of hyperparameters on scheduling problem instances. On this scheduling problem instance, the crossover probability was ranked first, and in another instance, it was ranked third. 

Our work takes an experimental approach to systematically analyze the importance of hyperparameters of state-of-the-art EAs on a testbench of state-of-the-art problems by applying Morris~\citep{MorrisOriginal} and Sobol~\citep{SOBOL} sensitivity analysis methodologies. Our methodology comprises both single-objective and multi-objective EAs. Our framework offers a ranking of hyperparameters and insights into their effectiveness on EA performance. {\color{insert} Our methodology is an \textit{Algorithm Configuration Problem} (ACP) framework as defined by~\cite{iommazzo2019algorithmic}. This approach is contextually similar to the AutoML approaches~\citep{he2021automl}, where the effort is to find the optimal configuration of algorithms and hyperparameters to solve machine learning tasks through automatic data preparation, feature engineering, hyperparameter optimization, and neural architecture search or even optimization of neural network components such as activation functions~\citep{ojha2014simultaneous}.{\color{blue} Table~\ref{tab:the_synthetic_table} is a summary of hyperparameter methods compared to sensitivity analysis methods.}

In fact, the ACP scope covers a wider range of methodologies and frameworks that seek to automate the design of algorithm configuration, such as AutoMOEA~\citep{bezerra2015comparing}, Auto Weka~\citep{thornton2013auto}, Auto-sklearn~\citep{feurer2020auto}, irace~\citep{lopez2016irace}, and others for machine learning hyperparameter optimization~\citep{feurer2019hyperparameter}. The goal of these methodologies is to perform hyperparameter optimization and automatic design of new algorithms by assessing components and parameters that offer the best performance on a set of problem instances~\citep{iommazzo2019algorithmic,thornton2013auto}. The critical issue in such categorization is whether one would consider, for instance, a new evolutionary operator design in an EA framework as a new algorithms design or hyperparameter optimization? In our work, we consider such a scenario as hyperparameter optimization. {\color{blue} However, we considered the ACP framework for the analysis of the sensitivity and influence of the hyperparameters on the performance of an algorithm rather than the optimization (or tuning) of the hyperparameters. For this, the framework systematically searches hyperparameters and assesses the performance of an algorithm, which is contrary to finding specific optimal values for a hyperparameter as other hyperparameter tuning methods would do. Hence, the goal of our ACP framework is to inform the ranking of the effectiveness of hyperparameters for a set of EAs.}

\begin{table}
    \color{blue}
    \centering
    \caption{\color{blue}Hyperparameter tuning methods and sensitivity analysis (our framework). Hyperparameter tuning methods are search techniques for optimal hyperparameter values whereas our framework finds ordering of their significance and their sensitivity of influence on the algorithm.}
    \begin{tabular}{lllll}
        \toprule
        & Method & Tuning & Type & Use \\
        \midrule
        \multirow{4}*{\rotatebox[origin=r]{90}{Search}} 
        & Manual Tuning  & static & requires intuitive guesses &  trails and errors \\
        & Grid Search & static & systematic search & uninformed search \\
        & Bayesian Search & static & informed search & expansive and specific to instances \\
        & AutoMOEA & dynamic & systematic and informed & expansive and subjective \\
        & AutoML & dynamic & informed search & expansive and specific to problems \\
        \midrule
        & Our Framework & Static & ranking and analysis & expansive but one at a time\\
        \bottomrule
    \end{tabular}
    \label{tab:the_synthetic_table}
\end{table}

}

\section{Evolutionary Algorithms}
\label{sec:ea_algos_theory}
EAs are population-based evolution-inspired algorithms. EAs iteratively find solutions to a problem by applying evolutionary operators to candidate solutions. Selection, recombination, and mutation are among evolutionary operators applied to candidate solutions that generate new solutions in each generation. Such a process guides a sequence of generations from an initial population of candidate solutions to a final population. Four different EAs are investigated in this research: two single-objective and two multi-objective algorithms. Each of these EAs has its own version of evolutionary operators. This Section briefly describes each of these EAs and their performance measure metrics.

\subsection{Single-objective Evolutionary Algorithms}
\label{sec:sing_obje_ea}
A single-objective optimization (SOO) algorithm (single solution-based or population-based) \textit{\textit{minimizes an objective function}} (a cost function or a problem) as 
\begin{equation}
    \begin{array}{ll}
        f: & \mathbb{R}^n \rightarrow \mathbb{R} \\
           & \textbf{x} \mapsto f(\textbf{x}),\\  
    \end{array}
\end{equation}
where $ \textbf{x} \in \mathbb{R}^n$ is a candidate solution (a search point in a solution space $X$), and we want $f(\textbf{x})$ to be as minimum as possible. An SOO algorithm converges to a solution $\textbf{x}^*$ such that 
$ f(\textbf{x}^*) \le f(\textbf{x}),~ \forall \textbf{x} \in X$. The solution $ \textbf{x}^* $, therefore, is a \textit{global minimum} (global optimum). However, if for $f(\textbf{x}^*) \le f(\textbf{x})$ there exists some $\delta > 0 $ such that $|\textbf{x} - \textbf{x}^*| \le \delta$ for any $\textbf{x} \in X $, {\color{correct}then} the solution $\textbf{x}^*$ is a \textit{local minimum} (near-optimum). 

We study two population-based single-objective global optimization algorithms: CMA-ES~\citep{CMAES} and DE~\citep{DEOriginal}. The basic steps and operators of CMA-ES and DE are as follows. 

\subsubsection{Covariance Matrix Adaptation Evolution Strategies (CMA-ES)}
CMA-ES is a population-based \textit{evolutionary strategy} optimization algorithm~\citep{CMAES}. CMA-ES algorithm generates new candidate solutions during its search by sampling solutions from a \textit{multivariate normal distribution}, $\mathcal{N}(\textbf{m}, C)$, uniquely determined by its mean $\textbf{m} \in \mathbb{R}^n$ and its symmetric positive definite covariance matrix $C \in \mathbb{R}^{n \times n}$. {\color{revise} The initial population of $\lambda$ candidate solutions at generation $g = 0$ is sampled as} 
\begin{equation}
   \textbf{x}_k^g \sim \textbf{m}^g + \sigma^g \mathcal{N}(\textbf{0},C^g)\quad \mbox{ for } k = 1, \ldots, \lambda,
\end{equation}
where $\mathcal{N}(\textbf{0},C)$ is a multivariate normal distribution with zero mean and covariance matrix $C^g \in \textbf{I}$, and $\sigma^g \in \mathbb{R}_{>0}$ is an initial step size.

For generation $g=1,2,\ldots$, multivariate normal distribution $\mathcal{N}(\textbf{m},C^{g+1})$ is generated (updated) with mean $\textbf{m} \in \mathbb{R}^n$ and covariance matrix $C \in \mathbb{R}^{n \times n}$ updated with scalar factor $\sigma^g \in \mathbb{R}_{>0}$. 
%
%
Selection and recombination operations in CMA-ES are equivalent to computing moving mean $ m^{g+1} $, a weighted average of selected points $\lambda_{\text{ratio}}$ from generation $g$. Adding a random vector with zero-mean acts as a mutation in CMA-ES during the offspring generation step. The steps size control and covariance matrix adaptation (learning rate $ \alpha_{\mu} $) are additional two necessary steps in a generation of CMA-ES~\citep{CMAES}. 

\subsubsection{Differential Evolution (DE)}
DE is a \textit{gradient-free} EA, originally proposed by ~\cite{DEOriginal}. DE iteratively searches for a solution. For an initial population $ X = [\textbf{x}_1, \textbf{x}_2,\ldots, \textbf{x}_\lambda] $ of size $ \lambda $, DE repeats its steps \textit{selection}, \textit{mutation}, and  \textit{recombination} until an optimum solution vector $ \mathrm{\mathbf{x}}^*$ is obtained, or until a maximum iteration is reached. At each generation $g =1,2,\ldots $, DE randomly selects three distinct candidate solutions $ \mathbf{x}^g_{r1}$, $ \mathbf{x}^g_{r2}$, and $\mathbf{x}^g_{r3}$ from $ X $ such that $ \mathbf{x}^g_{r1} \ne \mathbf{x}^g_{r2} \ne \mathbf{x}^g_{r3} $. The selection of a base vector $ \mathbf{x}^g_{r1} $ plays a crucial in DE.

A mutation operation is performed on a base vector $ \mathbf{x}^g_{r1} $ to generate a donor vector $\mathbf{v}^{g+1}$, which is generated using a mutation method $ \mathbf{b}_{\text{type}} $, a difference vector $(\mathbf{x}^g_{r2} \mathbf{x}^g_{r3})$, and acceleration coefficient $\beta$. A mutation method $ \mathbf{b}_{\text{type}} $ = ``DE/rand/1'' or similar mutation is performed as 
\begin{equation}
    \mathbf{v}^{g+1} = \mathbf{x}^g_{r1} + \beta (\mathbf{x}^g_{r2} \mathbf{x}^g_{r3}).
\end{equation}

A crossover operation using a crossover method \{bin, exp\} is performed to generate a trial vector $ \mathbf{u}^{g+1} $ which takes its elements from a donor vector $\mathbf{v}^{g+1}$ using a crossover probability $ P[\mathrm{X}] $. If the fitness $ f(\mathbf{u}^{g+1})$ is better than the target vector $ f(\mathbf{x}^{g+1}_t) $, then the trial vector $ \mathbf{u}^{g+1} $ replaces the target vector $\mathbf{x}^{g+1}_t$. 

\subsection{Multi-objective Evolutionary Algorithms}
A multi-objective optimization (MOO) algorithm minimizes two or more objective functions \textit{simultaneously} as
\begin{equation}
F(\textbf{x}) \equiv  (f_1(\textbf{x}), \ldots, f_k(\textbf{x})), \mbox{ i.e., } F: \mathbb{R}^n \rightarrow \mathbb{R}^k \mbox{ for } k \ge 2    
\end{equation}
such that no one objective of the problem can be improved without a simultaneous detriment to at least one of the other objectives. Each $ f_l(\textbf{x}), l=1,2,\ldots,k$ is a scalar objective, and MOO optimizes the objective vector  $F(\textbf{x})$ where $\textbf{x} \in \mathbb{R}^n$ is its feasible solution. More specifically, a MOO algorithm produces a set of non-dominated solutions $\{\textbf{x}_1, \textbf{x}_2, \ldots, \textbf{x}_{\lambda'}\}$, also known as the Pareto-optimal solutions set~\citep{NSGA_II}.

A solution $ \textbf{x}_i $ dominates other solution $ \textbf{x}_j $ if for $ j = 1, 2, \ldots, \lambda, i \ne j $, and for all objectives $ l = 1, 2, \ldots, k$, $f_l(\textbf{x}_i) \preccurlyeq f_l(\textbf{x}_j)$ holds, where $\preccurlyeq$ should be read as ``better off.'' On the contrary, a solution $ \textbf{x}_i $ is non-dominated if, for at least one objective $l$, $f_l(\textbf{x}_i) \preccurlyeq f_l(\textbf{x}_j)$ does not hold. For each $\textbf{x}_i$, a set of such non-dominated solutions are called a Pareto-optimal set of solutions.    

In this paper, we study the population-based multi-objective global optimization algorithms NSGA-III~\citep{NSGA_III} and MOEA/D~\citep{zhang2007moea} and investigate their algorithmic hyperparameter setting in obtaining a better Pareto-optimal set of solutions.

\subsubsection{Non-Dominated Sorting Genetic Algorithm--III (NSGA-III)}
NSGA-III is a population-based MOO algorithm~\citep{NSGA_III}.
NSGA-III uses fast non-dominated sorting and niching operations to guide an initial population $X $ of size  $ \lambda $ candidate solutions through a predefined number of generations to a final population while simultaneously optimizing trade-offs of multiple objectives. In each step of NSGA-III, crossover, mutation, and non-dominated sorting is performed.

The fast non-dominated sorting sorts the $ \lambda $ candidate solutions into several sets (called Fronts) of non-dominated solutions: $F_1, F_2, \ldots, F_s$ such that the Front $ F_1 $ contains all the non-dominated candidate solutions of population $ \mathrm{X} $. That is, no one solution in $F_1$ is dominated by any other solutions. From all the remaining solutions (i.e., except the ones already in $F_1$), a new Front $F_2$ that contains all the next non-dominated solutions of $ \mathrm{X} $ is determined. Similarly, Front $F_3$ and other Fronts are subsequently obtained using non-dominated sorting. Thus, it is possible to assign a rank to the candidate solutions such that those on the Front $F_1$ have rank 1, solutions in Front $F_2$ have rank 2, and so on.

NSGA-III performs \textit{niching} as its selection operation on non-dominated sorting solutions. Niching takes advantage of a predefined set of reference points placed on a normalized hyperplane of a $ k $-dimensional objective-space~\citep{das1998normal}, where each individual $ \textbf{x} \in X $ in the population is associated with reference points~\citep{NSGA_III}. The total number of reference points depends on the predefined number of divisions associated with each objective axis. 
NSGA-III  repeats its operations selection, crossover, mutation, and recombination until a maximum iteration or a termination condition is reached. The performance of NSGA-III is measured in terms of the quality of solutions it produces in its iteration and in the final population.

\subsubsection{Multi-objective Evolutionary Algorithm based on Decomposition (MOEA/D)}
MOEA/D solves a MOO problem by decomposing the MOO problem into many single (scalar) objective sub-problems~\citep{zhang2007moea}. Tchebycheff approach~\citep{miettinen2012nonlinear} or normal boundary interaction approach ~\citep{das1998normal} are typically used approaches for decomposing a MOO problem into (say) $ N $ scalar sub-problems. A uniform spread of $N$ weight vectors $\{\textbf{w}_1,\ldots, \textbf{w}_N \}$ and reference point $\textbf{z}^* = (z^1_j, \ldots, z^k_j) = \min\{f_i(\textbf{x})| \textbf{x} \in X\}, \mbox{ for } i = 1,\ldots, k $ is used for computing $j = 1,2, \ldots, N$ scalar objectives  $y^{te} (\textbf{x} | \textbf{w}_j)$.  

The scalar objective in Tchebycheff decomposition method is $y^{te} (\textbf{x} | \textbf{w}_j) = \max_{1\le i \le k} \{\textbf{w}^i_j | f_i(\textbf{x}) - \textbf{z}^* \} ) $, where the weight vector $\textbf{w}_j = (w^1_j, \ldots, w^k_j) $. The optimal solution of  $y^{te} (\textbf{x} | \textbf{w}_i)$ for weight vector $ \textbf{w}_i$ should be close to a solution  $y^{te} (\textbf{x} | \textbf{w}_j)$ for weight vector $ \textbf{w}_j$. Hence, in MOEA/D, a neighborhood of  weight vector $ \textbf{w}_i$ is defined with many closest points in $\{\textbf{w}_1,\ldots, \textbf{w}_N \}$. The neighborhood may play a vital role in MOEA/D. 

Moreover, each objective is optimized as a single (scalar) objective problem. That is, $ i $th objective is optimized such that it minimizes its distance from a reference point on a $k-$objective space. Thus, all decomposed sub-problems move towards the reference point $\textbf{z}^*$. MOEA/D maintains $ T $ closest solution vectors (Neighbor) for each candidate solution in successive steps. In each iteration, MOEA/D generates a new solution by selecting two solution vectors using genetic operators and evaluating them in order to update their neighborhood and the best solution $\textbf{x}^*$. The details of the MOEA/D algorithm are available in~\citep{zhang2007moea}.

\subsection{Performance Metrics}
\label{sec:perfromance_metric}
\subsubsection{Single Objective Metrics}
A population-based EA applied to solve a single-objective problem offers the best solution in its final population. The best solution, $ \textbf{x}^* $ is the one that has the lowest $f(\textbf{x})$ value among all solutions of all generations of a single-objective EA. Hence, the \textit{Best Solution} obtained in fewer generations in a lesser wall clock time measures the quality of a single-objective EA.

\subsubsection{Multi-Objective Metrics}
\label{sec:moo_metric}
\begin{figure}
	\centering
	\includegraphics[width=0.98\textwidth]{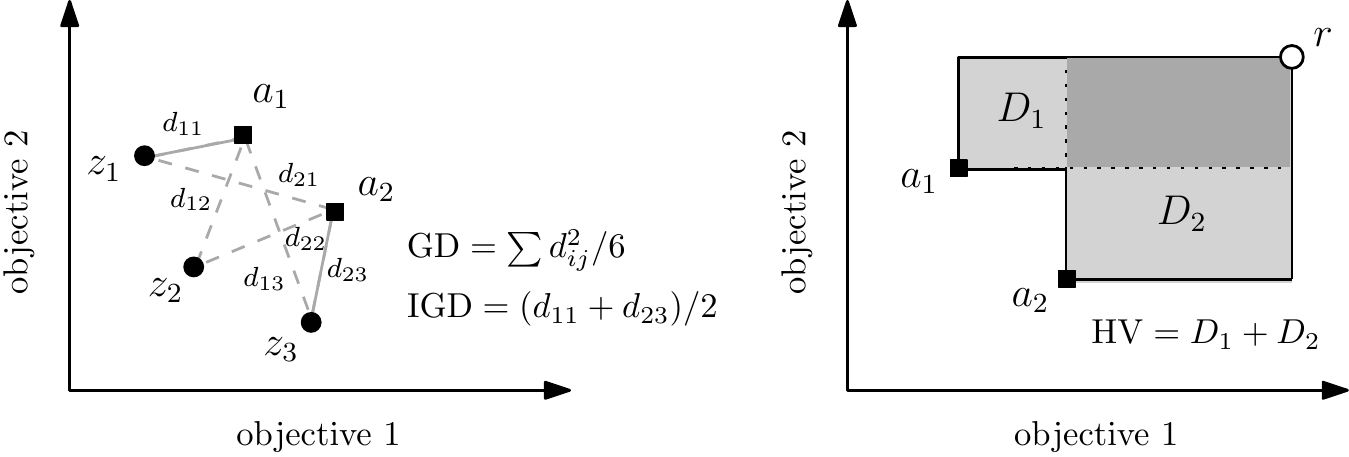}
	\caption{\color{insert} Example of a 2D objective space and computation of GD, IGD, and HV metrics. The current Pareto front is $\textbf{A} = \{a_1, a_2\}$ and true Pareto front is $\textbf{Z} = \{z_1, z_2, z_2\}$, and optimum of two-objective is a reference point $r$. The distance between two points is $d_{ij}$, and the area framed by a point with the reference point, $r$, is the area $D_i$. 
	\label{fig:moo_metric}}
\end{figure}

Multi-objective EAs applied to a MOO problem typically offer a set of solutions that satisfy trade-offs between the objectives. This set of solutions is non-dominated solutions which are also known as a Pareto-front. A multi-objective EA, therefore, guides a population of candidate solutions from \textit{current Pareto-front} $\textbf{A}$ toward a \textit{true Pareto-front} $\textbf{Z}$.
In such a setting, three indicators are used to measure and compare the performance of EAs on MOO problems: generational distance, inverse generational distance, and hyper-volume indicator (Fig.~\ref{fig:moo_metric}):

\paragraph{Generational Distance (GD).}
Generational distance GD$_i$ at an iteration, $ i $, measures the generational distance between \textit{current Pareto-front} and \textit{true Pareto-front} of a multi-objective problem~\citep{GDFirstAppearance,GDOtherAppearance}.  Generational distance GD$_i $ is a measure of error between \textit{current Pareto-front} and \textit{true Pareto-front} as
\begin{equation}
\text{GD}_i  \triangleq  \frac{\sqrt{\sum_{i=1} d_i^2}}{n},
\end{equation}
where $d_i^2$ is the distance of the $i$th solution in \textit{current Pareto-front} $\textbf{A}$ from the \textit{true Pareto-front} $\textbf{Z}$~\citep{GDOtherAppearance}, and GD is typically the average distance of such $ n $ solutions (Fig.~\ref{fig:moo_metric}). Hence, GD is a minimization metric where a low value {\color{correct} indicates} a better solution. 

\paragraph{Inverse Generational Distance (IGD).}
Inverse generational point provides combined information on the solutions' diversity and convergence quality. It makes use of a set of target reference points in $ k $-dimensional objective space. Like GD, IGD compares solutions in the \textit{current Pareto-front} $\textbf{A}$ with  \textit{true Pareto-front} $\textbf{Z}$. However, IGD uses a single reference and computes the average Euclidean distance between all solutions that are nearest to the target reference points~\citep{NSGA_III} as
\begin{equation}
        IGD (\textbf{A}, \textbf{Z}) \triangleq \frac{1}{|\textbf{Z}|} \sum\limits_{i = 1}^{|\textbf{Z}|} \min\limits_{j=1}^{|A|} d(\textbf{z}_i,\textbf{a}_j),
\end{equation}
where $d(\textbf{z}_i,\textbf{a}_j) = ||\textbf{z}_i,\textbf{a}_j||_2$. Similar to GD, IGD is a measure of error between {\color{correct}the} \textit{current Pareto-front} and \textit{true Pareto-front}. Hence, lower values of IGD indicate a better solution. 

\paragraph{Hypervolume Indicator (HV).}
Hyper-volume indicator, HV measures the dominance of Pareto-front solutions on a geometric space (e.g., area for a 2D objective space) framed by the $ k $-dimensional objective-space with respect to a positive semi-axle $r$ (see Fig.~\ref{fig:moo_metric}). Hence, HV measures the quality Pareto-optimal solutions set~\citep{fonseca2006improved}, and it is an indicator of the quality of the solutions obtained by two algorithms with respect to the same reference frame. The goal is to maximize the hyper-volume indicator index HV. A greater value indicates that the algorithm's overall performance is better with respect to another algorithm associated with a smaller hyper-volume value. Moreover, the greatest contributing point in a hyper-volume indicator analysis is the point covering the largest area, which can be considered the best solution~\citep{zitzler2003performance}.

\section{Global Sensitivity Analysis}
\label{sec:sensitivity_theory}
The goal of the \textit{sensitivity analysis} is to study how the uncertainty of a model's output depends on the uncertainty of its inputs~\citep{saltelli2002sensitivity,saltelli2008global}. 
The \textit{elementary effects} analysis, known as the ``Morris method''~\citep{MorrisOriginal}, and \textit{variance-based sensitivity analysis}, known as the ``Sobol method''~\citep{SOBOL}, are used in this research for the global sensitivity analysis of the hyperparameters of four EAs. {\color{insert} This framework of combining sensitivity analysis and EAs is an \textit{algorithms configuration problem} that aims to inform algorithm performance to variations in hyperparameter on problem instances~\citep{iommazzo2019algorithmic}. } 

\subsection{Elementary Effects}
The \textit{elementary effects} (EE) technique, known as the ``Morris method'' as it was originally introduced by~\cite{MorrisOriginal}, is an effective way to analyze the effects (sensitivity) of input variables on the outputs of a model or a system. In our case, the Morris method assesses the EE  of the algorithmic hyperparameters on the performances of an EA. This is useful in analyzing the sensitivity of EA hyperparameters as the Morris method determines whether the effects of a hyperparameter on a model’s outputs (EA performances on functions) are (a) insignificant and negligible, (b) linearly correlated, or (c) non-linearly correlated or involved in an interaction with other hyperparameters~\citep{saltelli2008global}. 

We briefly introduce the computation of EE as follows. Let us have $Y = f(\textbf{X})$, or simply $Y(\textbf{X})$ be the output of a model $f(\cdot)$ (an algorithm) that takes $k$ hyperparameters $\textbf{X} = \{X_1, X_2, \ldots, X_{k}\}$ from a hyperparameter space $\Omega$ of the \textit{p-level} grid. Then we compute the \textit{elementary effect} $\text{EE}_i$ of  $i$th hyperparameter $X_i$ as
\begin{equation}
\text{EE}_i = \frac{Y(X_1, \ldots, X_{i-1}, X_{i} + \Delta, \ldots, X_{k}) - Y(X_1,\ldots, X_{i-1} , X_{i}, \ldots, X_{n})}{\Delta},
\end{equation}
where  \(\Delta\) is a value in \(\left( \frac{1}{p-1},\ldots ,1-\frac{1}{p-1}\right) \) which is an incremental change in the values of hyperparameter $ X_i $ when $ X_i $ is sampled from \textit{p-level} grid hyperparameter space $\Omega$. In this scenario, for $k$ hyperparameters and $p$ discrete levels, \(\Delta = p/2(p-1) \) indicates the distance (length) between two levels in the hyperspace $\Omega$ along $i$th axis. The total points in the hyperparameter space $\Omega$, therefore, are $p^{k-1}[p-\Delta(p-1)]$ grid points, which increase exponentially as the number of hyperparameters $k$ increases. However, we use a \textit{one-at-a-time} (OAT) sampling technique for generating  $r$ sample points from this space to compute $r$ EEs for each hyperparameter. 

In the OAT sampling technique, hyperparameter $X_i$ value is changed from a grid point $X_i^{(j)}$ to the adjacent grid point $X_i^{(j\pm1)}$ by a length of \(\Delta\) while all other hyperparameters (say $X_{\sim i}$) remain as it is. Then the next hyperparameter $X_{i+1}$ is chosen, whose value is changed while others remain fixed. This way of sampling is a {\color{revise} \textit{uniform, non-repeating random walk} through the grid of hyperspace $\Omega$ (we call it \textit{Morris}~\citep{MorrisOriginal}). Another way of sampling points (a set $ \textbf{X} $ of hyperparameters) from the hyperspace $\Omega$ is  {\color{revise} to} use the Latin Hypercube Sample (LHS) based Morris method (\textit{Morris LHS})~\citep{Campolongo2007}, which is a stratified sampling approach to cover all region of the hyperspace $\Omega$.} 
Here, we typically select $r$ sample points for each hyperparameter $ X_i $. Hence, both OAT-based Morris LHS and Morris sampling methods give us $r(k+1)$ sample points.
 
We measure two indices $\mu_i$ and $\sigma_i$ indicate \textit{mean} (central tendency) and \textit{standard deviation} of  EE$_i$ of $i$th hyperparameter $X_i$. The measure 
\begin{equation}
\label{eq:EEmean}
\mu_i = \frac{1}{r}\sum^r_{j=1} EE^j_i
\end{equation}
indicates  the overall influence of a hyperparameter $X_i$ where a larger measure of $\mu_i $ means a larger \textit{overall} individual ability to influence the outputs of an algorithm. We also measure the standard deviation $\sigma_i$ of EE$_i$ as
\begin{equation}
\label{eq:EEsd}
\sigma_i = \sqrt{\frac{1}{r-1}\sum^r_{j=1} (EE^j_i - \mu_i)^2},
\end{equation}
where a large measure of $\sigma_i$ indicates that a hyperparameter has high interaction with other hyperparameters. The measure $\sigma$ is an ensemble influence. That is, if $\sigma_i$ has a high value, which means that the computed $r$ elementary effects EE$_i^r$ of $i$th hyperparameter $X_i$ varied a lot because of the variation in the values of other hyperparameters as well. Whereas a low value of $\sigma_i$ means small differences in the computed $r$ elementary effects EE$_i^r$ of the $i$th hyperparameter $X_i$. This indicates that the influence of a hyperparameter on a model’s output is independent of the choice of other hyperparameters values. However, to understand the influence of a hyperparameter, both $\mu$ and $\sigma$ measures need to be seen together (see Fig.~\ref{fig:sa_interpretation}). We normalized the values of $\mu_{i}$ and $\sigma_{i} $ between $ 0 $ and $ 1 $ to effectively show results as per Fig.~\ref{fig:sa_interpretation}.

\subsection{Variance-Based Sensitivity Analysis}
The \textit{variance-based sensitivity analysis} is known as the ``Sobol method''~\citep{SOBOL}, and it shows how much variance of a model's output depends on its inputs. It is an in-depth sensitivity analysis method that uses two sensitivity indices: (a) \textit{first-order effect} $ S_i $ to indicate a direct effect of a hyperparameter $ X_i $ on a model's output $ Y = f(\textbf{X}) $ and (b) \textit{total effect} $ ST_i $ to indicate a hyperparameter $ X_i $ interaction with its complementary parameters $ X_{\sim i} $. 

\begin{figure}
	\centering
	\includegraphics[width=0.8\linewidth]{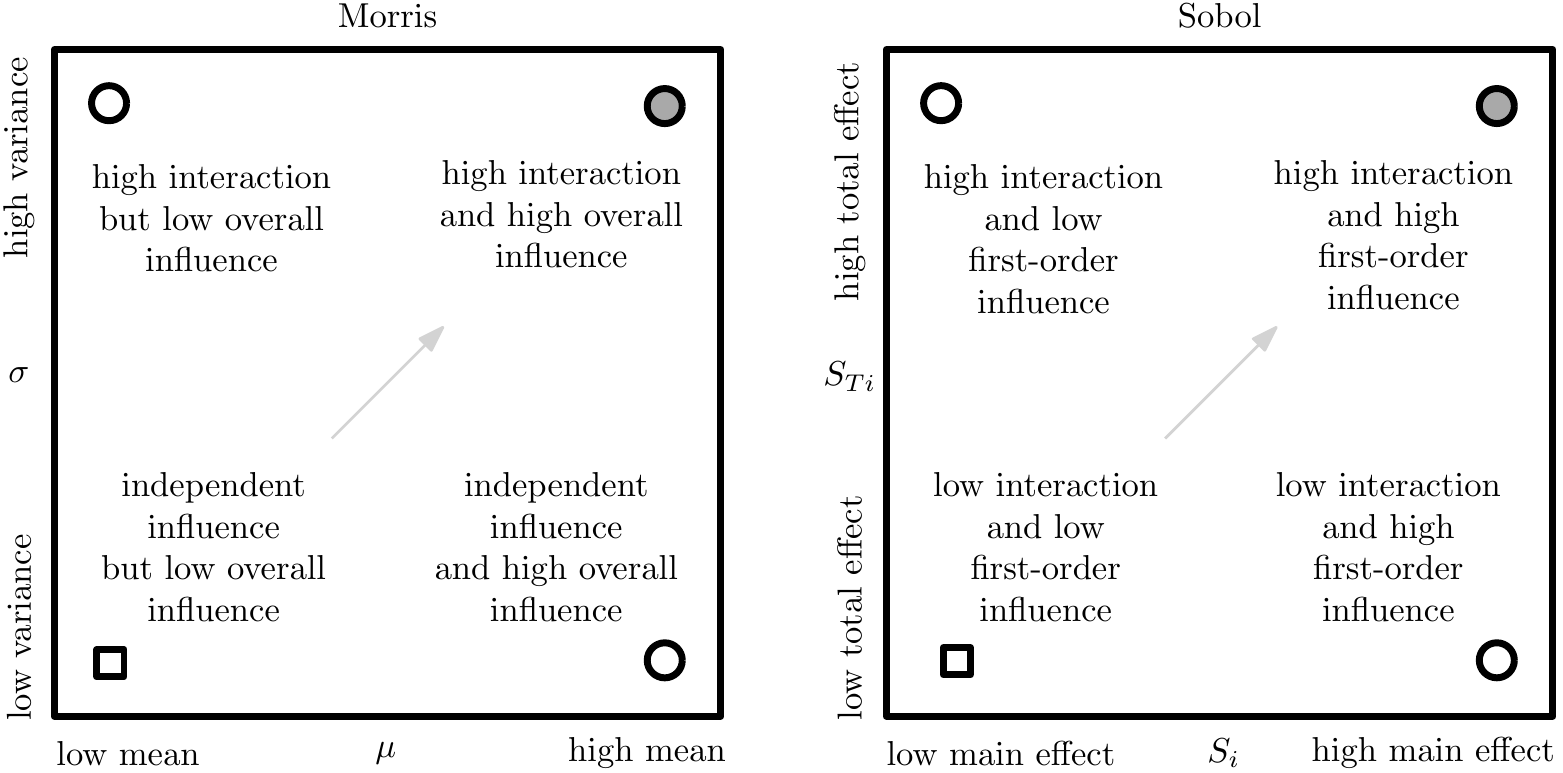}
	\caption{Morris (\textit{left}) and Sobol (\textit{right}) indices interpretation. Top right corner circle in dark gray is the ideal case where a hyperparameter has high individual influence and high interaction (or total effect). Circles in white at the top left and bottom right corners are cases that {\color{correct}have} high importance in at least one direction. Bottom left square in white shows the least ideal case where hyperparameters are non-influential, and fixing them at any values within their defined domain will not influence the algorithm’s performance. Arrow along the diagonal direction indicates the order of the hyperparameters' importance and influence. 
	\label{fig:sa_interpretation}}
\end{figure}

The direct effect $S_i $, irrespective of the hyperparameter interaction $ST_i$, indicates that, on average, how much the model's variance $ V[Y(\textbf{X})]$ could be reduced if the hyperparameter $X_i$ is fixed to a value. Meaning a low value of $S_i $ shows that the variance of the model’s output $ Y(\textbf{X} | X_i = x_i^*) $ does not depend on $ X_i $, and fixing $ X_i $ to a value does not have much impact on the model's output, while for a high value of $S_i $, it strongly does. Indeed, a low value of $S_i$ indicates that $i$th hyperparameter's influence is negligible. Similarly, the interaction effect or total effect $ST_i = 0$ indicates that the model's output $ Y(\textbf{X} | X_i) $ does not depend on $ X_i $, and it is a non-influential parameter. The large values of interaction effect or total-effect $ST_i$ show proportionally strong interactions between the hyperparameter $ X_i $ and its complementary parameter $ X_{\sim i} $. {\color{revise} The difference $ST_i - S_i \ge 0 $, i.e., total interaction influence minus direct influence, shows how much $i$th hyperparameter is involved in interaction with other hyperparameters.} We normalized the values of $S_{i}$ and $ST_i $ between $ 0 $ and $ 1 $ for lucid interpretation of their influence (see Fig.~\ref{fig:sa_interpretation}).

The {first-order effect} $ S_i $ and {total effect} $ ST_i $ of Sobol method are computed as
\begin{equation}
\label{eq:sobol1}
S_i = \frac{V(E(Y|X_i))}{V(Y)} =  \frac{y_A \cdot y_{C_i}-f_0^2}{y_A \cdot y_A - f_0^2} =  \frac{\frac{1}{N}\sum_{j=1}^N y^j_A  y^j_{C_{i}} - f_0^2}{\frac{1}{N}\sum_{j=1}^N (y^j_A)^2 - f_0^2} 
\end{equation}
and
\begin{equation}
\label{eq:sobol2}
ST_i = 1-\frac{V(E(Y|X_{\sim i}))}{V(Y)} = 1 - \frac{y_B \cdot y_{C_i} - f_0^2}{y_A \cdot y_A - f_0^2} = 1 - \frac{\frac{1}{N}\sum_{j=1}^N y^j_B y^j_{C_i}  - f_0^2}{\frac{1}{N}\sum_{j=1}^N (y^j_A)^2 - f_0^2},
\end{equation}
where \(N\) is the number of random samples, $y_A = f(A)$, $y_B = f(B)$ and $y_{C_i} = f(C_i)$ are model output vectors on sample matrix $A, B$ and $C_i$ respectively; and the estimated mean  \(f_0^2\) is
\begin{equation}
\label{eq:sobolmontecarlomean}
f_0^2 = \left( \frac{1}{N}\sum^N_{j=1} y^j_{A} \right)^2.
\end{equation}
Matrices $A_{N \times k}$ and $B_{N \times (k-2k)}$ are random sample points (hyperparameter values), and each matrix $C_i$ is formed by taking all columns of matrix $B$ except $i$th column, which is taken from $i$th  column of matrix ${A}$. Such a sampling is similar to OAT sampling, except its rows are not sorted in any specific order, and all elements in a row differ from the other elements in the row.

%
%
\section{Experiments}
\label{sec:experimentalSetup}
Our sensitivity analysis framework has four essential structural components:
\begin{enumerate}
    \item setup of EAs tunable hyperparameters and optimization problems
    \item sampling of hyperparameters from  hyperparameter space of respective sensitivity analysis methods for respective algorithms
    \item evaluation of EAs on optimization problem (testbench) for all sampled  hyperparameter points and for each hyperparameter sample, the evaluation of respective EAs over a number of independent instances to obtain stable results and to observe expected (average-case) performance of algorithms over performance measures
    \item computation of Morris and Sobol indices
\end{enumerate}
%
In the experiment, all EAs start with a population of initial candidate solutions (uniformly randomly drawn from $\mathbb{R}^n$, $n$ being dimensionality of the problem). Other commonalities among EAs are evolutionary operators like ``selection,''  ``mutation,'' and/or ``crossover'' for generating new (offspring) population and their evaluation. EAs repeat this process for a number of generations until a \textit{termination condition} is met. We set the \textit{termination condition} to be the desired number of \textit{function evaluations}, and we set this to a value of \num{10000} for all four algorithms for all problems. The other hyperparameters setting for our experiments were as follows: 

\subsection{Single-Objective Algorithm Hyperparameters \label{sec:single_exp_setup}}
We analyzed two single-objective EAs over {\color{revise}33 optimization problems:} {\color{insert}23 problems from testbench introduced in~\citep{Tesbench}, and we took 10 optimization problems regarding shifted problems from CEC2014 (shifted Sphere, Ellipsoid, Ackley, and Griewank; and shifted and rotated Rosenbrock, and Rastrigin) and CEC 2015 (shifted and rotated Weierstrass, Schwefel, Katsuura, HappyCat)~\citep{liang2006performance,liang2013problem,liang2014problem}}. An EA needs to find a single optimal solution for an SOO problem in a few generations at the expanse of some wall-clock  time. Hence, the \textit{Best Solution} was used for SOO evaluation. Table~\ref{tab:param_so_ea} lists the hyperparameter tuning space of CMA-ES~\citep{CMAES} and DE~\citep{DEOriginal} algorithms.

The sensitivity analysis method setup for single-objective optimization was as follows. We used {\color{revise}$p = 10$} grid levels to form the hyperparameter space $\Omega$ for respective single objective EAs. From this hyperparameter space, we select {\color{revise}$r = 50$} sample points for each hyperparameter of CMA-ES and DE in the cases of \textit{Morris LHS} and \textit{Morris} methods (see Equations~\eqref{eq:EEmean} and \eqref{eq:EEsd}). This gave us {\color{revise}$300$ and $400$} sample points in total for CMA-ES and DE algorithms, respectively. The Sobol analysis is $2+k$ times more expensive than Morris methods since it evaluates hyperparameter matrices $A$, $B$, and $C_i$, $i = 1,2,\ldots, k$. {\color{revise}For Sobol, we use $N = 100$, which} gave us $700$ and $900$ sample points in total for CMA-ES and DE algorithms, respectively.
\begin{table}
    \centering
    \small
    \setlength{\tabcolsep}{3pt}
    \caption{Hyperparameter domain range of CMA-ES~\citep{CMAES} and DE~\citep{DEOriginal}. For both algorithms, the termination condition was $10,000$ function evaluations.\label{tab:param_so_ea}}
    \begin{tabular}{c |l p{4cm} p{9cm}}
        \toprule
        \multicolumn{1}{c}{Algo} & Params & Domain & Description \\
        \midrule
        \multirow{5}*{\rotatebox[origin=r]{90}{CMA-ES}} 
        & $\lambda $ & $ [10, 1000] $  & Population size\\ 
        & $\alpha_{\mu} $ & $ [0, 4]  $ & Learning rate\\ 
        & $\sigma_0 $ & $  [0.1, 2]  $  & Initial step size\\
        & $\sigma_{0-scale} $ & $ \{\text{False}, \text{ True}\} $ &  Re-scaling of $ \sigma_0  $: convergence speed controller \\
        & $\mu\lambda_{\mathrm{ratio}} $ & $  [0.1, 1]  $ &   Percentage  of population's elements usage in co-variance matrix estimation and update\\
        
        \multicolumn{1}{c}{} & &  & \\
        \multirow{7}*{\rotatebox[origin=r]{90}{DE}}  
        & $ \lambda $  &  $ [10, 1000]  $ &   Population size  \\ 
        &  $\mathrm{X} $ & $ \{\text{bin}, \text{ exp}\}  $ &   Crossover methods: Binomial and Exponential\\ 
        & $ P[\mathrm{X}] $ & $ [0,1]  $ &   Crossover probability\\
        & $ \beta_{\mathrm{min}} $ & $ [0, 1]  $ &  Minimum Acceleration coefficient\\
        & $ \beta_{\mathrm{max}} $ & $ [0, 2]  $ &  Maximum Acceleration coefficient, $ \beta_{\mathrm{max}} = \beta_{\mathrm{min}} + \beta_{\mathrm{max}}$\\
        & $ \mathbf{b}_{\mathrm{type}} $ &  \{``best,'' ``target-to-best,'' ``rand-to-best,'' ``rand''\}  &   Base vector selection methods (mutation type or DE algorithm version)\\ 
        & $ \mathbf{b}\lambda_{\mathrm{ratio}} $ & $ [0.01, 0.5]  $ &   Percentage of base vectors (solution) to be used for difference vectors computation\\ 
        \bottomrule
    \end{tabular}
\end{table}

\subsection{Multi-objective Algorithm Hyperparameters \label{sec:mulobjectiove_exp_set}}
We analyzed multi-objective EAs over a testbench consisting of four families of optimization problems: (i) DTLZ1, DTLZ2, DTLZ3, and DTLZ4~\citep{ProblemSuiteDTLZ}; (ii) IDTLZ1 and IDTLZ2~\citep{ProblemSuiteDTLZ}; (iii) CDTLZ2~\citep{NSGA_III}; and (iv)  WFG3, WFG6, and WFG7~\citep{WFGToolkit}. EAs were evaluated and analyzed for each listed MOO problem for 3 objectives, and each problem was solved as a 10-dimensional problem. This setting was chosen based on the computation effort required for these MOO problems. 

Since the goal of the multi-objective EAs is to obtain a set of solutions where no one objective dominates over the other objectives~\citep{NSGA_II,zhang2007moea}, we use GD (minimization), IGD (minimization), and HV (maximization) as the measures of EA performances  (see Section~\ref{sec:moo_metric}). These metrics result in higher values for a large population size $ \lambda $ compared to a small population size $ \lambda $. Thus, for \textit{population-fair} performance analysis, the metrics were calculated from a union of populations of all generations of EAs and from not only the population of the last generation of the EAs. Moreover, the values were averaged over $30$ independent runs for each sampled set of hyperparameters. 

NSGA-III and MOEA/D have a few common tunable hyperparameters in addition to their subjective tunable hyperparameters. Table~\ref{tab:param_mo_ea}  shows the domain setting of these common and subjective tunable hyperparameters of NSGA-III and MOEA/D.

The sensitivity analysis method setup for multi-objective optimization was as follows. We used $p = 10$ grid levels to form the hyperparameter space $\Omega$ for respective single objective EAs. From this hyperparameter space, we select $r = 20$ sample points for each hyperparameter of CMA-ES and DE in the cases of \textit{Morris LHS} and \textit{Morris} methods (see Equations~\eqref{eq:EEmean} and \eqref{eq:EEsd}). This gave us $140$ and $160$ sample points in total for NSGA-III and MOEA/D algorithms, respectively. In the Sobol analysis, we {\color{correct}used} $N = 30$, and this gave us $240$ and $270$ sample points in total for NSGA-III and MOEA/D algorithms, respectively, for their matrices $A$ and $B$ from which $C_i$ matrices were {\color{correct}created}. The number of sampling points in this work is sufficiently large for good sensitivity analysis~\citep{Campolongo2007,saltelli2002sensitivity,saltelli2008global}.   
  
\begin{table}
    \centering
    \small
    \setlength{\tabcolsep}{3pt}
    \caption{Hyperparameter domain range of NSGA-III and MOEA/D and their shared (\textit{Common}) hyperparameters domain. For both algorithms, the termination condition was $10,000$ function evaluations.
        \label{tab:param_mo_ea}}
    \begin{tabular}{c | l  p{5.5cm} p{7.5cm}}
        \toprule
        \multicolumn{1}{c}{Algo} & Params & Domain & Description \\
        \midrule
        \multirow{5}*{\rotatebox[origin=r]{90}{Common}} 
        &  $ \lambda $ & $  [10, 1000] $ &  Population size.\\
        &  $ P[\text{X}] $ & $ [0, 1] $ & Simulated binary crossover (SBX) probability\\
        &  $ \text{X}_{DI} $ & $ [1, 200] $ &  SBX distribution index\\
        &  $ P[\text{PM}] $ & $ [0, 1] $ &  Polynomial mutation (PM) probability\\
        &  $ \text{PM}_{DI} $ & $ [1, 200] $ &  PM distribution index\\
        \multicolumn{1}{c}{} & &  & \\
        \multirow{2}*{\rotatebox[origin=l]{90}{NSGA-III}} 
        & K & $ [2, 10] $ & Tournament size  \\
        & Selection & Tournament &  Parents selection for offspring generation \\[1em]
        \multicolumn{1}{c}{} & &  & \\
        \multirow{5}*{\rotatebox[origin=r]{90}{MOEA/D}} 
        & $ Mode  $ & \raggedright \{``penalty based boundary intersection (PBI),'' ``Tchebycheff,'' ``Tchebycheff with normalization,'' ``modified Tchebycheff''\} & Method for MOO decomposition into many SOO subproblems\\
        & $\epsilon_N $  &  $[0.05, 0.5]$ & Neighbors: percentage of the population considered as neighbors for each sub-problem generation\\
        \bottomrule
    \end{tabular}
\end{table}

All algorithms, methods, and sensitivity analysis experiments were performed in MATLAB, and  implementations of individual components were taken from MATLAB libraries. We used a \textit{safe toolbox}~\citep{safe} to implement sensitivity analysis sampling methods, indices calculations, and workflows. Single objective algorithms were implemented using ypea library~\citep{ypea}. We used the implementation of multi-objective optimization problems and evaluation measure metrics related to optimization algorithms from PlatEMO library~\citep{PlatEMO}. The entire workflow framework was synchronized with the help of inbuilt functions of MATLAB.

{\color{insert}The whole experiment was expensive to run since the total number of function evaluations was \num{19 014 600 000}. The breakdown of this function evaluation was as follows (each multiplied by \num{10000} concerning termination condition): DE, \num{858580}; CMA-ES, \num{720080}; MOEA/D, \num{171600}; and NSGA-III, \num{151200}. For DE and CMA-ES, there were 33 objective functions, and each one was run at least 10 times for each combination of hyperparameter settings. Similarly, for MOEA/D and NSGA-III, there were 10 functions, and each was run 30 times for stable results for each set. The hyperparameter sets were sampled in three different ways for all algorithms: Morris LHS, Morris, and Sobol,  as mentioned in Sections~\ref{sec:single_exp_setup} and \ref{sec:mulobjectiove_exp_set}. Our implementation of this framework for sensitivity analysis of EAs and results are available in~\citep{ojha2022SaEAs}.}

\section{Results and Discussion}
\label{sec:results}
The results of {\color{revise} sensitivity analysis of each algorithm for their performances} on testbench were collected and processed to produce three indicators: (i) sensitivity analysis indices matrix as per Fig.~\ref{fig:sa_interpretation}, (ii) ordered bar plot arranged from low to high normalized sensitivity analysis total indices values, and (iii) mean score (average performance) of each hyperparameter over select performance measures. Additionally, the statistical tests and clustering analyses results are presented in supplementary Sections A and B. This section describes hyperparameter  \textit{influence}, \textit{ranking}, and \textit{quality} through these three indicators. 

{\color{revise} Each sensitivity analysis method varies how they sample hyperparameter sets as they use strategies such as LHS, OAT based uniform random walk, and OAT based uniform sampling.} Morris LHS and Sobol use the LHS strategy, which means they stratified the hyperspace to draw samples to cover most of the sample space. Morris uses uniform random walk sampling. In summary, each method may present its own ordering of hyperparameters that influence ranking and interpretation. Hence, we are also interested in the commonality of results among methods. 

\subsection{Single-Objective EAs}
\label{sec:soo_results}
\subsubsection{CMA-ES Analysis}
\begin{figure}
	\centering
	\includegraphics[width=\linewidth]{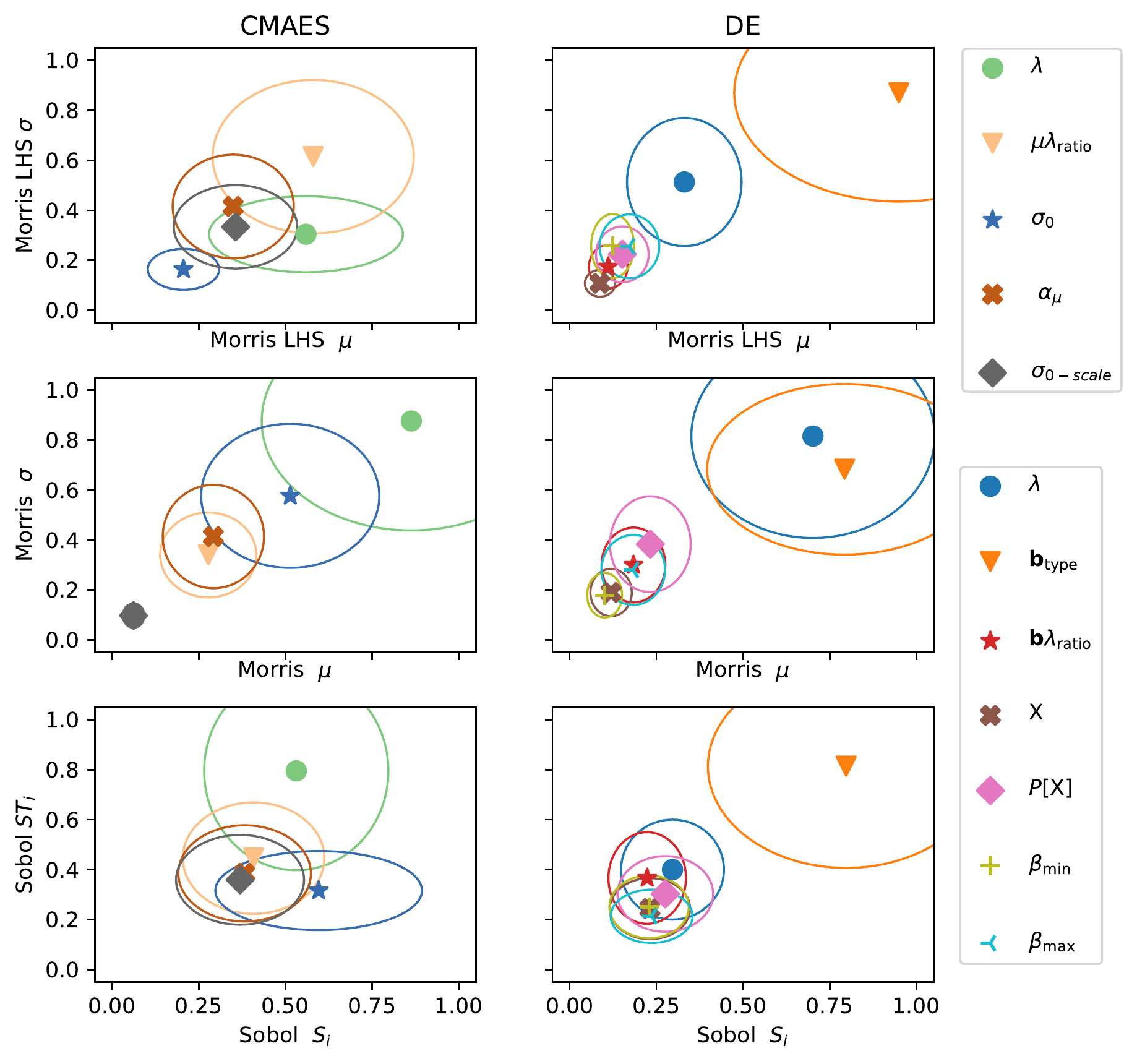}
	\caption{\color{revise}Single objective algorithms sensitivity analysis. CMA-ES and DE hyperparameters sensitivity analysis are shown in column 1 and column 2, respectively. Rows 1, 2, and 3,  respectively, indicate Morris LHS, Morris, and Sobol methods. The upper right legend belongs to CMA-ES and the lower right to DE. A symbol and a color represent each hyperparameter. An eclipse centered at a hyperparameter is the span of the standard deviation of the influence along with direct and interaction influences. A larger width of the eclipse of a hyperparameter in the x-axis direction means more variation in direct dominance of that hyperparameter, and a larger height in the y-axis direction means its variation in total (or interaction) influence. In each plot, the further apart a hyperparameter in the diagonal direction from the origin $(0,0)$ is, the higher its importance to the algorithm. CMA-ES hyperparameter $\lambda$, $\mu\lambda_{\mathrm{ratio}}$, $\sigma_0$, $\alpha_\mu$, and $\sigma_{0-scale}$ respectively are population size, percentage of the population for covariance matrix, initial step size, learning rate, and convergence speed controller. DE hyperparameters $ \lambda $,  $ \mathbf{b}_{\mathrm{type}} $, $ \mathbf{b}\lambda_{\mathrm{ratio}} $, $\mathrm{X} $, $ P[\mathrm{X}] $, $ \beta_{\mathrm{min}} $, and $ \beta_{\mathrm{max}} $ respectively are population size,  base vector selection type (mutation type), percentage of the population for base vector selection, crossover-type, crossover probability, minimum acceleration coefficient, and maximum acceleration coefficient. Table~\ref{tab:param_so_ea} contains the hyperparameter name, definition and domain information.  {\color{insert}Supporting statistical tests~\citep{Kalpic2011} between direct and interaction effects and clustering analysis are provided in supplementary Sections A and B.}
		\label{fig:SO_scatter}}
\end{figure}

\begin{figure}
	\centering
	\includegraphics[width=\linewidth]{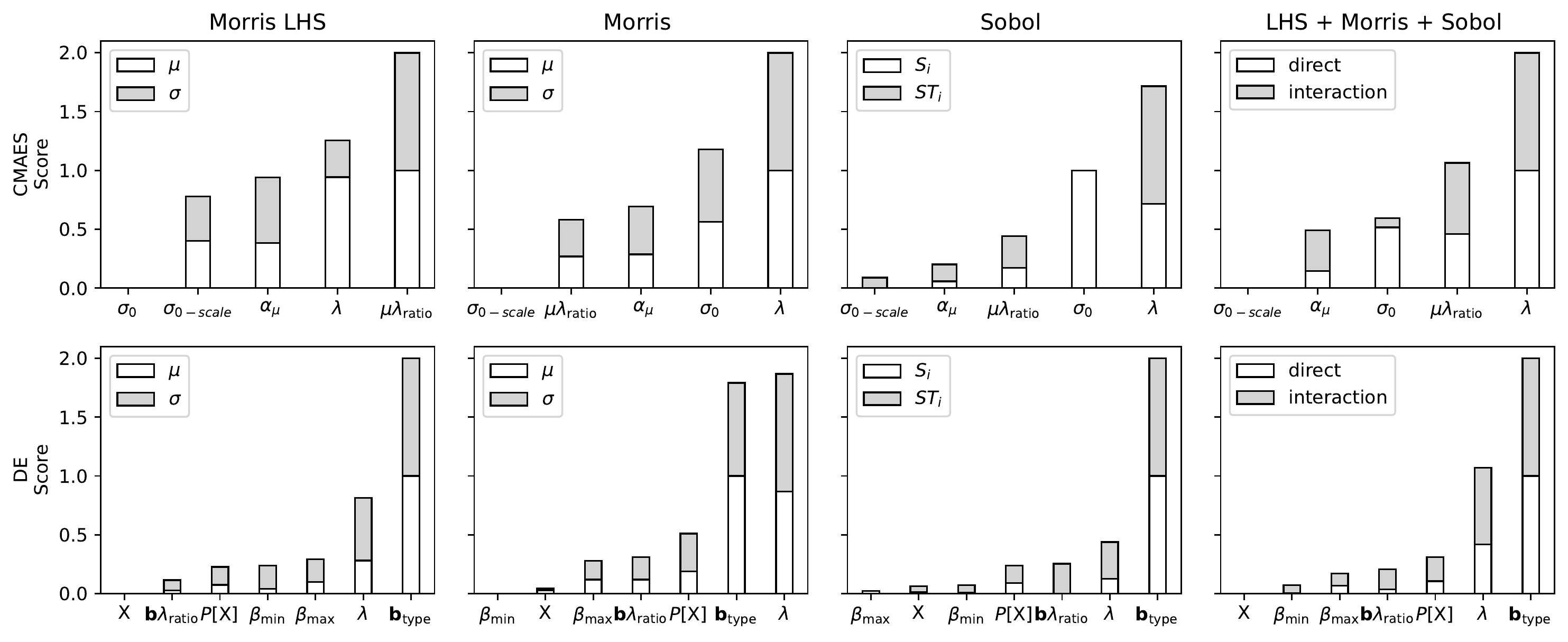}
	\caption{\color{revise}Ordering (small to larger) of the sum of sensitivity analysis indices of single objective algorithms. CMA-ES (row 1) and DE (row 2) algorithms hyperparameters performance across all problems (functions). Columns 1, 2, and 3 respectively show performance evaluated using Morris LHS, Morris, and Sobol methods. The white color portion of a bar is the direct influence normalized value in \([0,1]\) and gray color portion is interaction (total) influence value in \([0,1]\). Larger height bar implies a higher influence. Table~\ref{tab:param_so_ea} contains the hyperparameter name, definition and domain information. 
		\label{fig:SO_bar}}
\end{figure}

\begin{figure}
	\centering
	\includegraphics[width=\linewidth]{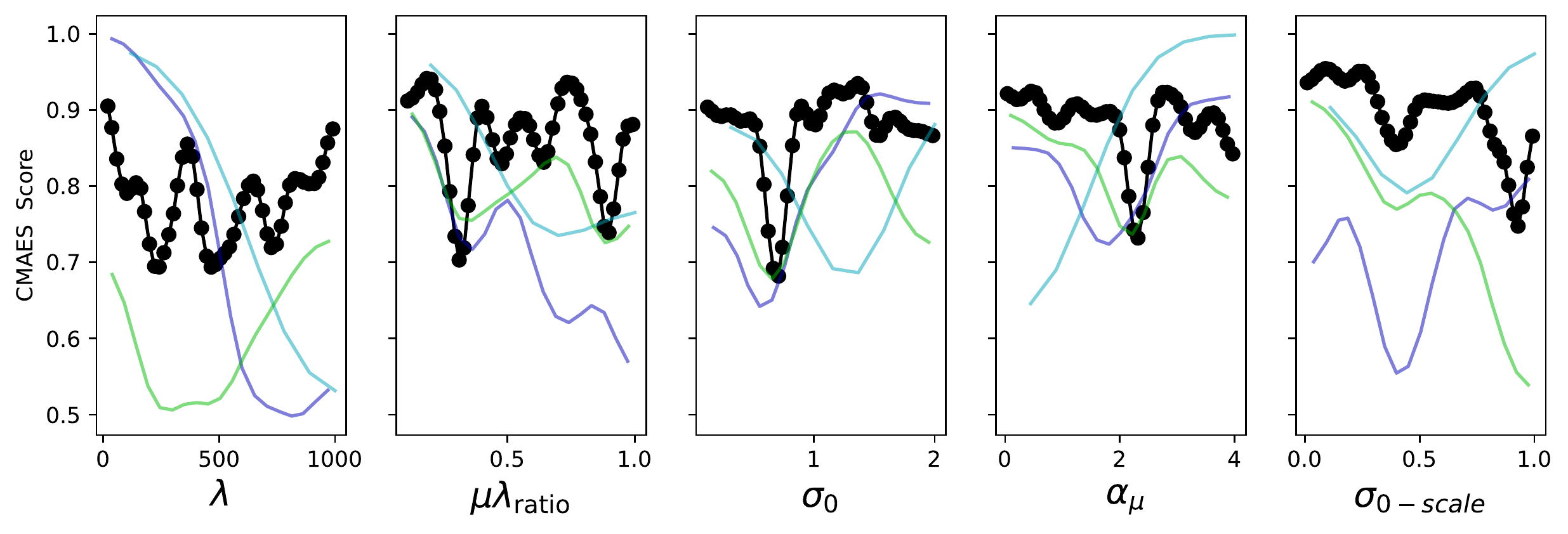}
	\includegraphics[width=\linewidth]{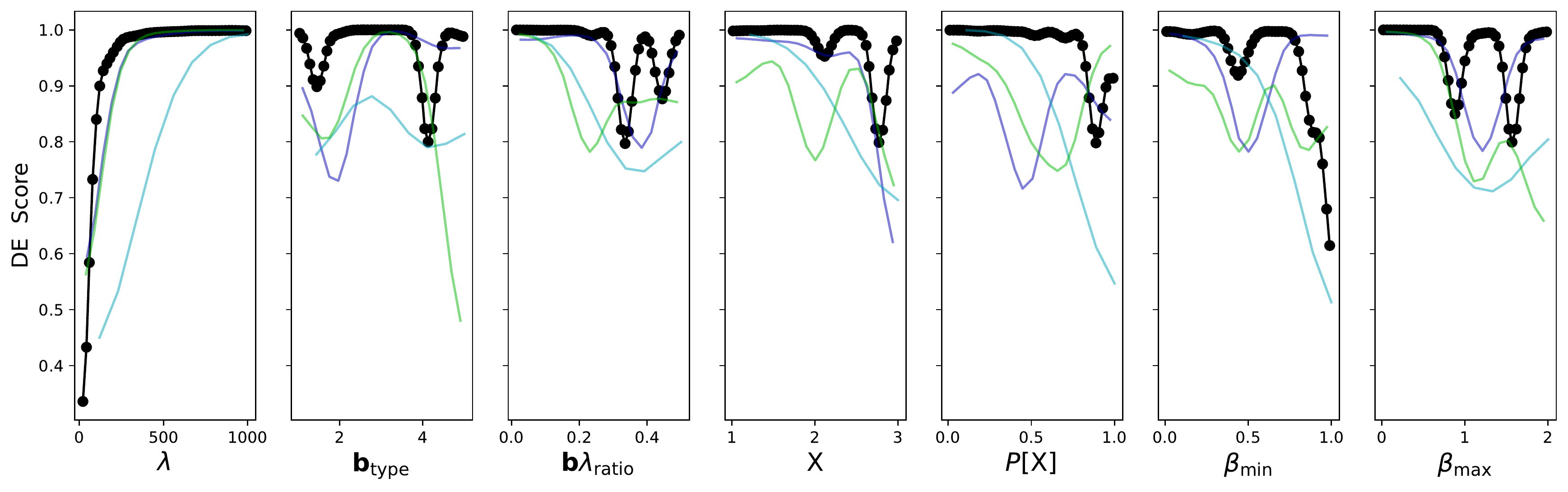}
	\caption{\color{revise}CMA-ES and DE algorithms average performance on {\color{insert}10 runs (for 72 cases, 30 runs)} of each hyperparameter set. CMA-ES and DE algorithms had \num{2740} and \num{2,980} hyperparameter sample sets evaluated in total (black dots) by Morris LHS (blue lines), Morris (cyan lines), Sobol (green lines) methods. Each dot in a subplot is a mean performance of a bin of total samples. Along x-axis there are 50 bins from lower to higher values which are plotted against each hyperparameter normalized score filtered (using Gaussian filter with sigma 2) in the y-axis. 
		\label{fig:SO_line}}
\end{figure}

CMA-ES results are shown in Figs.~\ref{fig:SO_scatter}, ~\ref{fig:SO_bar}, and ~\ref{fig:SO_line}, where Fig.~\ref{fig:SO_scatter} is a scatter plot that presents sensitivity analysis indices as per Fig.~\ref{fig:sa_interpretation}. It shows the tendency of the \textit{quality of influence} a hyperparameter has on CMA-ES performance on all {\color{revise}33 problems} in the testbench. {\color{revise} For instance, $\lambda$, the population size in CMA-ES has a high \textit{overall} influence and high \textit{interaction} influence in all three sensitivity analysis methods. Hence, $\lambda$} is the most significant hyperparameter of the CMA-ES algorithm, and this must be the first hyperparameter one must select to tune for the performance improvement when CMA-ES is applied to solve a problem.

~\\
{\color{revise}
\textbf{Population size $\lambda$.} Population size $\lambda$ is the most influential factor in CMA-ES algorithms. Both Morris and Sobol methods show a strong overall influence and high interaction for $\lambda$. Morris LHS ranked it as a high direct influence but slightly lower interaction influence than covariance matrix size controller $\mu\lambda_{\mathrm{ratio}}$ that has the highest interaction and direct influence in the Morris LHS method. Since MOEA/D decomposes problems into several single-objective problems, unsurprisingly, the size of the population and related hyperparameters are the most influential. This corroborates the fact that they offer exploration capabilities to population-based algorithms, allowing them to search a huge part of the search space concurrently. Fig.~\ref{fig:SO_bar} and Fig.~\ref{fig:SO_line} confirm the significance of $\lambda$ in CMA-ES. Figure~\ref{fig:SO_line} also suggests that variation in CMA-ES performance is very high due to this interaction of population size with other hyperparameters as we observe a highly fluctuating performance of CMA-EA for varied $\lambda$ values.
  

\textbf{Covariance matrix size controller $\mu\lambda_{\mathrm{ratio}}$.}  Hyperparameter $\mu\lambda_{\mathrm{ratio}}$, which controls the percentage of population $\lambda$ to be used for the covariance matrix estimation and update, has high interaction and direct influence on CMA-ES performance. The $\mu\lambda_{\mathrm{ratio}}$ is the second most influential hyperparameter across all three methods (see Fig.~\ref{fig:SO_bar}). The significance of  $\mu\lambda_{\mathrm{ratio}}$ is evident as its values  and the choice of $\lambda$ are closely linked, and the choice of this ratio will increase or decrease the size of the covariance matrix that is at the core of the CMA-ES algorithm functioning. Similar to the performance of $\lambda$, $\mu\lambda_{\mathrm{ratio}}$ performance is largely variable for its values (see Fig.~\ref{fig:SO_line}).

\textbf{Initial step size $\sigma_0$.} Fig.~\ref{fig:SO_bar} confirms the significance of $\sigma_0$ (initial step size) influence as this emerged as the next best hyperparameter in Morris and Sobol methods. Morris LHS, which is a stratified sampling method that covers the most hyperspace region, suggests that $\sigma_0$ is taken from most regions of its possible values and the CMA-ES performance had varied because of such sampling. However, the scores remain relatively high (see Fig.~\ref{fig:SO_line}). The performance $\sigma_0$ in Morris LHS is also impacted by the fact that for almost half of the time, its re-scaling was switched off by $\sigma_{0-scale}$. Accordingly, $\sigma_{0-scale}$ should have a higher influence on Morris LHS than $\sigma_0$, which indeed is the case (see Fig.~\ref{fig:SO_bar}). Examining Fig.~\ref{fig:SO_line}, we may observe that for range $[1,2]$ of $\sigma_0$ values, CMA-ES mean performances were largely consistent (or above certain high scores). More precisely, a range $[0.8, 1.5)$ $\sigma_0$ produces the best performance.}

\textbf{Learning-rate $\alpha_{\mu}$.} Learning-rate $\alpha_{\mu}$ was found to be non-influential. However, since the performance of CMA-ES was consistent with its chosen values across all three methods, the learning-rate $\alpha_{\mu}$ was better than re-scaling $\sigma_{0-scale}$. Moreover, the learning-rate $\alpha_{\mu}$ shows more interaction with other hyperparameters than the convergence speed controller $\sigma_{0-scale}$. This is also evident as the gray bars are larger than the white bars in Fig.~\ref{fig:SO_bar} and drop in performance for only a very small range of values around $2$ in Fig.~\ref{fig:SO_line}).

{\color{revise}
\textbf{Convergence controller $\sigma_{0-scale}$.} CMA-ES convergence controller hyperparameter $\sigma_{0-scale}$, a hyperparameter meant for re-scaling of initial step size $\sigma_0$ on and off, is the least influential in both Morris and Sobol methods (see Fig.~\ref{fig:SO_scatter}). This result is supported by both Fig.~\ref{fig:SO_bar} and  Fig.~\ref{fig:SO_line}. However, it is an influential hyperparameter in the sense that it has a very high influence on $\sigma_0$, which is the third most influential hyperparameter. From Fig.~\ref{fig:SO_line}, it is evident that no re-scaling of $\sigma_0$ performs better than re-scaled $\sigma_0$. This is the reason why for Morris LHS, $\sigma_0$ is the least influence hyperparameter.}

\textbf{CMA-ES hyperparameters ranking.} In summary, we may provide a \textit{ranking of hyperparameters} for CMA-ES from the most to least influential hyperparameter as {\color{revise} $\lambda$, $\mu\lambda_{\mathrm{ratio}}$, $\sigma_0$,}  $\alpha_\mu$, and $\sigma_{0-scale}$. One may ignore tuning $\alpha_\mu$ and $\sigma_{0-scale}$ completely as setting a sufficiently large function evaluation number would neutralize their importance in the CMA-ES algorithm. 


\subsubsection{DE Analysis}
\textbf{DE versions $ \mathbf{b}_{\mathrm{type}} $.} DE results are shown in Figs.~\ref{fig:SO_scatter}, ~\ref{fig:SO_bar}, and ~\ref{fig:SO_line}, where Fig.~\ref{fig:SO_scatter} shows the quality of influence each hyperparameter has on DE performance over {\color{revise}33 problems}. As per  the result of the three sensitivity analysis methods, it is clear that the type of DE base vector selection method (mutation type) $ \mathbf{b}_{\mathrm{type}} $ (DE version) is by far the most significant hyperparameter. Examining Fig.~\ref{fig:SO_line}, we observe that the type of DE mutation strategy \textit{best} and \textit{rand} have lower scores, and they fluctuate. In comparison, the base vector selection methods \textit{target-to-best} and \textit{best-to-best} performed more consistently with high scores. Therefore, the average performance of DE over the testbench was highly sensitive to the selection of $ \mathbf{b}_{\mathrm{type}} $. We also observe that $ \mathbf{b}_{\mathrm{type}} $ in Fig.~\ref{fig:SO_scatter} remains at $(1,1)$ corner of the plot, meaning it had both a high \textit{direct effect} and high \textit{interaction effect}. 

\textbf{Population size $\lambda$.} Overall population size $\lambda$ is the second most influential hyperparameter in DE (cf. Figs.~\ref{fig:SO_scatter} and~\ref{fig:SO_bar}). Comparatively, it had produced better scores for larger population sizes than the smaller population sizes (see Fig.~\ref{fig:SO_line}). However, the size of the population of DE was a distant {\color{revise}second} influential hyperparameter. This indicates that except for small population size (less than 200), DE’s performance was invariable when the population size was increased from 200 to 1000 (Fig.~\ref{fig:SO_line}). {\color{revise}This was when the number of function evaluations was the same for each population size, i.e., the number of function evaluations was \num{10000} for each population size.

\textbf{Crossover-type $ \mathrm{X} $ and crossover probability $ P[\mathrm{X}] $.} The crossover related hyperparameters are the type of crossover $ \mathrm{X} $ and the probability of crossover $ P[\mathrm{X}] $. Between these, $ P[\mathrm{X}] $ plays a vital role in DE's performance, and the type of DE $ \mathbf{b}_{\mathrm{type}} $ was the least influential (cf. Figs.~\ref{fig:SO_scatter} and~\ref{fig:SO_bar}). For crossover-type \textit{binomial} offered better scores than the crossover-type \textit{exponential} (see Fig.~\ref{fig:SO_line}). The crossover probability $ P[\mathrm{X}] $ has its usage only for binomial crossover. Hence, it was an influential hyperparameter in this setting.

\textbf{Base vectors selection pool $\mathbf{b}\lambda_{\mathrm{ratio}}$.} The hyperparameter $\mathbf{b}\lambda_{\mathrm{ratio}}$ defines the percentage of the population used for the selection of base vectors for DE. We found that  $\mathbf{b}\lambda_{\mathrm{ratio}}$ has a negligible influence on the performance of DE (cf. Figs.~\ref{fig:SO_scatter} and \ref{fig:SO_bar}).

\textbf{Acceleration coefficients $\beta_{\mathrm{min}}$ and $\beta_{\mathrm{max}}$.} Similar to $\mathbf{b}\lambda_{\mathrm{ratio}}$, acceleration coefficients hyperparameters $\beta_{\mathrm{min}}$ and $\beta_{\mathrm{max}}$ are least significant in DE. However, setting an appropriate range is vital for DE performance, as we observed in Fig.~\ref{fig:SO_line}. This is evident because $\beta_{\mathrm{min}}$ and $\beta_{\mathrm{max}}$ acquire a relatively moderate influence in Morris LHS methods (see Fig.~\ref{fig:SO_bar}). Since the Morris LHS method uses a stratified sampling approach, it forced the selection of $\beta_{\mathrm{min}}$ and $\beta_{\mathrm{max}}$  values across their whole range and the performance of DE is impacted negatively by the higher values of $\beta_{\mathrm{min}}$ and $\beta_{\mathrm{max}}$. Examining Fig.~\ref{fig:SO_line}, we observed that $\beta_{\mathrm{min}}$ scores for values in $[0.0, 0.5]$ performed consistently with  better scores than the values in $(0.5, 1.0]$. However, Morris and Sobol had a uniform distribution and show that the influence of this hyperparameter is non-influential; therefore, setting these values somewhere in $[0.0, 0.5]$ will suffice, and one may not need to exhaustively tune this hyperparameter.  

Similarly, $\beta_{\mathrm{max}}$ was found sensitive to its range selection. Fig.~\ref{fig:SO_line} offers us the ways to investigate which range had a positive influence and which had a negative. We observe that the lower values had higher scores than the larger values of $\beta_{\mathrm{max}}$ (see Fig.~\ref{fig:SO_line}). Investigating closely, we found that scores in $[0.2, 0.9]$ are by far better than the scores for other values. This means tuning $\beta_{\mathrm{max}}$ values within range $[0.2, 0.9]$ for a problem is an appropriate strategy. 
}

\textbf{DE hyperparameters ranking.} In summary, we \textit{rank} DE hyperparameters from the most significant to least significant as $ \mathbf{b}_{\mathrm{type}} $, $ \lambda $, {\color{revise} $ P[\mathrm{X}]$, $ \mathbf{b}\lambda_{\mathrm{ratio}} $, $\beta_{\mathrm{max}}$, $ \beta_{\mathrm{min}} $, and $\mathrm{X}$. That one would safely use DE with binomial crossover and set appropriate values (discussed above) of $\beta_{\mathrm{max}}$, $ \beta_{\mathrm{min}} $.}

{\color{insert}
\subsubsection{Remarks on SOO hyperparameter rankings and algorithms}
We evaluated two single objective optimization algorithms and presented rankings of their hyperparameter influence based on a combined assessment of three sensitivity analysis methods. Each method, as mentioned, has its own way of drawing samples from the hyperparameter space and thus has produced its own ranking. However, the results reveal some obvious signs of influence based on direct and interaction effects. 

Supplementary A provides rich information on statistical tests among hyperparameters that one can thoroughly examine to reach the presented ranking and may find more information should one is interested in studying specific hyperparameters. For instance, the interaction effect of population size in CMA-ES is more significant than its direct effect (see Supplementary A), which confirms the analysis presented in Fig.~\ref{fig:SO_line}. Additionally, clustering analysis of hyperparameters and objective function provides information behaviors of the algorithm on the class of problems they solve (see supplementary B). For example, the type of mutation in DE forms a distinct cluster of its performance characteristics, and all other hyperparameters are grouped together in one cluster (see supplementary B). 

As a consequence of this analysis and the results presented in Section~\ref{sec:soo_results}, it is clear that \textit{DE is a more stable algorithm than CMA-ES}. See variation in scores of the hyperparameters of the CMA-ES algorithm compared to DE's hyperparameters in Fig.~\ref{fig:SO_line} and high interaction among CMA-ES's hyperparameters. In contrast, DE has a clear ranking of hyperparameters. Additionally, during the experiments, CMA-ES failed to solve some classes of problems for some combination of hyperparameters (see results in \cite{ojha2022SaEAs}).}

\subsection{Multi-objective EAs}
\subsubsection{NSGA-III Analysis}
\begin{figure}
	\centering
	\includegraphics[width=\linewidth]{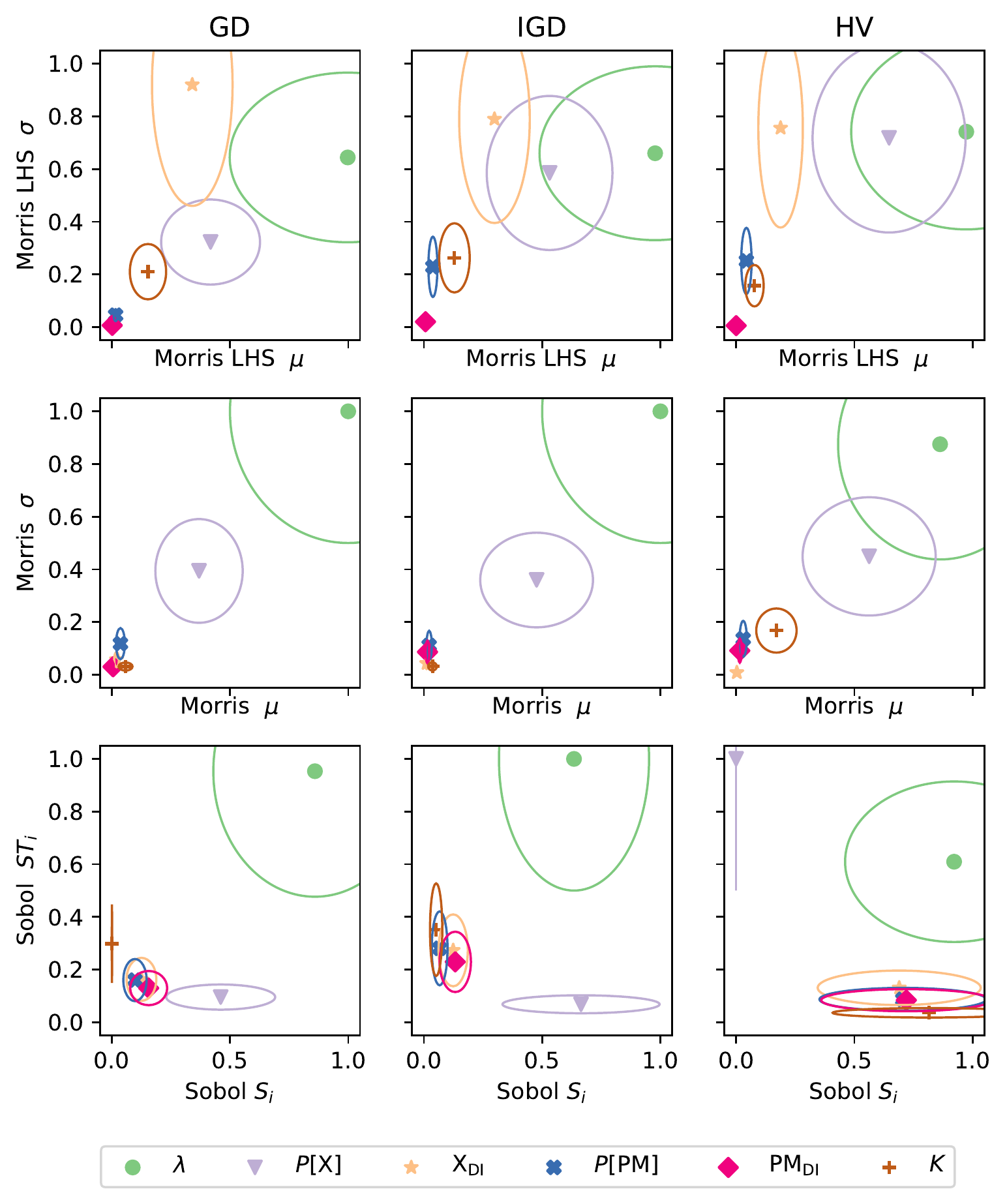}
	\caption{NSGA-III hyperparameters sensitivity analysis. Columns 1, 2, and 3 respectively indicate performance metrics GD, IGD, and HV. Rows 1, 2, and 3, respectively, indicate Morris LHS, Morris, and Sobol methods. Legends of hyperparameter are shown at the bottom. Each hyperparameter is represented by a symbol and a color. An eclipse centered at a hyperparameter is the standard deviation of its influence and direction of its influence. The further apart a hyperparameter in the diagonal direction from the origin $(0,0)$ is, the higher its importance to the algorithm. A larger width of the eclipse of a hyperparameter in the x-axis direction means more variation in the direct influence of a hyperparameter, and a larger height in the y-axis direction means variation in total (or interaction) influence. 
	{\color{insert}Supporting statistical tests and clustering analysis are provided in supplementary  Sections A and B.} 
		\label{fig:NSGA3_scatter}}
\end{figure}

\begin{figure}
	\centering
	\includegraphics[width=\linewidth]{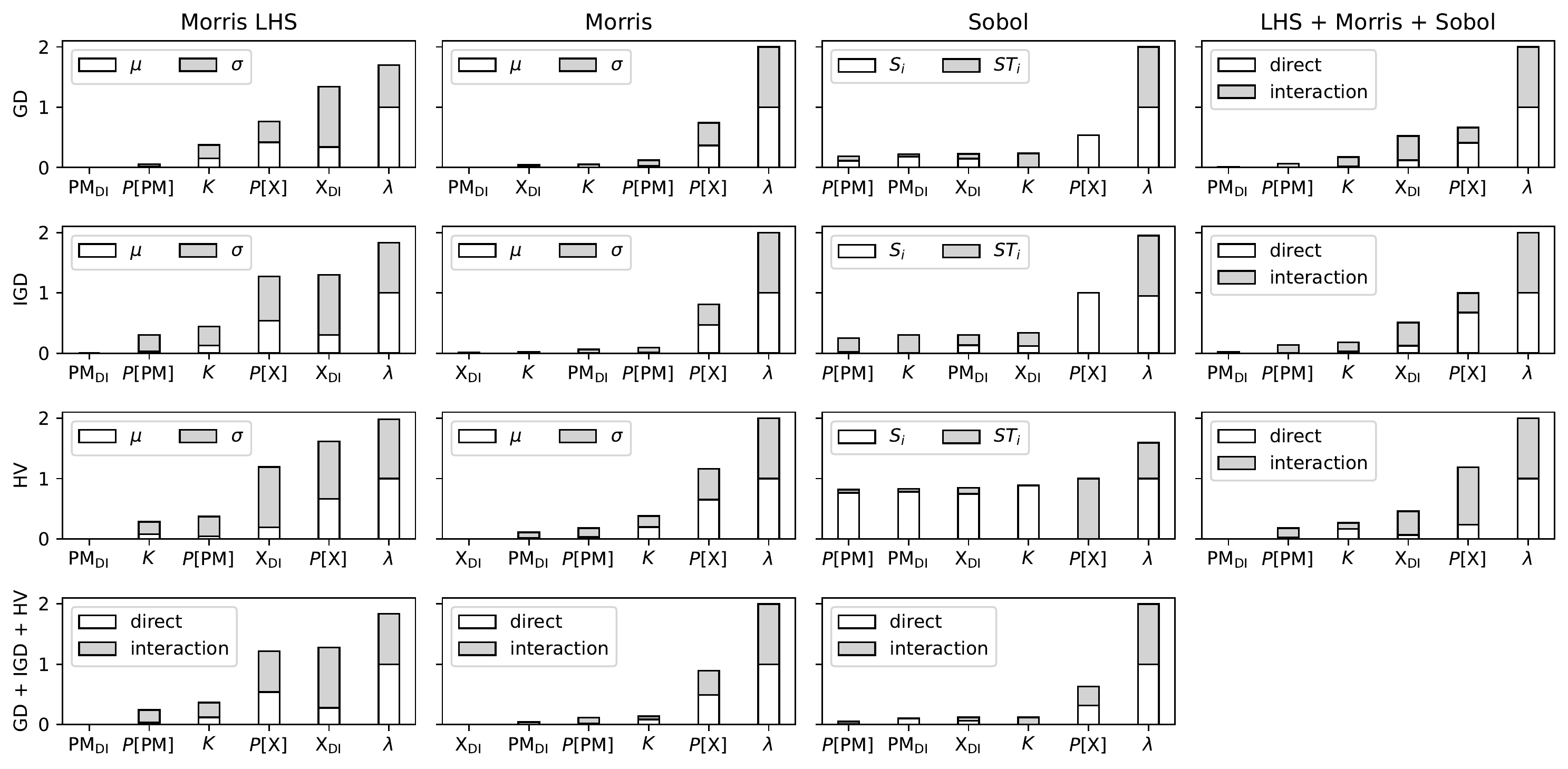}
	\caption{NSGA-III algorithm's hyperparameters performance across all problems (functions). Rows 1, 2, and 3 respectively, show performance evaluated using GD, IGD, and HV metrics. Columns 1, 2, and 3, respectively indicate Morris LHS, Morris, and Sobol methods. The white color portion of a bar is direct influence normalized value in \([0,1]\) and gray color portion is interaction (total) influence value in \([0,1]\). A larger height bar implies a higher influence, and hyperparameters in each subplot are arranged from low to high influence.
	\label{fig:NSGA3_bar}}
\end{figure}

\begin{figure}
	\centering
	\includegraphics[width=\linewidth]{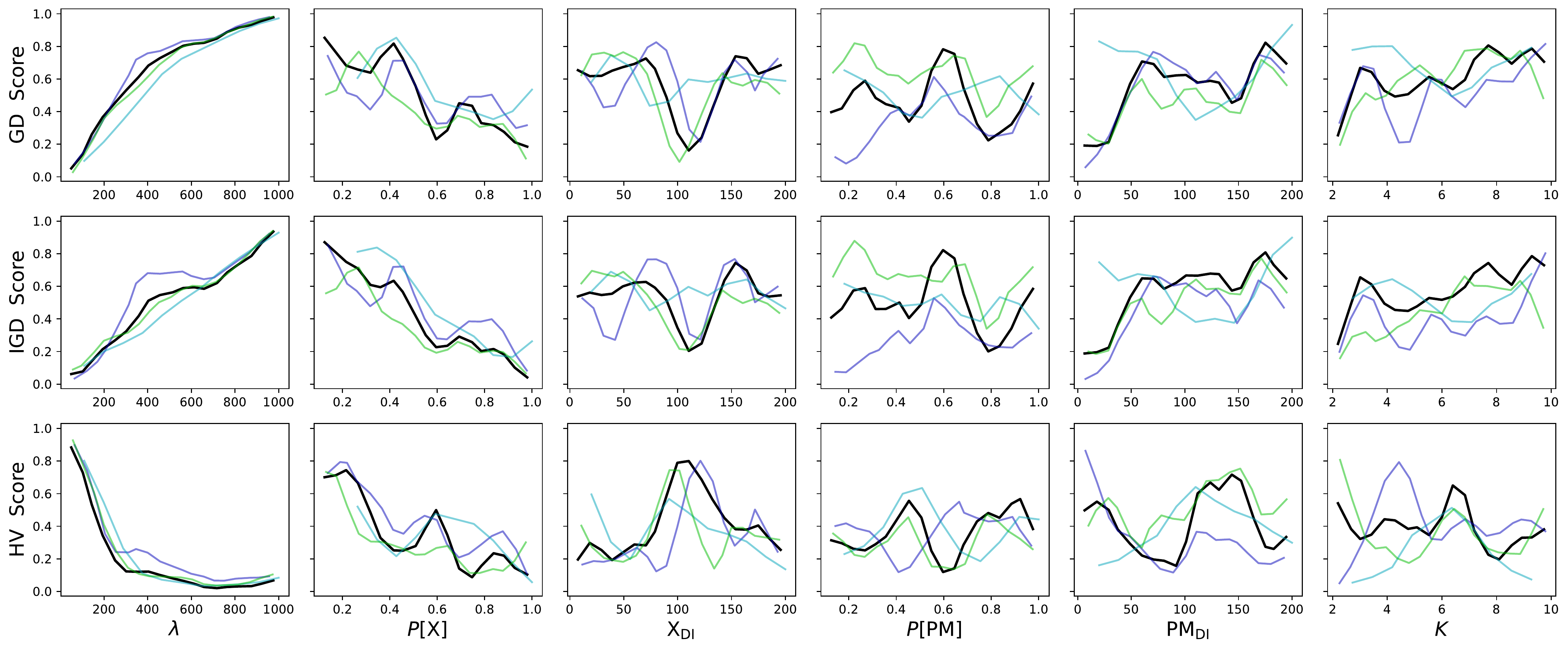}
	\caption{NSGA-III algorithm average performance on 30 runs of each set of hyperparameters. NSGA-III hyperparameter (x-axis) against the mean metric value (y-axis). Rows 1, 2, and 3, respectively, are GD, IGD, and HV metrics. The scores are normalized between 0 and 1  and smooth out using a Gaussian 1D filter with sigma 0.99. The y-axis is GD, IGD, and HV metrics values normalized between a score of 0 and 1, where 0 is the best score for GD and IGD, and 1 is the best score for HV. A total of 520 samples were evaluated for the NSGA-III algorithm jointly by  Morris LHS (blue lines), Morris (cyan lines), and Sobol (green lines) methods. The hyperparameter values are arranged in 20 bins (lower values to higher values) across the x-axis. Each line in each plot connects the mean values of 20 bins of such samples.
		\label{fig:NSGA3_line}}
\end{figure}

\textbf{Population size $ \lambda $.} Results of NSGA-III are presented in Figs.~\ref{fig:NSGA3_scatter}, \ref{fig:NSGA3_bar}, and \ref{fig:NSGA3_line}. In Fig.~\ref{fig:NSGA3_scatter}, we present results of three measures GD, IGD, and HV; see columns in Fig.~\ref{fig:NSGA3_scatter}, and along rows in Fig.~\ref{fig:NSGA3_scatter}, we present Morris LHS, Morris, and Sobol sensitivity methods. For NSGA-III, we clearly observe that the population size $ \lambda $ is a significant hyperparameter, and the probability of crossover is the second most significant hyperparameter. Population size influence has approximately equal high direct influence and high interaction influence. {\color{revise} That is, although population size is the most significant hyperparameter, NSGA-III performance varied because of the variation of the other hyperparameters as well (see NSGA-III has a monotonous line for $\lambda$ in Fig.~\ref{fig:NSGA3_line} that indicates a more liner influence on NSGA-III).} This fact was found true across all methods and all measures as the eclipse of its influence centered around coroner $(1,1)$ in Fig.~\ref{fig:NSGA3_scatter}, and the white and gray bars have comparable lengths in Fig.~\ref{fig:NSGA3_line}. 

An examination of scores of the population size shows that population size does not fluctuate much for the HV metric after a certain population size, but for GD and IGD metrics, the scores keep increasing for increasing population size (see Fig.~\ref{fig:NSGA3_line}). However, this is monotonous, and one would expect such performance for GD and IGD metrics. The probability of crossover shows more fluctuations in all three metrics. Therefore, the variations in the performance of NSGA-III after a sufficiently large population size (in this case, $ 200 $) come from the variations of other hyperparameters, {\color{insert}including  crossover probability.}

{\color{revise}
\textbf{Crossover and mutation hyperparameters.} The probability of crossover $ P[\mathrm{X}] $ shows a more linear relationship between its values and NSGA-III performance measures GD, IGD, and HV. For increasing values of crossover probability, we see decreasing GD and IGD scores (signs of better performance) and increasing scores of HV for some values (see Fig.~\ref{fig:NSGA3_line}). A crossover rate of around $0.6$ leads to better solutions along the problem's objective dimensions, i.e., increasing scores of HV and lower scores of GD and IGD. This fact is supported by the strong direct and interaction influence of crossover $ P[\mathrm{X}] $ {\color{revise} for IGD and HV metrics and relatively direct influence on GD.} The Sobol method on $ P[\mathrm{X}] $ does show a very strong total influence compared to direct influence on all metrics. In summary, the $ P[\mathrm{X}] $ performance has a behavior of monotonous increase and is one of the most influential hyperparameters in NSGA-III. 

For crossover related hyperparameter crossover distribution indices $X_{\text{DI}}$, the performance remains consistent and largely non-influential (cf. Figs~\ref{fig:NSGA3_scatter} and ~\ref{fig:NSGA3_line}) as only for a certain range of its value (a small range around 100), it shows a spike in the performance of NSGA-III. Similarly, the mutation distribution index $\text{PM}_{\text{DI}}$,  the performance of NSGA-III is better for a certain range (around $100 -- 150$ or low values of $\text{PM}_{\text{DI}}$, see Fig.~\ref{fig:NSGA3_line}). For both $X_{\text{DI}}$ and  $\text{PM}_{\text{DI}}$, this phenomenon occurred roughly around a value of $100$ of these indices, which aligned with the range for these hyperparameters suggested in \citep{deb2014analysing,deb1995simulated}.

Similar to the probability of crossover $ P[\mathrm{X}] $, the probability of mutation $P[\text{PM}]$ shows a sudden change in performance around a value of $0.6$, but in a complementary direction (see a drop in HV and spike in GD and IGD metric in Fig.~\ref{fig:NSGA3_line}). The direct and interaction influence of mutation related hyperparameters $P[\text{PM}]$ and $\text{PM}_{\text{DI}}$ is low for NSGA-III (cf. Figs.~\ref{fig:NSGA3_scatter} and \ref{fig:NSGA3_bar}).
} 

\textbf{Tournament size $K$.} Tournament size $K$, probability of polynomial mutation $P[\text{PM}]$, polynomial mutation distribution index $\text{PM}_{\text{DI}}$ and simulated binary crossover (SBX) distribution index $X_{\text{DI}}$ have comparable significance. However, they differ in different methods and metrics. Among these hyperparameters, tournament size $K$ clearly shows a high influence on NSGA-III performance. Tournament size $K$ shows more interaction influence than direct influence, except for the HV metric of the Sobol method. The high score of $K$ in Fig.~\ref{fig:NSGA3_line} with clear fluctuation is the evidence of its interaction with other hyperparameters, but the scores (especially in GD and IGD scores) show an upward trend, indicating it has comparatively less influence on guiding the population towards true Pareto-front than hyperparameters $P[\text{PM}]$, $\text{PM}_{\text{DI}}$ and $X_{\text{DI}}$. We may also confirm that the lower value of $K$ is more influential than its higher values.

\textbf{NSGA-III hyperparameters ranking.} Considering the hyperparameters' performance influence, we rank them from most influential to least influential hyperparameters as $\lambda$, $ P[\mathrm{X}] $, {\color{revise} $X_{\text{DI}}$, $K$}, $P[\text{PM}]$, and $\text{PM}_{\text{DI}}$. Here, $\lambda$ is effective up to a certain population size, and then $\lambda$ saturates. The tuning of crossover $ P[\mathrm{X}] $ linearly influences NSGA-III, and  $X_{\text{DI}}$,  $P[\text{PM}]$, and $\text{PM}_{\text{DI}}$ require setting a fixed value, but their influence fluctuates, i.e., they are affected by the setting of values of other hyperparameters a lot.

\subsubsection{MOEA/D Analysis}
\begin{figure}
	\centering
	\includegraphics[width=\linewidth]{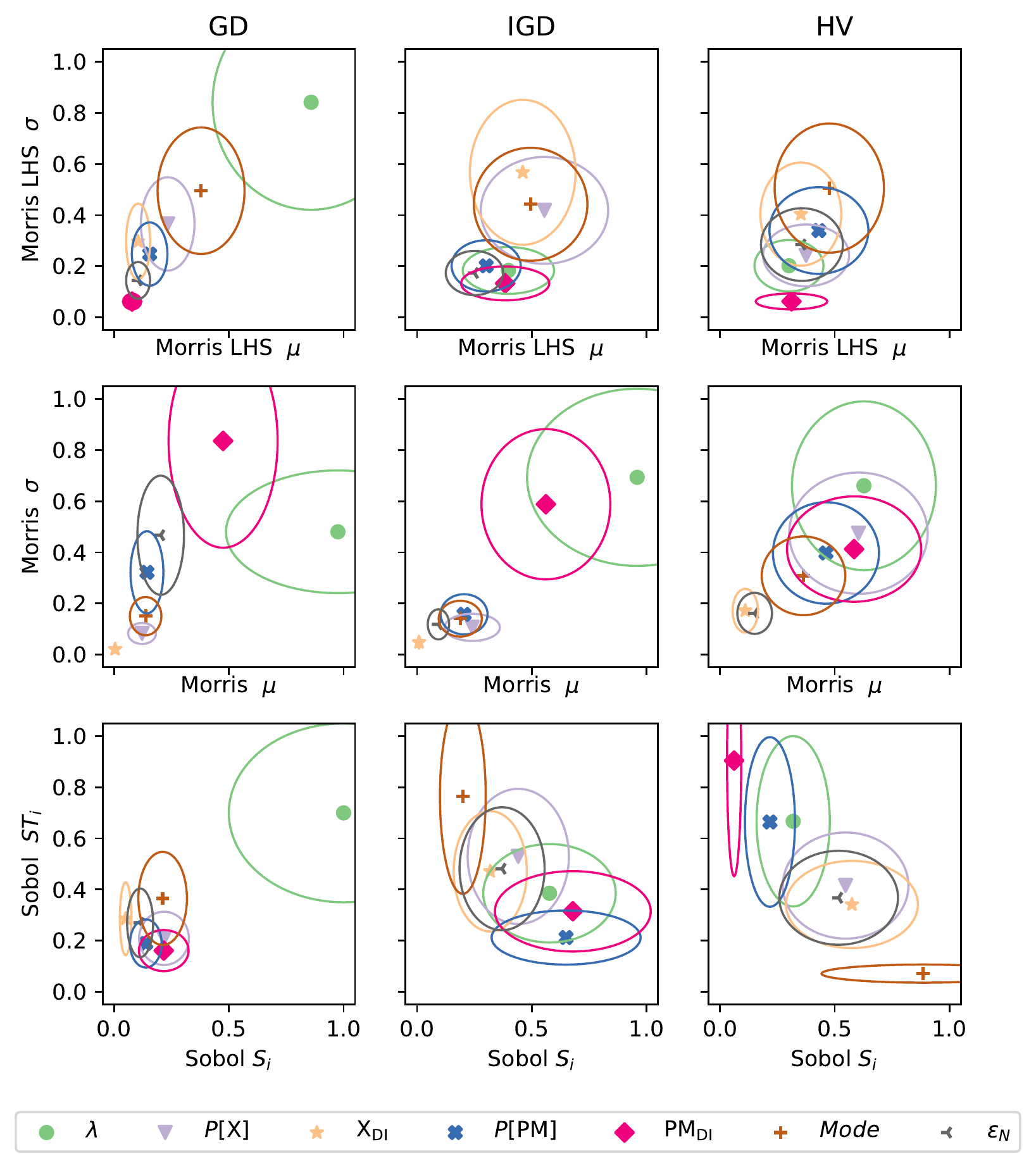}
	\caption{MOEA/D hyperparameters sensitivity analysis. Columns 1, 2, and 3 respectively indicate performance metrics GD, IGD, and HV. Rows 1, 2, and 3, respectively, indicate Morris LHS, Morris, and Sobol methods. Legends of hyperparameter are shown at the bottom. Each hyperparameter is represented by a symbol and a color. An eclipse centered at a hyperparameter is the standard deviation of its influence and direction of its influence. Further apart a hyperparameter in the diagonal direction from the origin $(0,0)$ is, the higher its importance to the algorithm. A larger width of the eclipse of a hyperparameter in the x-axis direction means more variation in the direct influence of a hyperparameter, and a larger height in the y-axis direction means variation in total (or interaction) influence. 
	{\color{insert}Supporting statistical tests and clustering analysis are provided in supplementary Sections A and B.}
		\label{fig:MOEAD_scatter}}
\end{figure}

\begin{figure}
	\centering
	\includegraphics[width=\linewidth]{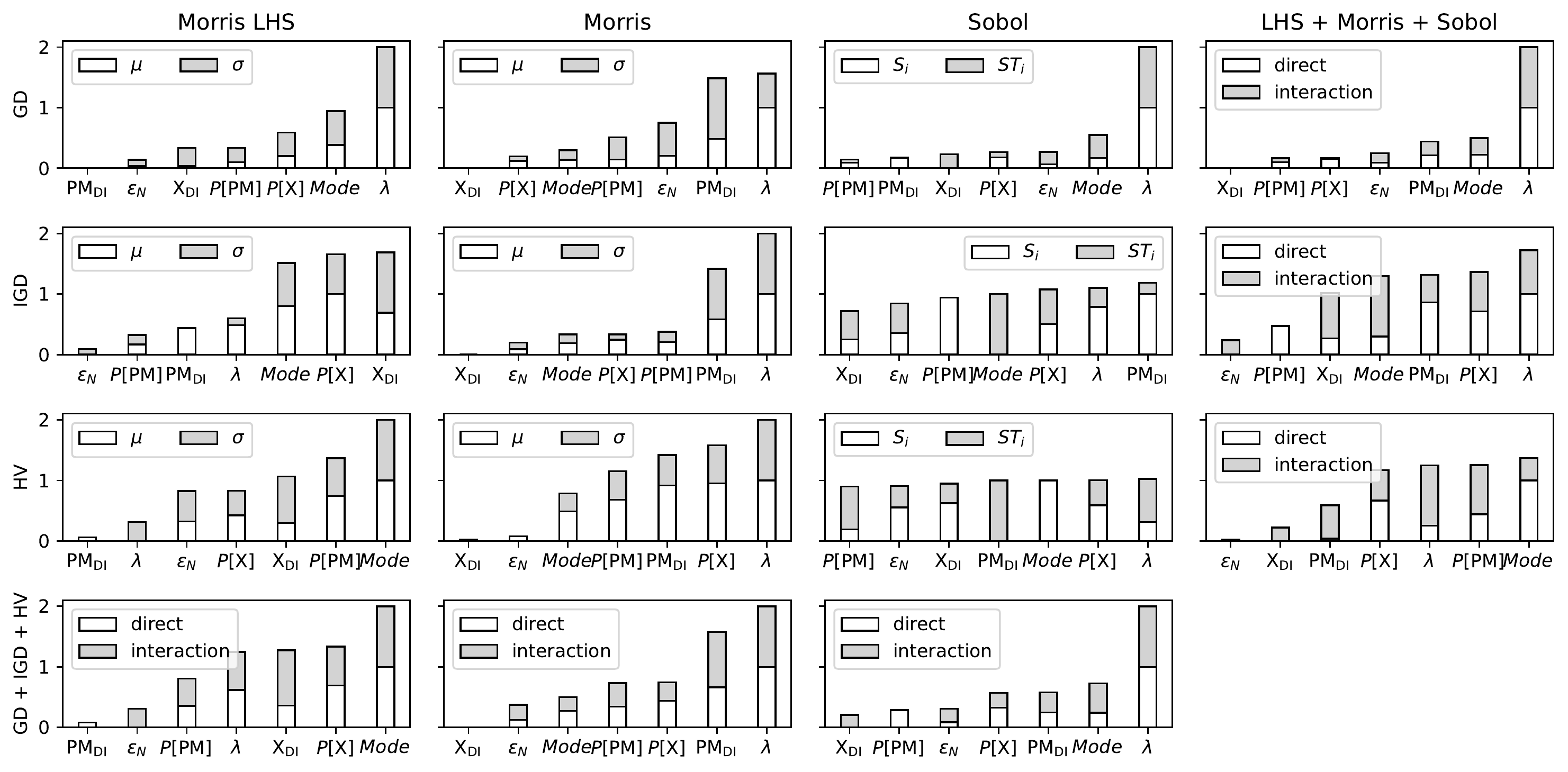}
	\caption{MOEA/D algorithm's hyperparameters performance across all problems (functions). Rows 1, 2, and 3, respectively, show performance evaluated using GD, IGD, and HV metrics. Columns 1, 2, and 3 respectively indicate metric Morris LHS, Morris, and Sobol methods. The white color portion of a bar is direct influence normalized value in \([0,1]\) and gray color portion is interaction (total) influence value in \([0,1]\). A larger height bar implies a higher influence, and hyperparameters in each subplot are arranged from low to high influence.
		\label{fig:MOEAD_bar}}
\end{figure}

\begin{figure}
	\centering
	\includegraphics[width=\linewidth]{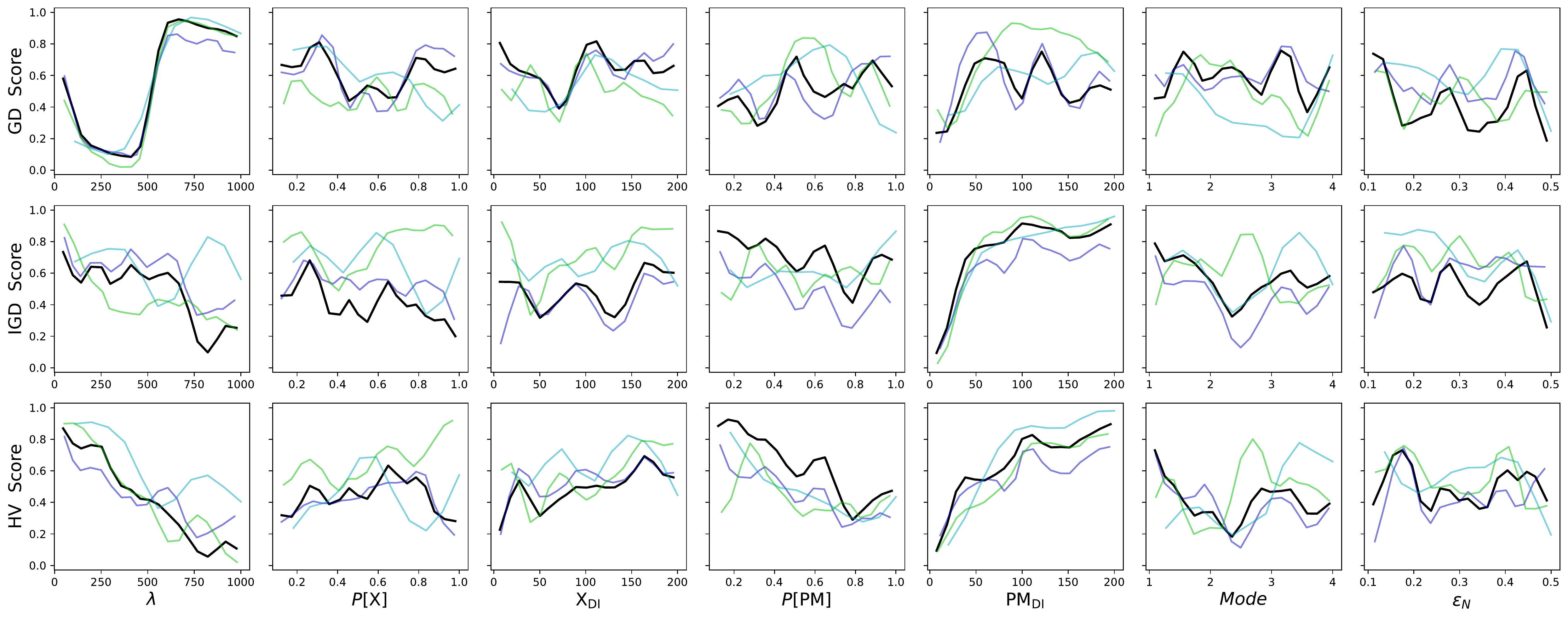}
	\caption{MOEA/D algorithm average performance on 30 runs of each set of hyperparameters. MOEA/D hyperparameter (x-axis) against the mean metric value (y-axis). Rows 1, 2, and 3, respectively, are GD, IGD, and HV metrics. The scores are normalized between 0 and 1 and smooth out using a Gaussian 1D filter with sigma 0.99. The y-axis is GD, IGD, and HV metrics values normalized between a score of 0 and 1, where 0 is the best score for GD and IGD, and 1 is the best score for HV. A total {\color{correct} of 590 samples were evaluated for the MOEA/D algorithm} jointly by  Morris LHS (blue lines), Morris (cyan lines), Sobol (green lines) methods. The hyperparameter values are arranged in 20 bins (lower values to higher values) across the x-axis. Each line in each plot connects the mean values of 20 bins of such samples.
		\label{fig:MOEAD_line}}
\end{figure}

\textbf{Population size $\lambda$.} MOEA/D results are shown in Figs.~\ref{fig:MOEAD_scatter}, \ref{fig:MOEAD_bar}, and \ref{fig:MOEAD_line}. In Fig.~\ref{fig:MOEAD_scatter}, the results of three sensitivity analysis methods for three metrics of MOEA/D performance are presented. Unlike NSGA-III results, population size $\lambda$ is not a clear most significant hyperparameter for MOEA/D multi-objective algorithm. Rather, MOEA/D's hyperparameters $Mode$, the MOO decomposition method, is also among the influential hyperparameters. {\color{correct}Morris LHS method shows that the $Mode$ is the most significant hyperparameter overall on three metrics.} Fig.~\ref{fig:MOEAD_line} also confirms this fact as for the population size values, the GD, IGD, and HV metrics show a strong relation. 

{\color{revise} 
For example, the HV metric in Fig.~\ref{fig:MOEAD_line} shows a linear trend, but it has clear fluctuations in scores. This is because population size has high interaction with other hyperparameters, and tuning population size alone cannot compensate for the role of the other hyperparameters in the performance of MOEA/D on the GD metric. However, for the IGD metric, population size improves the performance of MOEA/D. This shows a highly fluctuating behavior of population size in MOEA/D for varied metrics, i.e., MOEA/D performance has a nonlinear relationship with the population size. This means population size is rather highly involved with interaction with other hyperparameters as the variation in other hyperparameters also influences the performance of MOEA/D.}   

\textbf{MOO decomposition type $Mode$.} The next set of hyperparameters that we observe as highly influential is $Mode$, as it shows high interaction and high overall influence in Morris LHS, Morris, and Sobol for GD and HV metrics. HV metric for Sobol placed the hyperparameters on the direct influence to high total influence diagonal (see Fig.~\ref{fig:MOEAD_scatter}), which suggests that the hyperparameters either have a good  high interaction or good  overall influence. Hence, the sum of these,  presented in Fig.~\ref{fig:MOEAD_bar}, differs only marginally. Sobol rank $Mode$ {\color{correct} is second in the} GD metric as both high interaction and high overall influence and third in the HV metric as it has a high direct influence. 

Examining the performance of $Mode$ in Fig.~\ref{fig:MOEAD_line}, we confirm that the type of MOO decomposition ``Tchebycheff with normalization'' had the best performance, followed by ``penalty based boundary intersection (PBI)'' and  ``Tchebycheff'' has significantly poor performance and ``modified Tchebycheff,'' decomposition mode had the worse scores among MOO decomposition methods. MOEA/D hyperparameter $\epsilon_N$ refers to the number of neighbors for selecting the percentage of the population for sub-problems selection MOEA/D has an equivalent influence as the probability of mutation distribution index $\text{PM}_{\text{DI}}$. However,  $\epsilon_N$ value less than $0.2$ show a sharp improvement in MOEA/D performance.

\textbf{Crossover and mutation hyperparameters.} Genetic operator related hyperparameters $ P[\mathrm{X}] $, $X_{\text{DI}}$, $P[\text{PM}]$ and $\text{PM}_{\text{DI}}$ show varied significance on different metrics on different sensitivity methods. For example, the  probability of mutation distribution index $\text{PM}_{\text{DI}}$ has a high influence on HV metrics (pink diamond and eclipse in Fig.~\ref{fig:MOEAD_scatter}) and a high total influence on HV metrics in the Sobol method. The probability of mutation $P[\text{PM}]$ is second to $\text{PM}_{\text{DI}}$ in total influence on HV as per the Sobol method. This suggests that \textit{mutation has a high influence in diversifying the population} in MOEA/D, helping it produces a better Pareto-front. We also observe that $P[\text{PM}]$ and $\text{PM}_{\text{DI}}$ have mirror image  like performance (see Fig.~\ref{fig:MOEAD_line}), which suggests that values of $P[\text{PM}]$ around $0.8$ and higher values of $\text{PM}_{\text{DI}}$ are more effective in MOEA/D performance. {\color{revise}The probability of crossover $ P[\mathrm{X}] $ has competing performance in the MOEA/D, and it is similar to performances of mutation related hyperparameters. That is, unlike NSGA-III, the probability of crossover does not outshine the crossover and mutation related hyperparameters.}

\textbf{MOEA/D hyperparameters ranking.} In summary, the \textit{ranking of hyperparameters} of MOEA/D from the most influential to least influential hyperparameters is $\lambda$, $Mode$,  $\text{PM}_{\text{DI}}$,  $P[\text{PM}]$,  $ P[\mathrm{X}] $, $\epsilon_N$, and $X_{\text{DI}}$.

{\color{insert}
\subsubsection{Remarks on MOO hyperparameter rankings and algorithms}
Providing ranking to hyperparameters for MOO is more challenging than SOO since it uses three distinct sensitivity analysis methods and uses three distinct performance metrics. However, we look for potential agreement between these distinct measures. We observe that the population size $\lambda$ clearly emerged as the most influential hyperparameter in all three analyses and metrics for NSGA-III, and the probability of crossover was the second most influential. These two hyperparameters significantly dominate all other hyperparameters in NSGA-III. Whereas for MOEA/D, the population size $\lambda$ dominates only for the GD metric and for Morris analysis. For HV and IGD metric and Morris LHS and Sobol analysis, $Mode$ and mutation probability are dominant factors. Unlike NSGA-III, there is no clear, significantly dominant hyperparameter in MOEA/D. Therefore, considering hyperparameters' strong variability and dependency on the type of hyperparameter sampling methods and type of performance metrics, we may confirm that NSGA-III is a more stable algorithm than MOEA/D.   

}

\section{Conclusions}
\label{sec:conclusions}
We present a framework for systematic and methodological analysis of the effectiveness of the evolutionary algorithm hyperparameters. This analysis results in (i) identifying the pattern of influence each hyperparameter has on the algorithm, (ii) recommending rankings of hyperparameter influence, and (iii) analyzing the stability of algorithms related to hyperparameter sampling and performance metrics. We apply our methodology to state-of-the-art evolutionary algorithms: two single-objective algorithms and two multi-objective algorithms. The single-objective algorithms used are covariance matrix adaptation evolutionary strategy (CMA-ES), differential evolution (DE), and multi-objective algorithms used are non-dominated sorting genetic algorithm III (NSGA-III), and multi-objective evolutionary algorithm based on decomposition (MOEA/D). Our methodology involves two global sensitivity analysis methods, Morris and Sobol. This methodology is computationally heavy, but it produces widely usable and effective recommendations on hyperparameters ranking, being the order in which one can tune EA hyperparameters to achieve high performance. For example, the initial step size, base vector selection type (mutation), probability of crossover, and mode multi-objective problem decomposition were among the most influential hyperparameters of CMA-ES, DE, NSGA-III, and MOEA/D algorithms, respectively. The results show how the hyperparameters interact with one another when they are sampled differently, and different performance measures are used. {\color{blue} This framework can further analyze the sensitivity and influence of adaptive and dynamically tuneable hyperparameters for future work. Furthermore, since different hyperparameters sampling methods showed varied ranking, this work can further study the influence of the sampling method or sensitivity of an algorithm or its hyperparameters towards a particular type of sampling.}  

\small
\bibliographystyle{agsm}
\bibliography{0SAofEA}

@article{eggensperger2019pitfalls,
  title={Pitfalls and best practices in algorithm configuration},
  author={Eggensperger, Katharina and Lindauer, Marius and Hutter, Frank},
  journal={Journal of Artificial Intelligence Research},
  volume={64},
  pages={861--893},
  year={2019}
}

@article{cheng2021differential,
  title={Differential evolution algorithm with fitness and diversity ranking-based mutation operator},
  author={Cheng, Jianchao and Pan, Zhibin and Liang, Hao and Gao, Zhaoqi and Gao, Jinghuai},
  journal={Swarm and Evolutionary Computation},
  volume={61},
  pages={100816},
  year={2021},
  publisher={Elsevier}
}

@article{yazdani2021survey,
  title={A survey of evolutionary continuous dynamic optimization over two decades—Part B},
  author={Yazdani, Danial and Cheng, Ran and Yazdani, Donya and Branke, J{\"u}rgen and Jin, Yaochu and Yao, Xin},
  journal={IEEE Transactions on Evolutionary Computation},
  volume={25},
  number={4},
  pages={630--650},
  year={2021},
  publisher={IEEE}
}

@article{yuan2022constrained,
  title={A constrained multi-objective evolutionary algorithm using valuable infeasible solutions},
  author={Yuan, Jiawei and Liu, Hai-Lin and He, Zhaoshui},
  journal={Swarm and Evolutionary Computation},
  volume={68},
  pages={101020},
  year={2022},
  publisher={Elsevier}
}

@article{han2022surrogate,
  title={A surrogate-assisted evolutionary algorithm for expensive many-objective optimization in the refining process},
  author={Han, Dong and Du, Wenli and Wang, Xinjie and Du, Wei},
  journal={Swarm and Evolutionary Computation},
  volume={69},
  pages={100988},
  year={2022},
  publisher={Elsevier}
}

@article{rivera2022preference,
  title={Preference incorporation into many-objective optimization: An Ant colony algorithm based on interval outranking},
  author={Rivera, Gilberto and Coello, Carlos A Coello and Cruz-Reyes, Laura and Fernandez, Eduardo R and Gomez-Santillan, Claudia and Rangel-Valdez, Nelson},
  journal={Swarm and Evolutionary Computation},
  volume={69},
  pages={101024},
  year={2022},
  publisher={Elsevier}
}

@article{liang2021clustering,
  title={A clustering-based differential evolution algorithm for solving multimodal multi-objective optimization problems},
  author={Liang, Jing and Qiao, Kangjia and Yue, Caitong and Yu, Kunjie and Qu, Boyang and Xu, Ruohao and Li, Zhimeng and Hu, Yi},
  journal={Swarm and Evolutionary Computation},
  volume={60},
  pages={100788},
  year={2021},
  publisher={Elsevier}
}

@article{viktorin2019distance,
  title={Distance based parameter adaptation for success-history based differential evolution},
  author={Viktorin, Adam and Senkerik, Roman and Pluhacek, Michal and Kadavy, Tomas and Zamuda, Ales},
  journal={Swarm and Evolutionary Computation},
  volume={50},
  pages={100462},
  year={2019},
  publisher={Elsevier}
}

@article{piotrowski2018step,
  title={Step-by-step improvement of JADE and SHADE-based algorithms: Success or failure?},
  author={Piotrowski, Adam P and Napiorkowski, Jaroslaw J},
  journal={Swarm and evolutionary computation},
  volume={43},
  pages={88--108},
  year={2018},
  publisher={Elsevier}
}

@article{feurer2020auto,
	title        = {Auto-sklearn 2.0: Hands-free {AutoML} via meta-learning},
	author       = {Feurer, Matthias and Eggensperger, Katharina and Falkner, Stefan and Lindauer, Marius and Hutter, Frank},
	year         = 2020,
	journal      = {arXiv:2007.04074}
}

@inproceedings{bezerra2015comparing,
	title        = {Comparing decomposition-based and automatically component-wise designed multi-objective evolutionary algorithms},
	author       = {Bezerra, Leonardo CT and L{\'o}pez-Ib{\'a}nez, Manuel and St{\"u}tzle, Thomas},
	year         = 2015,
	booktitle    = {International Conference on Evolutionary Multi-Criterion Optimization},
	pages        = {396--410},
	organization = {Springer}
}

@incollection{feurer2019hyperparameter,
	title        = {Hyperparameter optimization},
	author       = {Feurer, Matthias and Hutter, Frank},
	year         = 2019,
	booktitle    = {Automated Machine Learning},
	publisher    = {Springer, Cham},
	pages        = {3--33}
}

@inproceedings{thornton2013auto,
	title        = {{Auto-WEKA:} Combined selection and hyperparameter optimization of classification algorithms},
	author       = {Thornton, Chris and Hutter, Frank and Hoos, Holger H and Leyton-Brown, Kevin},
	year         = 2013,
	booktitle    = {Proceedings of the 19th ACM SIGKDD International Conference on Knowledge Discovery and Data Mining},
	pages        = {847--855}
}

@article{liang2006performance,
	title        = {Performance evaluation of multiagent genetic algorithm},
	author       = {Liang, Jing J and Baskar, S and Suganthan, Ponnuthurai N and Qin, A Kai},
	year         = 2006,
	journal      = {Natural Computing},
	publisher    = {Springer},
	volume       = 5,
	number       = 1,
	pages        = {83--96}
}

@inproceedings{iommazzo2019algorithmic,
	title        = {Algorithmic configuration by learning and optimization},
	author       = {Iommazzo, Gabriele and d'Ambrosio, Claudia and Frangioni, Antonio and Liberti, Leo},
	year         = 2019,
	booktitle    = {Cologne-Twente Workshop on Graphs and Combinatorial Optimization}
}

@article{he2021automl,
	title        = {{AutoML}: A survey of the state-of-the-art},
	author       = {He, Xin and Zhao, Kaiyong and Chu, Xiaowen},
	year         = 2021,
	journal      = {Knowledge-Based Systems},
	publisher    = {Elsevier},
	volume       = 212,
	pages        = 106622
}

@article{deb2014analysing,
	title        = {Analysing mutation schemes for real-parameter genetic algorithms},
	author       = {Deb, Kalyanmoy and Deb, Debayan},
	year         = 2014,
	journal      = {International Journal of Artificial Intelligence and Soft Computing},
	publisher    = {Inderscience Publishers Ltd},
	volume       = 4,
	number       = 1,
	pages        = {1--28}
}

@article{cui2019improved,
	title        = {Improved {NSGA-III} with selection-and-elimination operator},
	author       = {Cui, Zhihua and Chang, Yu and Zhang, Jiangjiang and Cai, Xingjuan and Zhang, Wensheng},
	year         = 2019,
	journal      = {Swarm and Evolutionary Computation},
	publisher    = {Elsevier},
	volume       = 49,
	pages        = {23--33}
}

@inproceedings{voss2010improved,
	title        = {Improved step size adaptation for the {MO-CMA-ES}},
	author       = {Vo{\ss}, Thomas and Hansen, Nikolaus and Igel, Christian},
	year         = 2010,
	booktitle    = {Proceedings of the 12th Annual Conference on Genetic and Evolutionary Computation},
	publisher    = {ACM},
	pages        = {487--494}
}

@article{wang2015adaptive,
	title        = {Adaptive replacement strategies for {MOEA/D}},
	author       = {Wang, Zhenkun and Zhang, Qingfu and Zhou, Aimin and Gong, Maoguo and Jiao, Licheng},
	year         = 2015,
	journal      = {IEEE Transactions on Cybernetics},
	publisher    = {IEEE},
	volume       = 46,
	number       = 2,
	pages        = {474--486}
}

@article{qi2014moea,
	title        = {{MOEA/D} with adaptive weight adjustment},
	author       = {Qi, Yutao and Ma, Xiaoliang and Liu, Fang and Jiao, Licheng and Sun, Jianyong and Wu, Jianshe},
	year         = 2014,
	journal      = {Evolutionary Computation},
	publisher    = {MIT Press},
	volume       = 22,
	number       = 2,
	pages        = {231--264}
}

@inproceedings{zhang2010moea,
	title        = {{MOEA/D} with {NBI-style Tchebycheff} approach for portfolio management},
	author       = {Zhang, Qingfu and Li, Hui and Maringer, Dietmar and Tsang, Edward},
	year         = 2010,
	booktitle    = {IEEE Congress on Evolutionary Computation},
	pages        = {1--8},
	organization = {IEEE}
}

@article{das2010differential,
	title        = {Differential evolution: A survey of the state-of-the-art},
	author       = {Das, Swagatam and Suganthan, Ponnuthurai Nagaratnam},
	year         = 2010,
	journal      = {IEEE Transactions on Evolutionary Computation},
	publisher    = {IEEE},
	volume       = 15,
	number       = 1,
	pages        = {4--31}
}

@inproceedings{das2005two,
	title        = {Two improved differential evolution schemes for faster global search},
	author       = {Das, Swagatam and Konar, Amit and Chakraborty, Uday K},
	year         = 2005,
	booktitle    = {Proceedings of the 7th Annual Conference on Genetic and Evolutionary Computation},
	pages        = {991--998}
}

@article{biswas2009design,
	title        = {Design of fractional-order {PI$\lambda$D$\mu$} controllers with an improved differential evolution},
	author       = {Biswas, Arijit and Das, Swagatam and Abraham, Ajith and Dasgupta, Sambarta},
	year         = 2009,
	journal      = {Engineering Applications of Artificial Intelligence},
	publisher    = {Elsevier},
	volume       = 22,
	number       = 2,
	pages        = {343--350}
}

@article{das2009differential,
	title        = {Differential evolution using a neighborhood-based mutation operator},
	author       = {Das, Swagatam and Abraham, Ajith and Chakraborty, Uday K and Konar, Amit},
	year         = 2009,
	journal      = {IEEE Transactions on Evolutionary Computation},
	publisher    = {IEEE},
	volume       = 13,
	number       = 3,
	pages        = {526--553}
}

@article{das2007automatic,
	title        = {Automatic clustering using an improved differential evolution algorithm},
	author       = {Das, Swagatam and Abraham, Ajith and Konar, Amit},
	year         = 2007,
	journal      = {IEEE Transactions on Systems, Man, and Cybernetics-Part A: Systems and Humans},
	publisher    = {IEEE},
	volume       = 38,
	number       = 1,
	pages        = {218--237}
}

@article{islam2011adaptive,
	title        = {An adaptive differential evolution algorithm with novel mutation and crossover strategies for global numerical optimization},
	author       = {Islam, Sk Minhazul and Das, Swagatam and Ghosh, Saurav and Roy, Subhrajit and Suganthan, Ponnuthurai Nagaratnam},
	year         = 2011,
	journal      = {IEEE Transactions on Systems, Man, and Cybernetics, Part B (Cybernetics)},
	publisher    = {IEEE},
	volume       = 42,
	number       = 2,
	pages        = {482--500}
}

@article{deb1995simulated,
	title        = {Simulated binary crossover for continuous search space},
	author       = {Deb, Kalyanmoy and Agrawal, Ram Bhushan and others},
	year         = 1995,
	journal      = {Complex Systems},
	publisher    = {Citeseer},
	volume       = 9,
	number       = 2,
	pages        = {115--148}
}

@article{de2007parameter,
	title        = {Parameter setting in {EAs}: a 30 year perspective},
	author       = {De Jong, Kenneth},
	year         = 2007,
	journal      = {Studies in Computational Intelligence (SCI)},
	publisher    = {Springer},
	volume       = 54,
	pages        = {1--18}
}

@article{hansen2001completely,
	title        = {Completely derandomized self-adaptation in evolution strategies},
	author       = {Hansen, Nikolaus and Ostermeier, Andreas},
	year         = 2001,
	journal      = {Evolutionary Computation},
	publisher    = {MIT Press},
	volume       = 9,
	number       = 2,
	pages        = {159--195}
}

@inproceedings{de2016evolutionary,
  title={Evolutionary computation: a unified approach},
  author={De Jong, Kenneth},
  booktitle={Proceedings of the 2016 on Genetic and Evolutionary Computation Conference Companion},
  pages={185--199},
  year={2016}
}

@article{de2004choice,
	title        = {On the choice of the offspring population size in evolutionary algorithms},
	author       = {Jansen, Thomas and Jong, Kenneth A De and Wegener, Ingo},
	year         = 2005,
	journal      = {Evolutionary Computation},
	publisher    = {MIT Press},
	volume       = 13,
	number       = 4,
	pages        = {413--440}
}

@techreport{liang2013problem,
	title        = {Problem definitions and evaluation criteria for the {CEC 2014} special session and competition on single objective real-parameter numerical optimization},
	author       = {Liang, Jing J and Qu, Bo Y and Suganthan, Ponnuthurai N},
	year         = 2013,
	institution  = {Zhengzhou University, Zhengzhou China and Nanyang Technological University, Singapore}
}

@techreport{liang2014problem,
	title        = {Problem definitions and evaluation criteria for the {CEC 2015} competition on learning-based real-parameter single objective optimization},
	author       = {Liang, JJ and Qu, BY and Suganthan, PN and Chen, Q},
	year         = 2014,
	institution  = {Zhengzhou University, Zhengzhou China and Nanyang Technological University, Singapore}
}

@book{saltelli2004sensitivity,
	title        = {Sensitivity analysis in practice: A guide to assessing scientific models},
	author       = {Saltelli, Andrea and Tarantola, Stefano and Campolongo, Francesca and Ratto, Marco},
	year         = 2004,
	publisher    = {John Wiley \& Sons},
	volume       = 1
}

@inproceedings{lima2004parameter,
	title        = {Parameter-less optimization with the extended compact genetic algorithm and iterated local search},
	author       = {Lima, Cl{\'a}udio F and Lobo, Fernando G},
	year         = 2004,
	booktitle    = {Genetic and Evolutionary Computation Conference},
	pages        = {1328--1339},
	organization = {Springer}
}

@article{kramer2010evolutionary,
	title        = {Evolutionary self-adaptation: a survey of operators and strategy parameters},
	author       = {Kramer, Oliver},
	year         = 2010,
	journal      = {Evolutionary Intelligence},
	publisher    = {Springer},
	volume       = 3,
	number       = 2,
	pages        = {51--65}
}

@incollection{eiben2007parameter,
	title        = {Parameter control in evolutionary algorithms},
	author       = {Eiben, Agoston E and Michalewicz, Zbigniew and Schoenauer, Marc and Smith, James E},
	year         = 2007,
	booktitle    = {Parameter setting in evolutionary algorithms},
	publisher    = {Springer},
	pages        = {19--46}
}

@article{maturana2010autonomous,
	title        = {Autonomous operator management for evolutionary algorithms},
	author       = {Maturana, Jorge and Lardeux, Fr{\'e}d{\'e}ric and Saubion, Fr{\'e}d{\'e}ric},
	year         = 2010,
	journal      = {Journal of Heuristics},
	publisher    = {Springer},
	volume       = 16,
	number       = 6,
	pages        = {881--909}
}

@inproceedings{crossley2013quantifying,
	title        = {Quantifying the impact of parameter tuning on nature-inspired algorithms},
	author       = {Crossley, Matthew and Nisbet, Andy and Amos, Martyn},
	year         = 2013,
	booktitle    = {The 12th European Conference on Artificial Life},
	pages        = {925--932},
	organization = {MIT Press}
}

@article{Bergstra,
	title        = {Random Search for Hyper-Parameter Optimization},
	author       = {Bergstra, James and Bengio, Yoshua},
	year         = 2012,
	journal      = {Journal of Machine Learning Research},
	volume       = 13,
	pages        = {281–305},
	issn         = {1532-4435}
}

@book{miettinen2012nonlinear,
	title        = {Nonlinear multiobjective optimization},
	author       = {Miettinen, Kaisa},
	year         = 2012,
	publisher    = {Springer},
	volume       = 12
}

@inproceedings{fonseca2006improved,
	title        = {An improved dimension-sweep algorithm for the hypervolume indicator},
	author       = {Fonseca, Carlos M and Paquete, Lu{\'\i}s and L{\'o}pez-Ib{\'a}nez, Manuel},
	year         = 2006,
	booktitle    = {IEEE International Conference on Evolutionary Computation},
	pages        = {1157--1163},
	organization = {IEEE}
}

@inproceedings{SensitivityTuning,
	title        = {Automatic Tuning of Algorithms Through Sensitivity Minimization},
	author       = {Conca, Piero and Stracquadanio, Giovanni and Nicosia, Giuseppe},
	year         = 2015,
	booktitle    = {Machine Learning, Optimization, and Big Data},
	publisher    = {Springer},
	pages        = {14--25},
	isbn         = {978-3-319-27926-8},
	editor       = {Pardalos, Panos and Pavone, Mario and Farinella, Giovanni Maria and Cutello, Vincenzo}
}

@article{Simplifying,
	title        = {Simplifying {Sirius}: sensitivity analysis and development of a meta-model for wheat yield prediction},
	author       = {Brooks, Roger and Semenov, Mikhail and Jamieson, Peter},
	year         = 2001,
	month        = {01},
	journal      = {European Journal of Agronomy},
	volume       = 14,
	pages        = {43--60},
	doi          = {10.1016/S1161-0301(00)00089-7}
}

@article{SAUse,
	title        = {Practical use of computationally frugal model analysis methods},
	author       = {Hill, Mary C and Kavetski, Dmitri and Clark, Martyn and Ye, Ming and Arabi, Mazdak and Lu, Dan and Foglia, Laura and Mehl, Steffen},
	year         = 2016,
	journal      = {Groundwater},
	publisher    = {Wiley},
	volume       = 54,
	number       = 2,
	pages        = {159--170}
}

@inproceedings{TuningSensitivity,
	title        = {Evolutionary Algorithm Parameter Tuning with Sensitivity Analysis},
	author       = {Pinel, Fr{\'e}d{\'e}ric and Danoy, Gr{\'e}goire and Bouvry, Pascal},
	year         = 2012,
	booktitle    = {Security and Intelligent Information Systems},
	publisher    = {Springer},
	pages        = {204--216},
	editor       = {Bouvry, Pascal and K{\l}opotek, Mieczys{\l}aw A. and Lepr{\'e}vost, Franck and Marciniak, Ma{\l}gorzata and Mykowiecka, Agnieszka and Rybi{\'{n}}ski, Henryk}
}

@article{iglesias2007study,
	title        = {Study of sensitivity of the parameters of a genetic algorithm for design of water distribution networks},
	author       = {Iglesias, Pedro L and Mora, Daniel and Martinez, F Javier and Fuertes, Vicente S},
	year         = 2007,
	journal      = {Journal of Urban and Environmental Engineering},
	publisher    = {JSTOR},
	volume       = 1,
	number       = 2,
	pages        = {61--69}
}

@inproceedings{AltroArticolo,
	title        = {Assessing Algorithm Parameter Importance Using Global Sensitivity Analysis},
	author       = {Greco, Alessio and Riccio, Salvatore Danilo and Timmis, Jon and Nicosia, Giuseppe},
	year         = 2019,
	booktitle    = {Analysis of Experimental Algorithms},
	publisher    = {Springer},
	pages        = {392--407},
	doi          = {10.1007/978-3-030-34029-2\_26},
	editor       = {Kotsireas, Ilias and Pardalos, Panos and Parsopoulos, Konstantinos E. and Souravlias, Dimitris and Tsokas, Arsenis}
}

@article{zhang2007moea,
	title        = {{MOEA/D}: A multiobjective evolutionary algorithm based on decomposition},
	author       = {Zhang, Qingfu and Li, Hui},
	year         = 2007,
	journal      = {IEEE Transactions on Evolutionary Computation},
	publisher    = {IEEE},
	volume       = 11,
	number       = 6,
	pages        = {712--731}
}

@inproceedings{paul2011sensitivity,
	title        = {Sensitivity analysis from evolutionary algorithm search paths},
	author       = {Paul, Gr{\'e}gory and M{\"u}ller, Christian L and Sbalzarini, Ivo F},
	year         = 2011,
	booktitle    = {EVOLVE - A bridge between Probability, Set Oriented Numerics and Evolutionary Computation},
	publisher    = {Springer},
	series       = {Studies in Computational Intelligence}
}

@article{lopez2016irace,
	title        = {The irace package: Iterated racing for automatic algorithm configuration},
	author       = {L{\'o}pez-Ib{\'a}{\~n}ez, Manuel and Dubois-Lacoste, J{\'e}r{\'e}mie and C{\'a}ceres, Leslie P{\'e}rez and Birattari, Mauro and St{\"u}tzle, Thomas},
	year         = 2016,
	journal      = {Operations Research Perspectives},
	publisher    = {Elsevier},
	volume       = 3,
	pages        = {43--58}
}

@article{saltelli1999quantitative,
	title        = {A quantitative model-independent method for global sensitivity analysis of model output},
	author       = {Saltelli, Andrea and Tarantola, Stefano and Chan, KP-S},
	year         = 1999,
	journal      = {Technometrics},
	publisher    = {Taylor \& Francis},
	volume       = 41,
	number       = 1,
	pages        = {39--56}
}

@article{safe,
	title        = {An effective screening design for sensitivity analysis of large models},
	author       = {Pianosi, Francesca and Sarrazin, Fanny and Wagener, Thorsten},
	year         = 2015,
	journal      = {Environmental Modelling \& Software},
	volume       = 70,
	pages        = {80--85}
}

@article{PlatEMO,
	title        = {{PlatEMO}: A {MATLAB} platform for evolutionary multi-objective optimization},
	author       = {Tian, Ye and Cheng, Ran and Zhang, Xingyi and Jin, Yaochu},
	year         = 2017,
	journal      = {IEEE Computational Intelligence Magazine},
	volume       = 12,
	number       = 4,
	pages        = {73--87}
}

@misc{ypea,
	title        = {{YPEA}: Yarpiz Evolutionary Algorithms},
	author       = {S.M.K. Heris},
	year         = 2019,
	publisher    = {YarPiz},
	note         = {\url{https://github.com/smkalami/ypea}. Accessed on 22 September 2021}
}

@article{lou2021non,
  title={Non-revisiting stochastic search revisited: Results, perspectives, and future directions},
  author={Lou, Yang and Yuen, Shiu Yin and Chen, Guanrong},
  journal={Swarm and Evolutionary Computation},
  volume={61},
  pages={100828},
  year={2021},
  publisher={Elsevier}
}

@inproceedings{ojha2014aco,
  title={{ACO} for continuous function optimization: A performance analysis},
  author={Ojha, Varun Kumar and Abraham, Ajith and Sn{\'a}{\v{s}}el, V{\'a}clav},
  booktitle={2014 14th International Conference on Intelligent Systems Design and Applications},
  pages={145--150},
  year={2014},
  organization={IEEE}
}

@inproceedings{taylor2021sensitivity,
  title={Sensitivity analysis for deep learning: ranking hyper-parameter influence},
  author={Taylor, Rhian and Ojha, Varun and Martino, Ivan and Nicosia, Giuseppe},
  booktitle={2021 IEEE 33rd International Conference on Tools with Artificial Intelligence (ICTAI)},
  pages={512--516},
  year={2021},
  organization={IEEE}
}

@inproceedings{ojha2014simultaneous,
  title={Simultaneous optimization of neural network weights and active nodes using metaheuristics},
  author={Ojha, Varun Kumar and Abraham, Ajith and Sn{\'a}{\v{s}}el, V{\'a}clav},
  booktitle={2014 14th International Conference on Hybrid Intelligent Systems},
  pages={248--253},
  year={2014},
  organization={IEEE}
}

@misc{ojha2022SaEAs,
	title        = {Sensitivity Analysis Evolutionary Algorithms},
	author       = {Ojha, Varun and Timmis, Jon and Nicosia, Giuseppe},
	year         = 2022,
	note         = {\url{https://github.com/vojha-code/SAofEAs}. Accessed on 10 February 2022}
}

@article{saltelli2002sensitivity,
	title        = {Sensitivity analysis for importance assessment},
	author       = {Saltelli, Andrea},
	year         = 2002,
	journal      = {Risk Analysis},
	publisher    = {John Wiley \& Sons},
	volume       = 22,
	number       = 3,
	pages        = {579--590}
}

@book{saltelli2008global,
	title        = {Global Sensitivity Analysis: {T}he primer},
	author       = {Saltelli, Andrea and Ratto, Marco and Andres, Terry and Campolongo, Francesca and Cariboni, Jessica and Gatelli, Debora and Saisana, Michaela and Tarantola, Stefano},
	year         = 2008,
	publisher    = {John Wiley \& Sons}
}

@article{MorrisOriginal,
	title        = {Factorial Sampling Plans for Preliminary Computational Experiments},
	author       = {Max D. Morris},
	year         = 1991,
	journal      = {Technometrics},
	publisher    = {Taylor & Francis},
	volume       = 33,
	number       = 2,
	pages        = {161--174}
}

@article{Campolongo2007,
	title        = {An effective screening design for sensitivity analysis of large models},
	author       = {Campolongo, Francesca and Saltelli, Andrea and Cariboni, Jessica},
	year         = 2007,
	month        = {01},
	journal      = {Environmental Modelling \& Software},
	volume       = 22,
	pages        = {1509--1518}
}

@article{SOBOL,
	title        = {Global sensitivity indices for nonlinear mathematical models, Review},
	author       = {Sobol, Ilya M. and Kucherenko, Sergei},
	year         = 2005,
	month        = {01},
	journal      = {Wilmott Magazine},
	volume       = 2005,
	pages        = {56--61},
	doi          = {10.1002/wilm.42820050114}
}

@inproceedings{CMAES,
	title        = {Adapting arbitrary normal mutation distributions in evolution strategies: the covariance matrix adaptation},
	author       = {Hansen, Nikolaus and Ostermeier, Andreas},
	year         = 1996,
	booktitle    = {Proceedings of IEEE International Conference on Evolutionary Computation},
	pages        = {312--317}
}

@article{DEOriginal,
	title        = {Differential Evolution - A Simple and Efficient Heuristic for Global Optimization over Continuous Spaces},
	author       = {Storn, Rainer and Price, Kenneth},
	year         = 1997,
	month        = {01},
	journal      = {Journal of Global Optimization},
	volume       = 11,
	pages        = {341--359},
	doi          = {10.1023/A:1008202821328}
}

@article{Tesbench,
	title        = {Evolutionary programming made faster},
	author       = {Xin Yao and  Yong Liu and  Guangming Lin},
	year         = 1999,
	journal      = {IEEE Transactions on Evolutionary Computation},
	volume       = 3,
	number       = 2,
	pages        = {82--102}
}

@article{NSGA_II,
  title={A fast and elitist multiobjective genetic algorithm: {NSGA-II}},
  author={Deb, Kalyanmoy and Pratap, Amrit and Agarwal, Sameer and Meyarivan, TAMT},
  journal={IEEE Transactions on Evolutionary Computation},
  volume={6},
  number={2},
  pages={182--197},
  year={2002},
  publisher={IEEE}
}

@article{NSGA_III,
	title        = {An Evolutionary Many-Objective Optimization Algorithm Using Reference-Point-Based Nondominated Sorting Approach, {Part I}: Solving Problems With Box Constraints},
	author       = {Deb, Kalyanmoy and Jain, Himanshu},
	year         = 2013,
	journal      = {IEEE Transactions on Evolutionary Computation},
	publisher    = {IEEE},
	volume       = 18,
	number       = 4,
	pages        = {577--601}
}

@inproceedings{GDFirstAppearance,
	title        = {Evolutionary Computation and Convergence to a Pareto Front},
	author       = {D. A. Van Veldhuizen and G. B. Lamont},
	year         = 1998,
	booktitle    = {Late Breaking Papers at the 1998 Genetic Programming Conference},
	publisher    = {Stanford University},
	pages        = {221--228}
}

@phdthesis{GDOtherAppearance,
	title        = {Multiobjective Evolutionary Algorithms: Classifications, Analyses, and New Innovations},
	author       = {David A. Van Veldhuizen},
	year         = 1999,
	address      = {Wright-Patterson AFB, Ohio},
	school       = {School of Engineering, Air Force Institute of Technology}
}

@inproceedings{HVFirstAppearance,
	title        = {Multiobjective optimization using evolutionary algorithms—a comparative case study},
	author       = {Zitzler, Eckart and Thiele, Lothar},
	year         = 1998,
	booktitle    = {International Conference on Parallel Problem Solving from Nature},
	publisher    = {Springer},
	pages        = {292--301}
}

@article{zitzler2003performance,
	title        = {Performance assessment of multiobjective optimizers: An analysis and review},
	author       = {Zitzler, Eckart and Thiele, Lothar and Laumanns, Marco and Fonseca, Carlos M and Da Fonseca, Viviane Grunert},
	year         = 2003,
	journal      = {IEEE Transactions on Evolutionary Computation},
	publisher    = {IEEE},
	volume       = 7,
	number       = 2,
	pages        = {117--132}
}

@article{das1998normal,
	title        = {Normal-boundary intersection: A new method for generating the Pareto surface in nonlinear multicriteria optimization problems},
	author       = {Das, Indraneel and Dennis, John E},
	year         = 1998,
	journal      = {SIAM Journal on Optimization},
	publisher    = {SIAM},
	volume       = 8,
	number       = 3,
	pages        = {631--657}
}

@inproceedings{ProblemSuiteDTLZ,
	title        = {Scalable multi-objective optimization test problems},
	author       = {Deb, Kalyanmoy and Thiele, Lothar and Laumanns, Marco and Eckart Zitzler},
	year         = 2002,
	month        = {06},
	booktitle    = {Proceedings of 2002 IEEE Congress on Evolutionary Computation},
	volume       = 1,
	pages        = {825--830},
	doi          = {10.1109/CEC.2002.1007032}
}

@article{WFGToolkit,
	title        = {A review of multiobjective test problems and a scalable test problem toolkit},
	author       = {Simon Huband and Phil Hingston and Luigi Barone and Lydon While},
	year         = 2006,
	journal      = {IEEE Transactions on Evolutionary Computation},
	volume       = 10,
	number       = 5,
	pages        = {477--506}
}

@inproceedings{Sensitivity,
	title        = {Introduction to Sensitivity Analysis},
	author       = {Iooss, Bertrand and Saltelli, Andrea},
	year         = 2016,
	booktitle    = {Handbook of Uncertainty Quantification},
	publisher    = {Springer},
	pages        = {1--20},
	doi          = {10.1007/978-3-319-11259-6\_31-1},
	isbn         = {978-3-319-11259-6},
	editor       = {Ghanem, Roger and Higdon, David and Owhadi, Houman}
}

@inbook{Kalpic2011,
	title        = {Student's t-Tests},
	author       = {Kalpi{\'{c}}, Damir and Hlupi{\'{c}}, Nikica and Lovri{\'{c}}, Miodrag},
	year         = 2011,
	booktitle    = {International Encyclopedia of Statistical Science},
	publisher    = {Springer},
	pages        = {1559--1563},
	editor       = {Lovric, Miodrag}
}

@article{KMeans,
	title        = {Least squares quantization in {PCM}},
	author       = {Stuart P. Lloyd},
	year         = 1982,
	journal      = {IEEE Transactions on Information Theory},
	volume       = 28,
	pages        = {129--137}
}

@article{Silhouette,
	title        = {Silhouettes: A graphical aid to the interpretation and validation of cluster analysis},
	author       = {Peter J. Rousseeuw},
	year         = 1987,
	journal      = {Journal of Computational and Applied Mathematics},
	volume       = 20,
	pages        = {53--65},
	doi          = {doi.org/10.1016/0377-0427(87)90125-7}
}

\clearpage

\appendix
\normalsize

\renewcommand{\thetable}{\Alph{section}\arabic{table}}
\renewcommand{\thefigure}{\Alph{section}\arabic{figure}}
\setcounter{table}{0}
\setcounter{figure}{0}

	
	\section{Statistical Tests of Hyperparameters Influence}
	\label{apn:statistical_test}
	We present a pairwise statistical test~\citep{Kalpic2011} of hyperparameters in groups of direct effect and interaction effect of each algorithm. 
	Tables~\ref{tab:cmaes_stats} \ref{tab:de_stats}, 
	\ref{tab:nsgaiii_gd_stats}, 
	\ref{tab:nsgaiii_igd_stats}, 
	\ref{tab:nsgaiii_hv_stats},
	\ref{tab:moead_gd_stats}, 
	\ref{tab:moead_igd_stats}, and
	\ref{tab:moead_hv_stats} are statistical independent two-sample t-test results, where symbol `t' indicates t-statistics and `$p$' is the $p$-value. The samples for this t-test are direct effect and interaction (or total) effect values of each hyperparameter for the functions evaluated in this study. For single objective testbench, the sample size is 33, and for multi-objective testbench, the sample size is 10. In the t-test, one may choose a value of 0.05 or less to indicate the t-test is significant. 
	
	The common assumptions made when doing a t-test include those regarding the scale of measurement, random sampling, normality of data distribution, adequacy of sample size, and equality of variance in standard deviation.  The following assumptions for the t-test have been satisfied: the scale of measurement, random sampling, normality of data distribution, adequacy of sample size, and equality of variance in standard deviation.
	
	We consider the equality of variance of the two samples in the pair-wise t-test. We follow a standard t-test. The magnitude of t-statistics indicates the significance of the mean, and the sign (negative or positive) indicates the direction, that is, which sample is more significant than the other.
	
	For example, in Table~\ref{tab:cmaes_stats}, t-test between pair CMA-ES parameters $\lambda$ (first  column) and $\mu\lambda_{\mathrm{ratio}}$ (first row) indicate no significance of $\lambda$, although it indicates that the mean $\lambda$ is better than the mean of $\mu\lambda_{\mathrm{ratio}}$. This is because the $p$-value does not confirm its significance. Table~\ref{tab:cmaes_stats} interestingly shows that the interaction effect of $\lambda$ is much higher than the direct effect of $\lambda$, which is confirmed by its statistical significance (see row 9 and 10 and column 1 of Table~\ref{tab:cmaes_stats} where mean of $\lambda$ indicating interaction is better (t-stat is -2.14) since it has the $p$-value 0.04).  Moreover, the interaction effect of $\lambda$ is significantly better than all other hyperparameters. 
	In summary, 
	Tables~\ref{tab:cmaes_stats} \ref{tab:de_stats}, 
	\ref{tab:nsgaiii_gd_stats}, 
	\ref{tab:nsgaiii_igd_stats}, 
	\ref{tab:nsgaiii_hv_stats},
	\ref{tab:moead_gd_stats}, 
	\ref{tab:moead_igd_stats}, and
	\ref{tab:moead_hv_stats} have arranged such a way that one needs to read \textit{row} to find that the t-stat is significant only if it is negative and corresponding $p$-value less than $0.05$. Contrarily, if one reads \textit{column} direction, t-stat is significant only if it is positive and the corresponding $p$-value is less than $0.05$.
	
	In the case of single-objective optimization algorithms, results are shown in Tables~\ref{tab:cmaes_stats} and \ref{tab:de_stats} that support the results of the ranking of CMA-ES and DE hyperparameters. For instance, in Table~\ref{tab:cmaes_stats}, the high interaction of $\lambda$ is found to be more statistically significant than that of all other hyperparameters, including its own direct influence (read row-wise). This is followed by the strong direct influence of $\sigma_{0}$ (read column-wise). For DE, $\mathbf{b}_{\mathrm{type}}$ has both strong (and statistically significant) direct and interaction effects on the DE performance compared to all other hyperparameters (read column-wise in Table~\ref{tab:de_stats}). The hyperparameter $\lambda$ has higher statistically significant interaction influence on DE performance compared to hyperparameters $\mathbf{b}\lambda_{\mathrm{ratio}}$, $\mathrm{X}$, $\beta_{\mathrm{min}}$, and $\beta_{\mathrm{max}}$ (read row-wise).
	
	For multi-objective optimization, NSGA-III hyperparameters $\lambda$ and $P[\mathrm{X}]$ are observed to have a higher influence on NSGA-III performance than other hyperparameters (read columns in GD, IGD and HV metrics in Tables~\ref{tab:nsgaiii_gd_stats}, 
	\ref{tab:nsgaiii_igd_stats}, and \ref{tab:nsgaiii_hv_stats}). Similarly, for MOEAD, hyperparameter $\lambda$ and mutation related hyperparameters $P[\mathrm{PM}]$ and  $\mathrm{PM}_{\mathrm{DI}}$ observed to have a higher statistical significance of their performance (see Tables~\ref{tab:moead_gd_stats}, 
	\ref{tab:moead_igd_stats}, and
	\ref{tab:moead_hv_stats}).
	
	\begin{table}[!h]
		\centering
		\footnotesize
		\renewcommand{\arraystretch}{1}
		\setlength{\tabcolsep}{8.1pt}
		\caption{CMA-ES  Best Solution Metric T-Test Statistics
			\label{tab:cmaes_stats}}
		\begin{tabular}{llrrrrrrrrrr}
			\toprule
			& & & \multicolumn{5}{c}{Direct Effect} & \multicolumn{4}{c}{Interaction Effect}\\
			& Param & & $\lambda$ & $\mu\lambda_{\mathrm{ratio}}$ & $\sigma_0$ & $\alpha_{\mu}$ & $\sigma_{0-scale}$ & $\lambda$ & $\mu\lambda_{\mathrm{ratio}}$ & $\sigma_0$ & $\alpha_{\mu}$ \\
			\midrule
			\multirow{8}*{\rotatebox[origin=r]{90}{Direct Effect}} 
			& $\mu\lambda_{\mathrm{ratio}}$ & t & 1.05 &  & & & & &  & & \\
			&  & $p$ & 0.30 &  & & & & &  & & \\
			& $\sigma_0$ & t & -0.50  & -1.72 & & & & &  & & \\
			&  & $p$ & 0.62 & 0.09 & & & & &  & & \\
			& $\alpha_{\mu}$ & t & 1.30 & 0.28 & 2.01  & & & &  & & \\
			&  & $p$ & 0.20 & 0.78 & 0.05  & & & &  & & \\
			& $\sigma_{0-scale}$ & t & 1.38 & 0.41 & 2.06  & 0.14 & & &  & & \\
			&  & $p$ & 0.18 & 0.68 & 0.04  & 0.89 & & &  & & \\[1em]
			\multirow{10}*{\rotatebox[origin=r]{90}{Interaction Effect}} &  $\lambda$  & t & -2.14  & -3.80 & -1.72 & -4.20 & -4.15 & &  & & \\
			&  & $p$ & 0.04 & 0.00 & 0.09  & 0.00 & 0.00 & &  & & \\
			& $\mu\lambda_{\mathrm{ratio}}$ & t & 0.72 & -0.40 & 1.36  & -0.70 & -0.80 & 3.41 &  & & \\
			&  & $p$ & 0.47 & 0.69 & 0.18  & 0.49 & 0.43 & 0.00 &  & & \\
			& $\sigma_0$ & t & 1.67 & 0.85 & 2.29  & 0.63 & 0.48 & 4.16 & 1.19 & & \\
			&  & $p$ & 0.10 & 0.40 & 0.03  & 0.53 & 0.63 & 0.00 & 0.24 & & \\
			& $\alpha_{\mu}$ & t & 1.25 & 0.25 & 1.93  & -0.03 & -0.17 & 4.04 & 0.65 & -0.64 & \\
			&  & $p$ & 0.22 & 0.81 & 0.06  & 0.98 & 0.87 & 0.00 & 0.52 & 0.53  & \\
			& $\sigma_{0-scale}$ & t & 1.56 & 0.57 & 2.32  & 0.28 & 0.11 & 4.64 & 1.01 & -0.42 & 0.30 \\
			&  & $p$ & 0.13 & 0.57 & 0.02  & 0.78 & 0.91 & 0.00 & 0.32 & 0.67  & 0.76  \\
			\bottomrule
		\end{tabular}
	\end{table}

	\begin{table}[]
		\centering
		\footnotesize
		\renewcommand{\arraystretch}{1}
		\setlength{\tabcolsep}{4pt}
		\caption{DE Best Solution Metric T-Test Statistics
			\label{tab:de_stats}}
		\begin{tabular}{llrrrrrrrrrrrrrr}
			\toprule
			& & & \multicolumn{7}{c}{Direct Effect} & \multicolumn{6}{c}{Interaction Effect}\\
			& Param & & $\lambda$ & $\mathbf{b}_{\mathrm{type}}$ & $\mathbf{b}\lambda_{\mathrm{ratio}}$ & $\mathrm{X}$ & $P[\mathrm{X}]$ & $\beta_{\mathrm{min}}$ & $\beta_{\mathrm{max}}$ & $\lambda$ & $\mathbf{b}_{\mathrm{type}}$ & $\mathbf{b}\lambda_{\mathrm{ratio}}$ & $\mathrm{X}$ & $P[\mathrm{X}]$ & $\beta_{\mathrm{min}}$  \\
			\midrule
			\multirow{12}*{\rotatebox[origin=r]{90}{Direct Effect}} 
			& $\mathbf{b}_{\mathrm{type}}$ & t & -5.99 & & & & & & & & & & & & \\
			&  & $p$ & 0.00  & & & & & & & & & & & & \\
			& $\mathbf{b}\lambda_{\mathrm{ratio}}$ & t & 0.85  & 7.05 & & & & & & & & & & & \\
			&  & $p$ & 0.40  & 0.00 & & & & & & & & & & & \\
			& $\mathrm{X}$ & t & 0.81  & 7.50 & -0.10 & & & & & & & & & & \\
			&  & $p$ & 0.42  & 0.00 & 0.92  & & & & & & & & & & \\
			& $P[\mathrm{X}]$ & t & 0.27  & 7.13 & -0.70 & -0.65  & & & & & & & & & \\
			&  & $p$ & 0.79  & 0.00 & 0.49  & 0.52 & & & & & & & & & \\
			& $\beta_{\mathrm{min}}$ & t & 0.85  & 7.54 & -0.07 & 0.04 & 0.68  & & & & & & & & \\
			&  & $p$ & 0.40  & 0.00 & 0.94  & 0.97 & 0.50  & & & & & & & & \\
			& $\beta_{\mathrm{max}}$ & t & 0.75  & 7.39 & -0.16 & -0.07  & 0.57  & -0.10 & & & & & & & \\
			&  & $p$ & 0.46  & 0.00 & 0.87  & 0.95 & 0.57  & 0.92  & & & & & & & \\[0.5em]
			\multirow{14}*{\rotatebox[origin=r]{90}{Interaction Effect}} 
			& $\lambda$ & t & -1.19 & 4.79 & -2.08 & -2.13  & -1.62 & -2.17 & -2.06 & & & & & & \\
			& & $p$ & 0.24  & 0.00 & 0.04  & 0.04 & 0.11  & 0.03  & 0.04  & & & & & & \\
			& $\mathbf{b}_{\mathrm{type}}$ & t & -6.34 & -0.21 & -7.44 & -7.96  & -7.61 & -8.00 & -7.85 & -5.12 & & & & & \\
			&  & $p$ & 0.00  & 0.83 & 0.00  & 0.00 & 0.00  & 0.00  & 0.00  & 0.00  & & & & & \\
			& $\mathbf{b}\lambda_{\mathrm{ratio}}$ & t & -0.76 & 4.90 & -1.60 & -1.60  & -1.11 & -1.63 & -1.54 & 0.37  & 5.21 & & & & \\
			&  & $p$ & 0.45  & 0.00 & 0.12  & 0.11 & 0.27  & 0.11  & 0.13  & 0.71  & 0.00 & & & & \\
			& $\mathrm{X}$ & t & 0.61  & 6.79 & -0.25 & -0.16  & 0.42  & -0.19 & -0.10 & 1.84  & 7.19 & 1.37  & & & \\
			&  & $p$ & 0.54  & 0.00 & 0.81  & 0.87 & 0.67  & 0.85  & 0.92  & 0.07  & 0.00 & 0.18  & & & \\
			& $P[\mathrm{X}]$ & t & -0.07 & 6.31 & -0.98 & -0.96  & -0.37 & -0.99 & -0.89 & 1.19  & 6.71 & 0.74  & -0.73  & & \\
			&  & $p$ & 0.94  & 0.00 & 0.33  & 0.34 & 0.71  & 0.33  & 0.38  & 0.24  & 0.00 & 0.46  & 0.47 & & \\
			& $\beta_{\mathrm{min}}$ & t & 0.59  & 7.46 & -0.36 & -0.28  & 0.38  & -0.32 & -0.21 & 1.94  & 7.94 & 1.41  & -0.09  & 0.72  & \\
			&  & $p$ & 0.56  & 0.00 & 0.72  & 0.78 & 0.71  & 0.75  & 0.83  & 0.06  & 0.00 & 0.16  & 0.93 & 0.48  & \\
			& $\beta_{\mathrm{max}}$ & t & 1.10  & 8.26 & 0.14  & 0.27 & 0.98  & 0.24  & 0.34  & 2.51  & 8.80 & 1.92  & 0.42 & 1.28  & 0.59  \\
			&  & $p$ & 0.27  & 0.00 & 0.89  & 0.79 & 0.33  & 0.81  & 0.73  & 0.01  & 0.00 & 0.06  & 0.68 & 0.20  & 0.56   \\
			\bottomrule
		\end{tabular}
	\end{table}

	\begin{table}[]
		\centering
		\footnotesize
		\renewcommand{\arraystretch}{1}
		\setlength{\tabcolsep}{5pt}
		\caption{NSGA-III GD Metric T-Test Statistics
			\label{tab:nsgaiii_gd_stats}}
		\begin{tabular}{llrrrrrrrrrrrr}
			\toprule
			& & & \multicolumn{6}{c}{Direct Effect} & \multicolumn{5}{c}{Interaction Effect}\\
			& Param & & $\lambda$ & $P[\mathrm{X}]$ & $\mathrm{X}_{\mathrm{DI}}$ & $P[\mathrm{PM}]$ & $\mathrm{PM}_{\mathrm{DI}}$ & $K$ & $\lambda$ & $P[\mathrm{X}]$ & $\mathrm{X}_{\mathrm{DI}}$ & $P[\mathrm{PM}]$ & $\mathrm{PM}_{\mathrm{DI}}$ \\
			\midrule
			\multirow{10}*{\rotatebox[origin=r]{90}{Direct Effect}} 
			& $P[\mathrm{X}]$ & t & 2.62  & & & & & & & & & &   \\
			&  & $p$ & 0.02  & & & & & & & & & &   \\
			& $\mathrm{X}_{\mathrm{DI}}$ & t & 7.98  & 2.53 & & & & & & & & &  \\
			&  & $p$ & 0.00  & 0.02 & & & & & & & & &  \\
			& $P[\mathrm{PM}]$ & t & 8.87  & 2.83 & 0.64 & & & & & & & &  \\
			&  & $p$ & 0.00  & 0.01 & 0.53 & & & & & & & &  \\
			& $\mathrm{PM}_{\mathrm{DI}}$ & t & 7.61  & 2.29 & -0.56 & -1.32 & & & & & & &  \\
			&  & $p$ & 0.00  & 0.03 & 0.58 & 0.20 & & & & & & &  \\
			& $K$ & t & 10.27  & 3.63 & 3.26 & 5.07 & 3.95 & & & & & & \\
			&  & $p$ & 0.00  & 0.00 & 0.00 & 0.00 & 0.00 & & & & & & \\[0.5em]
			\multirow{12}*{\rotatebox[origin=r]{90}{Interaction Effect}} 
			& $\lambda$ & t & -0.98  & -3.64 & -13.71 & -16.94 & -13.06 & -20.40 & & & & & \\
			&  & $p$ & 0.34  & 0.00 & 0.00 & 0.00 & 0.00 & 0.00 & & & & & \\
			& $P[\mathrm{X}]$ & t & 8.29  & 2.76 & 0.55 & 0.07 & 1.10 & -2.44 & 14.15  & & & & \\
			&  & $p$ & 0.00  & 0.01 & 0.59 & 0.95 & 0.29 & 0.03 & 0.00  & & & & \\
			& $\mathrm{X}_{\mathrm{DI}}$ & t & 6.21  & 2.02 & -0.44 & -0.84 & -0.08 & -2.17 & 8.97  & -0.80 & & & \\
			&  & $p$ & 0.00  & 0.06 & 0.66 & 0.41 & 0.94 & 0.04 & 0.00  & 0.43 & & & \\
			& $P[\mathrm{PM}]$ & t & 6.14  & 2.02 & -0.40 & -0.77 & -0.04 & -2.06 & 8.78  & -0.75 & 0.03 & & \\
			&  & $p$ & 0.00  & 0.06 & 0.70 & 0.45 & 0.97 & 0.05 & 0.00  & 0.47 & 0.98 & & \\
			& $\mathrm{PM}_{\mathrm{DI}}$ & t & 6.55  & 2.26 & -0.05 & -0.41 & 0.32 & -1.75 & 9.45  & -0.41 & 0.32 & 0.28 & \\
			& & $p$ & 0.00  & 0.04 & 0.96 & 0.68 & 0.75 & 0.10 & 0.00  & 0.69 & 0.75 & 0.78 & \\
			& $K$  & t & 4.78  & 1.08 & -1.90 & -2.37 & -1.56 & -3.61 & 6.92  & -2.23 & -1.22 & -1.22 & -1.53 \\
			&  & $p$ & 0.00  & 0.30 & 0.07 & 0.03 & 0.14 & 0.00 & 0.00  & 0.04 & 0.24 & 0.24 & 0.14 \\
			\bottomrule
		\end{tabular}
	\end{table}

	\begin{table}[]
		\centering
		\footnotesize
		\renewcommand{\arraystretch}{1}
		\setlength{\tabcolsep}{5pt}
		\caption{NSGA-III IGD  Metric T-Test Statistics
			\label{tab:nsgaiii_igd_stats}}
		\begin{tabular}{llrrrrrrrrrrrr}
			\toprule
			& & & \multicolumn{6}{c}{Direct Effect} & \multicolumn{5}{c}{Interaction Effect}\\
			& Param  & & $\lambda$ & $P[\mathrm{X}]$ & $\mathrm{X}_{\mathrm{DI}}$ & $P[\mathrm{PM}]$ & $\mathrm{PM}_{\mathrm{DI}}$ & $K$ & $\lambda$ & $P[\mathrm{X}]$ & $\mathrm{X}_{\mathrm{DI}}$ & $P[\mathrm{PM}]$ & $\mathrm{PM}_{\mathrm{DI}}$ \\
			\midrule
			\multirow{10}*{\rotatebox[origin=r]{90}{Direct Effect}} 
			& $P[\mathrm{X}]$ & t & -0.16  & & &  & & & & & &  & \\
			&  & $p$ & 0.87 & & &  & & & & & &  & \\
			& $\mathrm{X}_{\mathrm{DI}}$ & t & 3.86 & 3.92 & &  & & & & & &  & \\
			&  & $p$ & 0.00 & 0.00 & &  & & & & & &  & \\
			& $P[\mathrm{PM}]$ & t & 4.39 & 4.42 & 1.34  &  & & & & & &  & \\
			&  & $p$ & 0.00 & 0.00 & 0.20  &  & & & & & &  & \\
			& $\mathrm{PM}_{\mathrm{DI}}$ & t & 3.86 & 3.92 & -0.20 & -1.87 & & & & & &  & \\
			&  & $p$ & 0.00 & 0.00 & 0.85  & 0.08 & & & & & &  & \\
			& $K$  & t & 4.36 & 4.40 & 1.32  & 0.31 & 1.65 & & & & &  & \\
			&  & $p$ & 0.00 & 0.00 & 0.20  & 0.76 & 0.12 & & & & &  & \\[0.5em]
			
			\multirow{12}*{\rotatebox[origin=r]{90}{Interaction Effect}} 
			& $\lambda$ & t & -2.87  & -2.52 & -24.39 & -40.08 & -32.61 & -23.14 & & & &  & \\
			&  & $p$ & 0.01 & 0.02 & 0.00  & 0.00 & 0.00 & 0.00 & & & &  & \\
			& $P[\mathrm{X}]$ & t & 4.28 & 4.32 & 1.09  & -0.04 & 1.43 & -0.30 & 25.81  & & &  & \\
			& & $p$ & 0.00 & 0.00 & 0.29  & 0.97 & 0.17 & 0.77 & 0.00 & & &  & \\
			& $\mathrm{X}_{\mathrm{DI}}$ & t & 2.32 & 2.43 & -1.54 & -2.22 & -1.49  & -2.23 & 8.06 & -2.11 & &  & \\
			&  & $p$ & 0.03 & 0.03 & 0.14  & 0.04 & 0.15 & 0.04 & 0.00 & 0.05 & &  & \\
			& $P[\mathrm{PM}]$ & t & 2.16 & 2.27 & -1.43 & -2.01 & -1.38  & -2.05 & 6.94 & -1.93 & -0.06 &  & \\
			&  & $p$ & 0.04 & 0.04 & 0.17  & 0.06 & 0.18 & 0.05 & 0.00 & 0.07 & 0.96  &  & \\
			& $\mathrm{PM}_{\mathrm{DI}}$ & t & 2.53 & 2.64 & -1.02 & -1.63 & -0.96  & -1.68 & 7.94 & -1.55 & 0.33  & 0.36 & \\
			&  & $p$ & 0.02 & 0.02 & 0.32  & 0.12 & 0.35 & 0.11 & 0.00 & 0.14 & 0.74  & 0.72 & \\
			& $K$  & t & 1.74 & 1.87 & -2.12 & -2.74 & -2.09  & -2.74 & 6.40 & -2.63 & -0.58 & -0.49 & -0.87  \\
			&  & $p$ & 0.10 & 0.08 & 0.05  & 0.01 & 0.05 & 0.01 & 0.00 & 0.02 & 0.57  & 0.63 & 0.40 \\
			\bottomrule
		\end{tabular}
	\end{table}
	
	\begin{table}[]
		\centering
		\footnotesize
		\renewcommand{\arraystretch}{1}
		\setlength{\tabcolsep}{5pt}
		\caption{NSGA-III HV Metric T-Test Statistics
			\label{tab:nsgaiii_hv_stats}}
		\begin{tabular}{llrrrrrrrrrrrr}
			\toprule
			& & & \multicolumn{6}{c}{Direct Effect} & \multicolumn{5}{c}{Interaction Effect}\\
			& Param &  & $\lambda$ & $P[\mathrm{X}]$ & $\mathrm{X}_{\mathrm{DI}}$ & $P[\mathrm{PM}]$ & $\mathrm{PM}_{\mathrm{DI}}$ & $K$  & $\lambda$ & $P[\mathrm{X}]$ & $\mathrm{X}_{\mathrm{DI}}$ & $P[\mathrm{PM}]$ & $\mathrm{PM}_{\mathrm{DI}}$ \\
			\midrule
			\multirow{10}*{\rotatebox[origin=r]{90}{Direct Effect}} 
			& $P[\mathrm{X}]$ & t & 19.13 & &  & & &  & & &  & & \\
			&  & $p$ & 0.00 & &  & & &  & & &  & & \\
			& $\mathrm{X}_{\mathrm{DI}}$ & t & 2.60 & -9.16 &  & & &  & & &  & & \\
			&  & $p$ & 0.02 & 0.00 &  & & &  & & &  & & \\
			& $P[\mathrm{PM}]$ & t & 2.29 & -8.67 & -0.14   & & &  & & &  & & \\
			&  & $p$ & 0.04 & 0.00 & 0.89 & & &  & & &  & & \\
			& $\mathrm{PM}_{\mathrm{DI}}$ & t & 2.17 & -9.02 & -0.27   & -0.13 & &  & & &  & & \\
			&  & $p$ & 0.05 & 0.00 & 0.79 & 0.90 & &  & & &  & & \\
			& $K$  & t & 1.02 & -8.86 & -1.06   & -0.90 & -0.79 &  & & &  & & \\
			&  & $p$ & 0.33 & 0.00 & 0.31 & 0.38 & 0.44 &  & & &  & & \\[0.5em]
			\multirow{12}*{\rotatebox[origin=r]{90}{Interaction Effect}} 
			& $\lambda$  & t & 2.43 & -5.12 & 0.57 & 0.66 & 0.77 & 1.37 & & &  & & \\
			&  & $p$ & 0.03 & 0.00 & 0.58 & 0.52 & 0.46 & 0.20 & & &  & & \\
			& $P[\mathrm{X}]$ & t & -1.62 & --   & -4.12   & -3.63 & -3.52 & -2.00 & -3.28 & &  & & \\
			&  & $p$ & 0.13 & 0.00 & 0.00 & 0.00 & 0.00 & 0.07 & 0.01 & &  & & \\
			& $\mathrm{X}_{\mathrm{DI}}$ & t & 15.88 & -10.34   & 7.32 & 6.98 & 7.29 & 7.37 & 4.00 & 68.71 &  & & \\
			&  & $p$ & 0.00 & 0.00 & 0.00 & 0.00 & 0.00 & 0.00 & 0.00 & 0.00 &  & & \\
			& $P[\mathrm{PM}]$ & t & 14.17 & -2.57 & 7.30 & 7.02 & 7.30 & 7.43 & 4.22 & 26.94 & 1.21 & & \\
			&  & $p$ & 0.00 & 0.02 & 0.00 & 0.00 & 0.00 & 0.00 & 0.00 & 0.00 & 0.25 & & \\
			& $\mathrm{PM}_{\mathrm{DI}}$ & t & 16.08 & -4.21 & 7.78 & 7.42 & 7.74 & 7.77 & 4.36 & 46.06 & 1.99 & 0.08 & \\
			&  & $p$ & 0.00 & 0.00 & 0.00 & 0.00 & 0.00 & 0.00 & 0.00 & 0.00 & 0.07 & 0.94 & \\
			& $K$  & t & 15.30 & -1.12 & 7.99 & 7.65 & 7.95 & 8.00 & 4.65 & 29.99 & 2.75 & 1.09 & 1.27 \\
			&  & $p$ & 0.00 & 0.29 & 0.00 & 0.00 & 0.00 & 0.00 & 0.00 & 0.00 & 0.02 & 0.30 & 0.23 \\
			\bottomrule
		\end{tabular}
	\end{table}

	\begin{table}[]
		\centering
		\footnotesize
		\renewcommand{\arraystretch}{0.8}
		\setlength{\tabcolsep}{3pt}
		\caption{MOEA/D GD Metric T-Test Statistics
			\label{tab:moead_gd_stats}}
		\begin{tabular}{llrrrrrrrrrrrrrr}
			\toprule
			& & & \multicolumn{7}{c}{Direct Effect} & \multicolumn{6}{c}{Interaction Effect}\\
			& Param & & $\lambda$ & $P[\mathrm{X}]$ & $\mathrm{X}_{\mathrm{DI}}$ & $P[\mathrm{PM}]$ & $\mathrm{PM}_{\mathrm{DI}}$ & $Mode$ & $\epsilon_N$ & $\lambda$ & $P[\mathrm{X}]$ & $\mathrm{X}_{\mathrm{DI}}$ & $P[\mathrm{PM}]$ & $\mathrm{PM}_{\mathrm{DI}}$ & $Mode$ \\
			\midrule
			\multirow{14}*{\rotatebox[origin=r]{90}{Direct Effect}} 
			& $P[\mathrm{X}]$ & t & 9.23 & & & & & & & & & & & & \\
			&  & $p$ & 0.00 & & & & & & & & & & & & \\
			& $\mathrm{X}_{\mathrm{DI}}$ & t & 50.53 & 1.92 & & & & & & & & & & & \\
			&  & $p$ & 0.00 & 0.07 & & & & & & & & & & & \\
			& $P[\mathrm{PM}]$ & t & 25.00 & 0.87 & -2.24 & & & & & & & & & & \\
			&  & $p$ & 0.00 & 0.40 & 0.04 & & & & & & & & & & \\
			& $\mathrm{PM}_{\mathrm{DI}}$ & t & 12.55 & 0.01 & -2.55 & -1.10 & & & & & & & & & \\
			&  & $p$ & 0.00 & 0.99 & 0.02 & 0.29 & & & & & & & & & \\
			& $Mode$ & t & 11.84 & 0.05 & -2.33 & -0.98 & 0.05 & & & & & & & & \\
			& & $p$ & 0.00 & 0.96 & 0.03 & 0.34 & 0.96 & & & & & & & & \\
			& $\epsilon_N$ & t & 19.16 & 1.08 & -1.26 & 0.43 & 1.33 & 1.21 & & & & & & & \\
			& & $p$ & 0.00 & 0.30 & 0.22 & 0.67 & 0.20 & 0.24 & & & & & & & \\[0.5em]
			\multirow{14}*{\rotatebox[origin=r]{90}{Interaction Effect}} 
			& $\lambda$ & t & 1.96 & -2.76 & -4.22 & -3.59 & -2.93 & -2.93 & -3.68 & & & & & & \\
			&  & $p$ & 0.07 & 0.01 & 0.00 & 0.00 & 0.01 & 0.01 & 0.00 & & & & & & \\
			& $P[\mathrm{X}]$ & t & 9.63 & 0.07 & -1.88 & -0.79 & 0.08 & 0.03 & -1.01 & 2.83 & & & & & \\
			&  & $p$ & 0.00 & 0.94 & 0.08 & 0.44 & 0.94 & 0.98 & 0.33 & 0.01 & & & & & \\
			& $\mathrm{X}_{\mathrm{DI}}$ & t & 8.66 & -0.58 & -2.78 & -1.65 & -0.67 & -0.70 & -1.82 & 2.39 & -0.66 & & & & \\
			&  & $p$ & 0.00 & 0.57 & 0.01 & 0.12 & 0.51 & 0.49 & 0.09 & 0.03 & 0.52 & & & & \\
			& $P[\mathrm{PM}]$ & t & 8.11 & 0.22 & -1.36 & -0.48 & 0.23 & 0.19 & -0.68 & 2.80 & 0.15 & 0.75 & & & \\
			&  & $p$ & 0.00 & 0.83 & 0.19 & 0.64 & 0.82 & 0.85 & 0.50 & 0.01 & 0.88 & 0.46 & & & \\
			& $\mathrm{PM}_{\mathrm{DI}}$ & t & 9.52 & 0.46 & -1.23 & -0.24 & 0.51 & 0.46 & -0.48 & 3.06 & 0.40 & 1.03 & 0.21 & & \\
			& & $p$ & 0.00 & 0.65 & 0.24 & 0.81 & 0.61 & 0.65 & 0.64 & 0.01 & 0.70 & 0.31 & 0.84 & & \\
			& $Mode$ & t & 4.70 & -0.93 & -2.31 & -1.63 & -1.00 & -1.02 & -1.76 & 1.64 & -0.99 & -0.50 & -1.05 & -1.27 & \\
			&  & $p$ & 0.00 & 0.37 & 0.03 & 0.12 & 0.33 & 0.32 & 0.09 & 0.12 & 0.34 & 0.62 & 0.31 & 0.22 & \\
			& $\epsilon_N$ & t & 9.21 & -0.45 & -2.69 & -1.52 & -0.52 & -0.56 & -1.70 & 2.50 & -0.53 & 0.14 & -0.63 & -0.91 & 0.61 \\
			& & $p$ & 0.00 & 0.66 & 0.02 & 0.15 & 0.61 & 0.58 & 0.11 & 0.02 & 0.60 & 0.89 & 0.54 & 0.37 & 0.55 \\
			\bottomrule
		\end{tabular}
	\end{table}

	\begin{table}[]
		\centering
		\footnotesize
		\renewcommand{\arraystretch}{0.8}
		\setlength{\tabcolsep}{3pt}
		\caption{MOEA/D IGD Metric T-Test Statistics
			\label{tab:moead_igd_stats}}
		\begin{tabular}{llrrrrrrrrrrrrrr}
			\toprule
			& & & \multicolumn{7}{c}{Direct Effect} & \multicolumn{6}{c}{Interaction Effect}\\
			& Param & & $\lambda$ & $P[\mathrm{X}]$ & $\mathrm{X}_{\mathrm{DI}}$ & $P[\mathrm{PM}]$ & $\mathrm{PM}_{\mathrm{DI}}$ & $Mode$ & $\epsilon_N$ & $\lambda$ & $P[\mathrm{X}]$ & $\mathrm{X}_{\mathrm{DI}}$ & $P[\mathrm{PM}]$ & $\mathrm{PM}_{\mathrm{DI}}$ & $Mode$ \\
			\midrule
			\multirow{12}*{\rotatebox[origin=r]{90}{Direct Effect}} 
			& $P[\mathrm{X}]$ & t & 0.77 & & & & & & & & & & & & \\
			&  & $p$ & 0.45 & & & & & & & & & & & & \\
			& $\mathrm{X}_{\mathrm{DI}}$ & t & 1.64 & 1.02 & & & & & & & & & & & \\
			&  & $p$ & 0.12 & 0.32 & & & & & & & & & & & \\
			& $P[\mathrm{PM}]$ & t & -0.43 & -1.54 & -3.02 & & & & & & & & & & \\
			&  & $p$ & 0.68 & 0.14 & 0.01 & & & & & & & & & & \\
			& $\mathrm{PM}_{\mathrm{DI}}$ & t & -0.59 & -1.71 & -3.15 & -0.22 & & & & & & & & & \\
			&  & $p$ & 0.56 & 0.10 & 0.01 & 0.83 & & & & & & & & & \\
			& $Mode$ & t & 2.27 & 1.85 & 1.15 & 3.69 & 3.80 & & & & & & & & \\
			&  & $p$ & 0.04 & 0.08 & 0.26 & 0.00 & 0.00 & & & & & & & & \\
			& $\epsilon_N$ & t & 1.27 & 0.56 & -0.53 & 2.40 & 2.56 & -1.54 & & & & & & & \\
			&  & $p$ & 0.22 & 0.58 & 0.60 & 0.03 & 0.02 & 0.14 & & & & & & & \\[0.5em]
			\multirow{14}*{\rotatebox[origin=r]{90}{Interaction Effect}} 
			& $\lambda$ & t & 0.92 & 0.31 & -0.41 & 1.51 & 1.64 & -1.08 & -0.09 & & & & & & \\
			&  & $p$ & 0.37 & 0.76 & 0.69 & 0.15 & 0.12 & 0.29 & 0.93 & & & & & & \\
			& $P[\mathrm{X}]$ & t & 0.26 & -0.57 & -1.57 & 0.81 & 0.98 & -2.28 & -1.14 & -0.75 & & & & & \\
			&  & $p$ & 0.80 & 0.58 & 0.13 & 0.43 & 0.34 & 0.03 & 0.27 & 0.46 & & & & & \\
			& $\mathrm{X}_{\mathrm{DI}}$ & t & 0.65 & -0.24 & -1.57 & 1.54 & 1.72 & -2.45 & -0.96 & -0.51 & 0.42 & & & & \\
			&  & $p$ & 0.52 & 0.81 & 0.13 & 0.14 & 0.10 & 0.02 & 0.35 & 0.62 & 0.68 & & & & \\
			& $P[\mathrm{PM}]$ & t & 2.19 & 1.75 & 1.03 & 3.57 & 3.68 & -0.11 & 1.42 & 1.01 & 2.19 & 2.33 & & & \\
			&  & $p$ & 0.04 & 0.10 & 0.32 & 0.00 & 0.00 & 0.92 & 0.17 & 0.33 & 0.04 & 0.03 & & & \\
			& $\mathrm{PM}_{\mathrm{DI}}$ & t & 1.63 & 1.03 & 0.05 & 2.94 & 3.07 & -1.06 & 0.55 & 0.42 & 1.56 & 1.54 & -0.94 & & \\
			&  & $p$ & 0.12 & 0.32 & 0.96 & 0.01 & 0.01 & 0.30 & 0.59 & 0.68 & 0.14 & 0.14 & 0.36 & & \\
			& $Mode$ & t & -1.08 & -2.31 & -3.85 & -0.88 & -0.64 & -4.43 & -3.23 & -2.12 & -1.54 & -2.41 & -4.31 & -3.75 & \\
			&  & $p$ & 0.29 & 0.03 & 0.00 & 0.39 & 0.53 & 0.00 & 0.00 & 0.05 & 0.14 & 0.03 & 0.00 & 0.00 & \\
			& $\epsilon_N$ & t & 0.58 & -0.31 & -1.58 & 1.39 & 1.57 & -2.43 & -1.01 & -0.56 & 0.33 & -0.09 & -2.31 & -1.55 & 2.24 \\
			&  & $p$ & 0.57 & 0.76 & 0.13 & 0.18 & 0.13 & 0.03 & 0.33 & 0.58 & 0.74 & 0.93 & 0.03 & 0.14 & 0.04 \\
			\bottomrule
		\end{tabular}
	\end{table}

	\begin{table}[]
		\centering
		\footnotesize
		\renewcommand{\arraystretch}{0.8}
		\setlength{\tabcolsep}{3pt}
		\caption{MOEA/D HV Metric T-Test Statistics
			\label{tab:moead_hv_stats}} 
		\begin{tabular}{llrrrrrrrrrrrrrr}
			\toprule
			& & & \multicolumn{7}{c}{Direct Effect} & \multicolumn{6}{c}{Interaction Effect}\\
			& Param  &  & $\lambda$ & $P[\mathrm{X}]$ & $\mathrm{X}_{\mathrm{DI}}$ & $P[\mathrm{PM}]$ & $\mathrm{PM}_{\mathrm{DI}}$ & $Mode$ & $\epsilon_N$ & $\lambda$ & $P[\mathrm{X}]$ & $\mathrm{X}_{\mathrm{DI}}$ & $P[\mathrm{PM}]$ & $\mathrm{PM}_{\mathrm{DI}}$ & $Mode$ \\
			\midrule
			\multirow{14}*{\rotatebox[origin=r]{90}{Direct Effect}} 
			& $P[\mathrm{X}]$  & t & -1.27   & & & & & &  & & & & & & \\
			&   & $p$ & 0.23   & & & & & &  & & & & & &  \\
			& $\mathrm{X}_{\mathrm{DI}}$ & t & -1.38   & -0.15 & & & & &  & & & & & & \\
			&  & $p$ & 0.19   & 0.88 & & & & &  & & & & & & \\
			& $P[\mathrm{PM}]$ & t & 0.71   & 2.32 & 2.41  & & & &  & & & & & & \\
			&  & $p$ & 0.49   & 0.04 & 0.03  & & & &  & & & & & & \\
			& $\mathrm{PM}_{\mathrm{DI}}$ & t & 1.82   & 3.44 & 3.47  & 1.77  & & &  & & & & & & \\
			&  & $p$ & 0.09   & 0.00 & 0.00  & 0.10  & & &  & & & & & & \\
			& $Mode$  & t & -2.67   & -1.59 & -1.43  & -3.68 & -4.56  & &  & & & & & & \\
			&   & $p$ & 0.02   & 0.13 & 0.17  & 0.00  & 0.00  & &  & & & & & & \\
			& $\epsilon_N$   & t & -1.12   & 0.17 & 0.32  & -2.19 & -3.34  & 1.76  &  & & & & & & \\
			&    & $p$ & 0.28   & 0.87 & 0.75  & 0.05  & 0.00  & 0.10  &  & & & & & & \\[0.5em]
			\multirow{14}*{\rotatebox[origin=r]{90}{Interaction Effect}} 
			& $\lambda$ & t & -1.68   & -0.58 & -0.44  & -2.57 & -3.47  & 0.93  & -0.74 & & & & & & \\
			&  & $p$ & 0.12   & 0.57 & 0.67  & 0.02  & 0.00  & 0.37  & 0.47 & & & & & & \\
			& $P[\mathrm{X}]$  & t & -0.60   & 0.82 & 0.96  & -1.70 & -3.07  & 2.40  & 0.65 & 1.32   & & & & & \\
			&  & $p$ & 0.56   & 0.42 & 0.35  & 0.11  & 0.01  & 0.03  & 0.52 & 0.21   & & & & & \\
			& $\mathrm{X}_{\mathrm{DI}}$ & t & -0.14   & 1.28 & 1.40  & -1.05 & -2.38  & 2.76  & 1.12 & 1.70   & 0.53 & & & & \\
			&  & $p$ & 0.89   & 0.22 & 0.18  & 0.31  & 0.03  & 0.02  & 0.28 & 0.11   & 0.60 & & & & \\
			& $P[\mathrm{PM}]$ & t & -2.00   & -0.68 & -0.50  & -3.36 & -4.56  & 1.07  & -0.87 & 0.01   & -1.64 & -2.10  & & & \\
			& & $p$ & 0.07   & 0.51 & 0.62  & 0.00  & 0.00  & 0.30  & 0.40 & 0.99   & 0.12 & 0.05  & & & \\
			& $\mathrm{PM}_{\mathrm{DI}}$ & t & -2.95   & -1.80 & -1.62  & -4.17 & -5.14  & -0.08 & -1.99 & -1.06   & -2.71 & -3.10  & -1.25 & & \\
			&  & $p$ & 0.01   & 0.09 & 0.13  & 0.00  & 0.00  & 0.93  & 0.07 & 0.31   & 0.02 & 0.01  & 0.23  & & \\
			& $Mode$  & t & 1.70   & 3.28 & 3.32  & 1.55  & -0.10  & 4.43  & 3.17 & 3.35   & 2.86 & 2.21  & 4.35  & 4.97  & \\
			&  & $p$ & 0.11   & 0.01 & 0.01  & 0.14  & 0.92  & 0.00  & 0.01 & 0.00   & 0.01 & 0.04  & 0.00  & 0.00  & \\
			& $\epsilon_N$ & t & -0.31   & 1.16 & 1.28  & -1.37 & -2.81  & 2.70  & 0.99 & 1.61   & 0.36 & -0.20  & 2.02  & 3.04  & -2.60 \\
			&  & $p$ & 0.76   & 0.27 & 0.22  & 0.19  & 0.01  & 0.02  & 0.34 & 0.13   & 0.72 & 0.85  & 0.06  & 0.01  & 0.02  \\
			\bottomrule
		\end{tabular}
	\end{table}

	\clearpage
	
	\section{Clustering Analysis of Hyperparameters and Testbench}
	\label{apn:clustering}
	We present a clustering analysis of hyperparameters' performance. For this analysis, we take direct and interaction (total) influence values together in a matrix form and apply the k-means algorithm~\citep{KMeans}. The number of clusters was automatically chosen based on silhouette scores~\citep{Silhouette} as per the formula: K = max\{2, argmax(silhouette scores)\}. This formula selects a value of K equal to 2 or more, for which K-means clustering produces the maximum silhouette score. We analyze algorithms' performance for hyperparameters and functions in the testbench for each sampling and metric used.
	
\begin{figure}
		\centering
		\includegraphics[width=\linewidth]{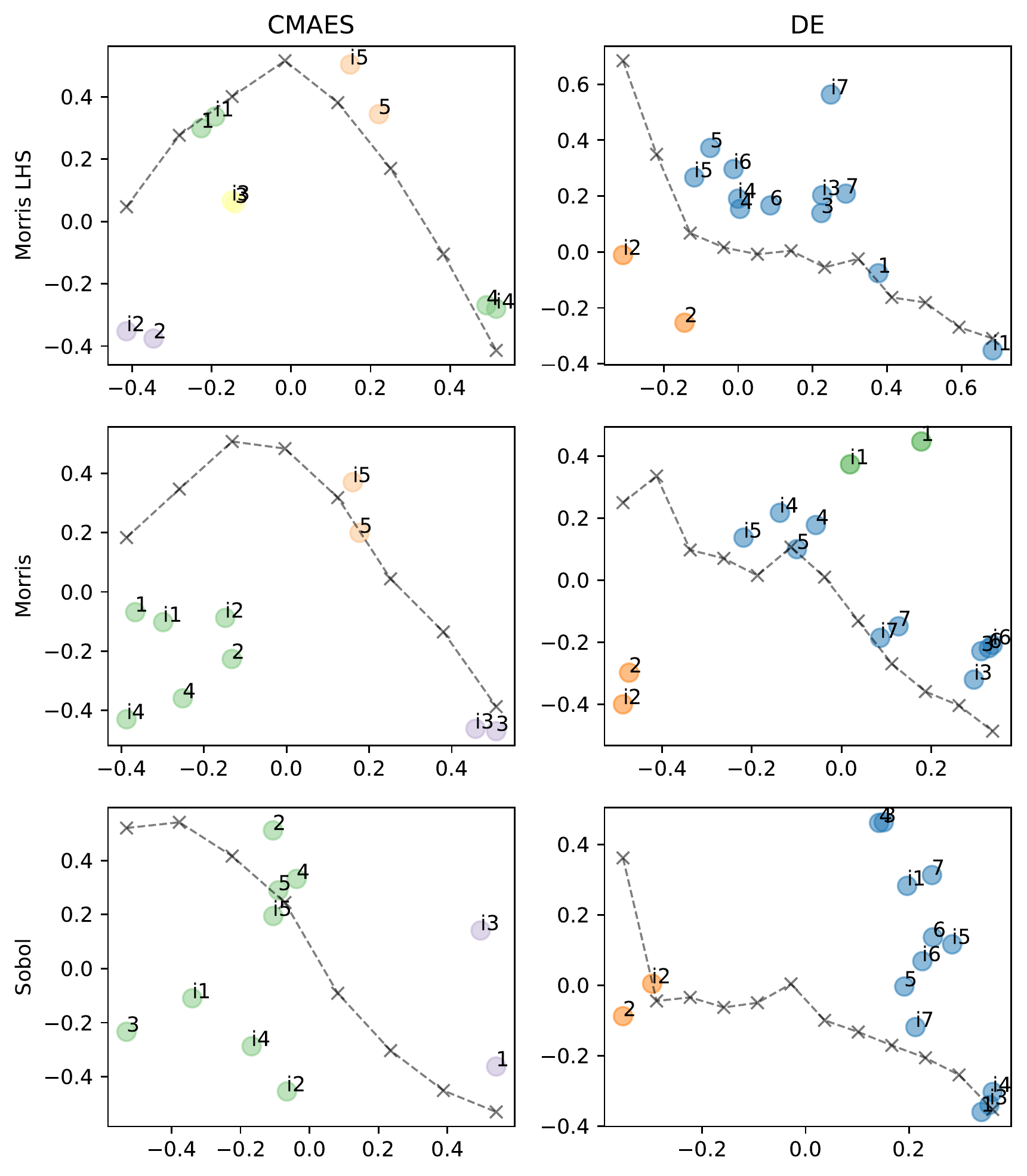}
		\caption{DE and CMA-ES algorithm hyperparameters K-means influence clustering. K is chosen automatically by using the formula K = max\{2, argmax(silhouette scores)\}, which is the cross at the highest peak of the dotted line. CMA-ES hyperparameters numbered from 1 to 5 are $\lambda$, $\mu\lambda_{\mathrm{ratio}}$, $\sigma_0$, and $\alpha_{\mu}$, $\sigma_{0-scale}$, respectively and DE hyperparameters numbered from 1 to 7 are $\lambda$, $\mathbf{b}_{\mathrm{type}}$, $\mathbf{b}\lambda_{\mathrm{ratio}}$, $\mathrm{X}$, $P[\mathrm{X}]$, $\beta_{\mathrm{min}}$, and $\beta_{\mathrm{max}}$, respectively.  A hyperparameter number appearing twice indicates the values of direct and interaction effect. A letter `i' as a prefix to a number in the plot indicates interaction/total effect, and the of 'i' indicates direct effect. Different colors represent different clusters.   
			\label{fig:DE_CMA_Cluster_param}}
	\end{figure}

\begin{figure}
		\centering
		\includegraphics[width=\linewidth]{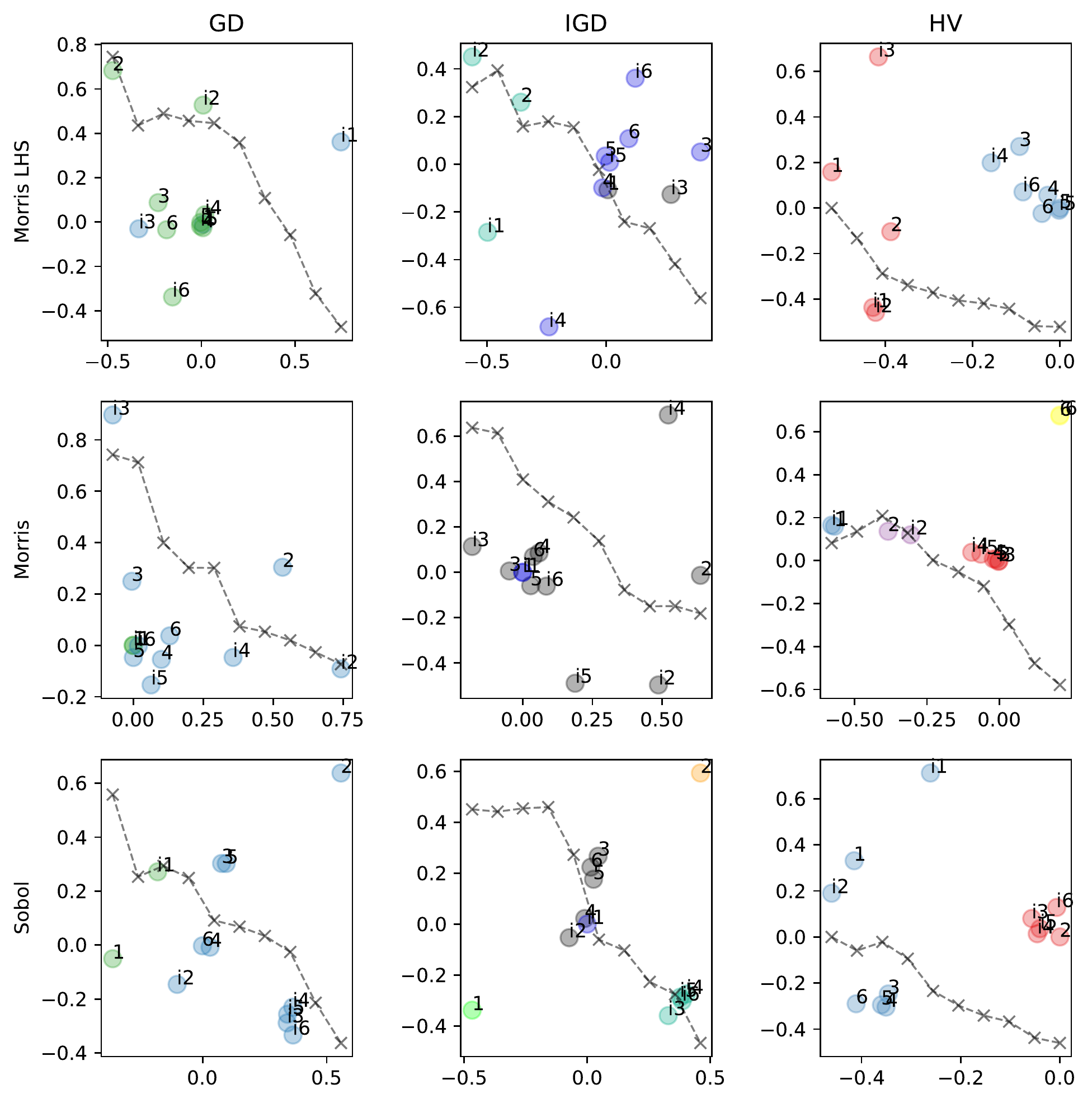}
		\caption{NSGA-III algorithm hyperparameters K-means influence clustering. K is chosen automatically by using the formula K = max\{2, argmax(silhouette scores)\}, which is the cross at the highest peak of the dotted line. NSGA-III hyperparameters numbered from 1 to 6  are $\lambda$, $P[\mathrm{X}]$, $\mathrm{X}_{\mathrm{DI}}$, $P[\mathrm{PM}]$, $\mathrm{PM}_{\mathrm{DI}}$, and $K$, respectively. A hyperparameter number appearing twice indicates the values of direct and interaction effect. A letter `i' as a prefix to a number in the plot indicates interaction/total effect, and the of 'i' indicates direct effect. Different colors represent different clusters.
			\label{fig:NSGAIII_cluster_param}}
	\end{figure}

\begin{figure}
		\centering
		\includegraphics[width=\linewidth]{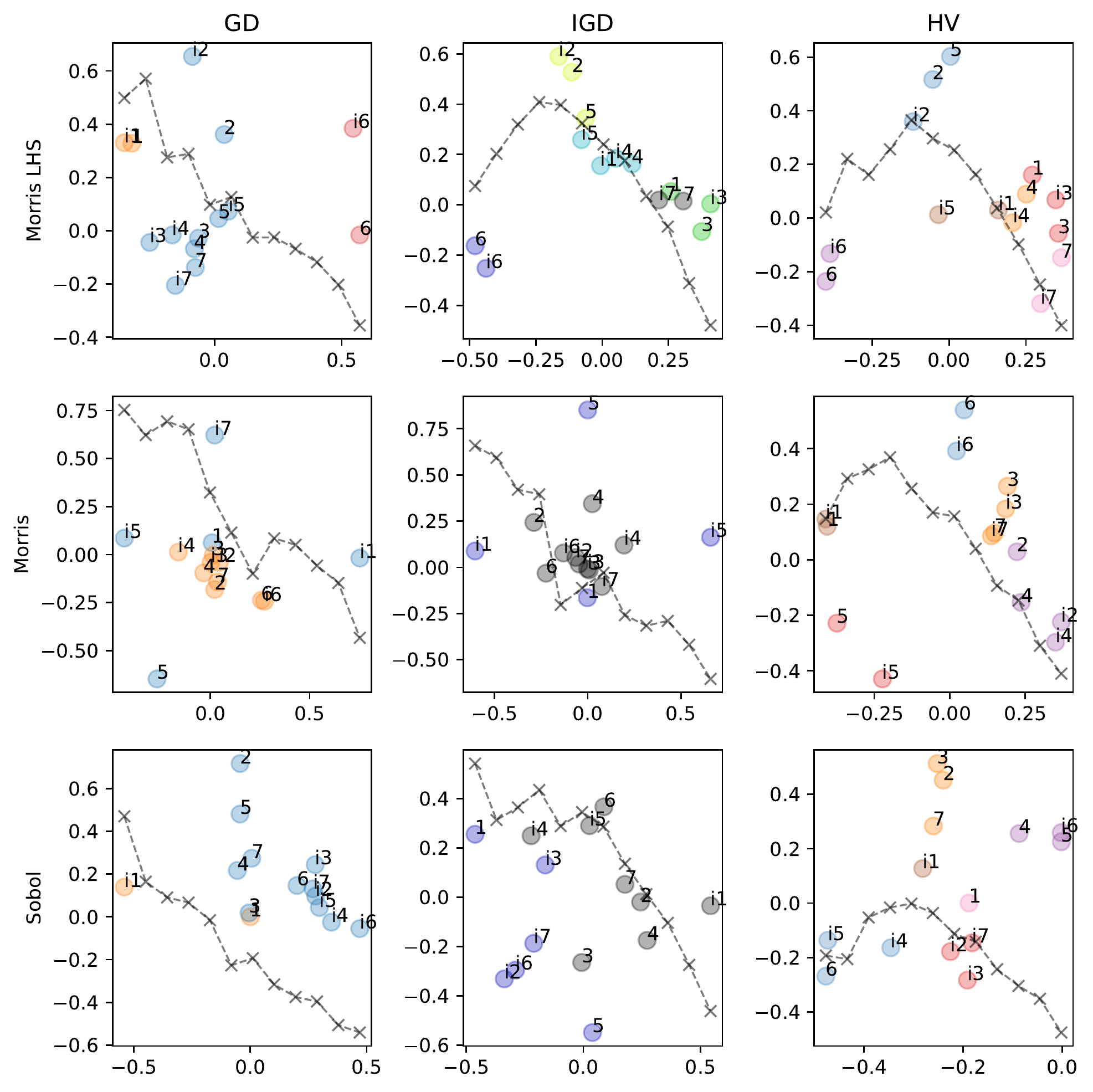}
		\caption{MOEA/D algorithm hyperparameters K-means influence clustering. K is chosen automatically by using the formula K = max\{2, argmax(silhouette scores)\}, which is the cross at the highest peak of the dotted line. MOEA/D hyperparameters numbered from 1 to 7  are $\lambda$, $P[\mathrm{X}]$, $\mathrm{X}_{\mathrm{DI}}$, $P[\mathrm{PM}]$, $\mathrm{PM}_{\mathrm{DI}}$, $Mode$, and $\epsilon_N$, respectively. A hyperparameter number appearing twice indicates the values of direct and interaction effect. A letter `i' as a prefix to a number in the plot indicates interaction/total effect, and the of 'i' indicates direct effect. Different colors represent different clusters.
			\label{fig:MOEAD_cluster_param}}
	\end{figure}

\begin{figure}
		\centering
		\includegraphics[width=\linewidth]{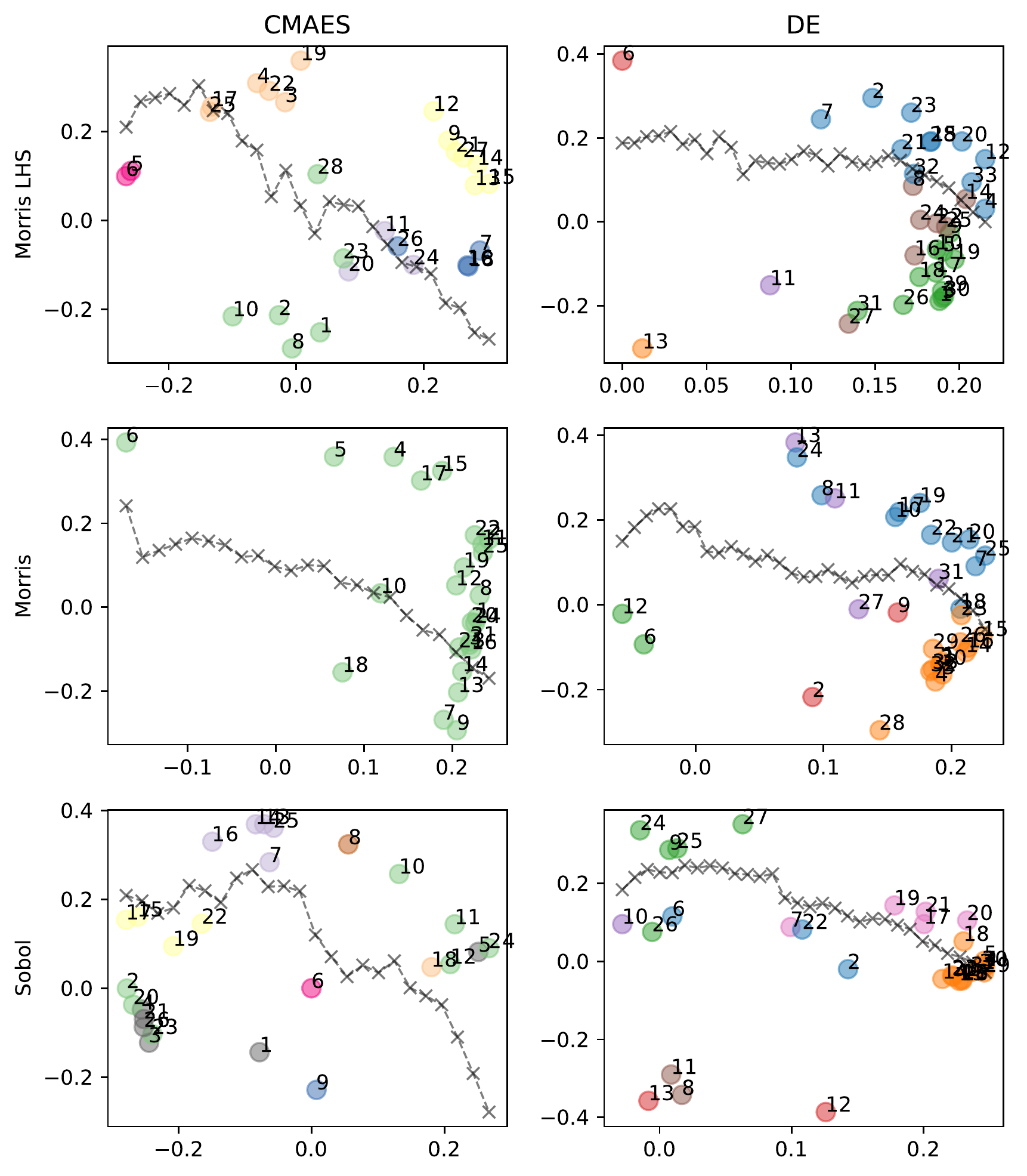}
		\caption{DE and CMA-ES function clustering based on hyperparameters influence on them. The numbers 1 to 23 are of the function in~\citep{Tesbench}, 24 to 27 are shifted functions~\citep{liang2013problem} Sphere, Ellipsoid, Ackley, and Griewank; and 28 to 33 are shifted and rotated functions~\citep{liang2013problem,liang2014problem} Rosenbrock, and Rastrigin, Weierstrass, Schwefel, Katsuura, and HappyCat. Different colors represent different clusters.
			\label{fig:DE_CMA_Cluster_func}}
	\end{figure}

\begin{figure}
		\centering
		\includegraphics[width=\linewidth]{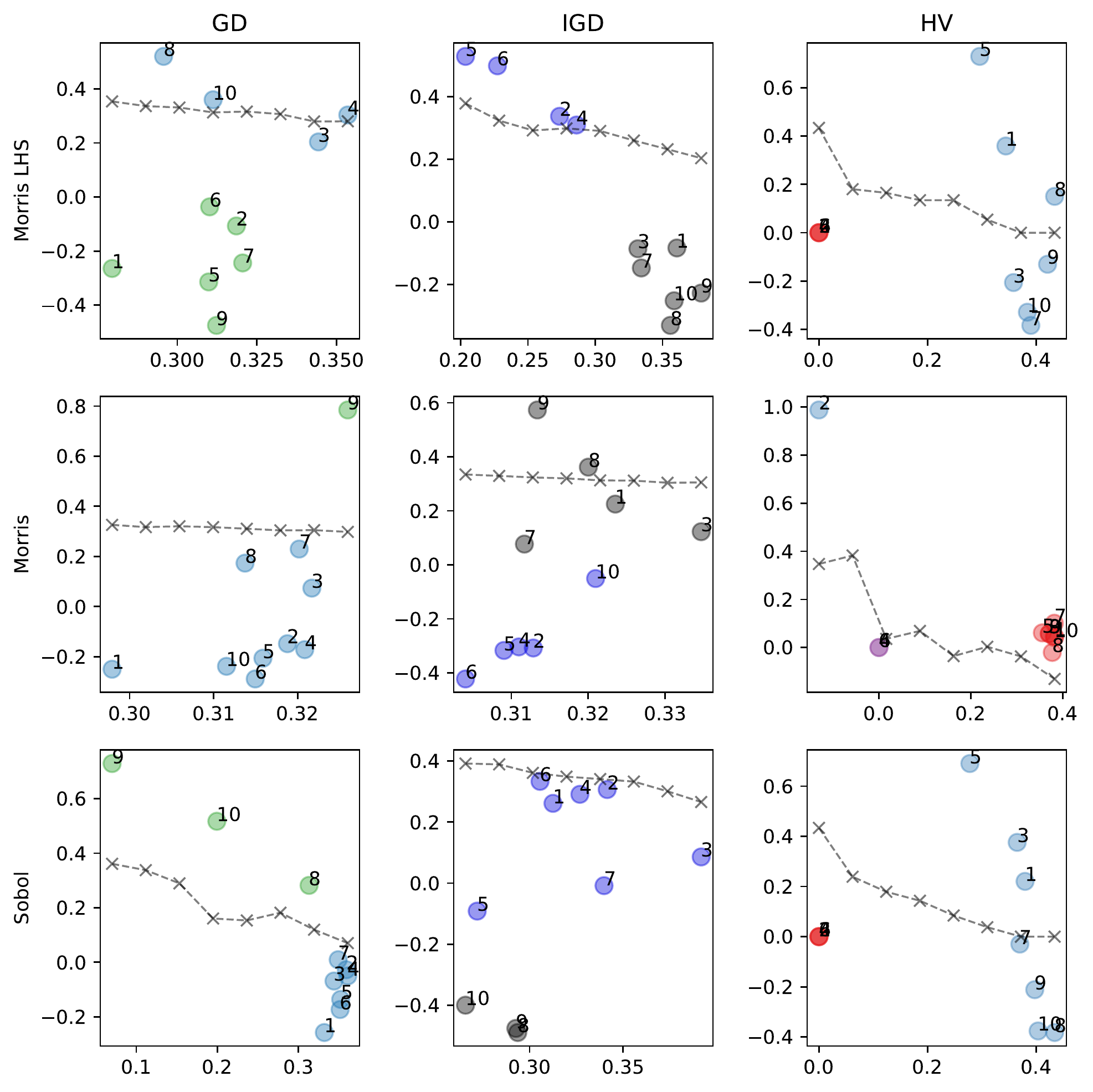}
		\caption{NSGA-III performance on multi-objective functions. Clustering based on the influence of the hyperparameters on the functions. The numbers 1 to 10 represent optimization problems CDTLZ2, DTLZ1, DTLZ2, DTLZ3, DTLZ4, IDTLZ1, IDTLZ2, WFG3, WFG6, and WFG7, respectively. Different colors represent different clusters.
			\label{fig:NSGAIII_cluster_func}}
	\end{figure}

\begin{figure}
		\centering
		\includegraphics[width=\linewidth]{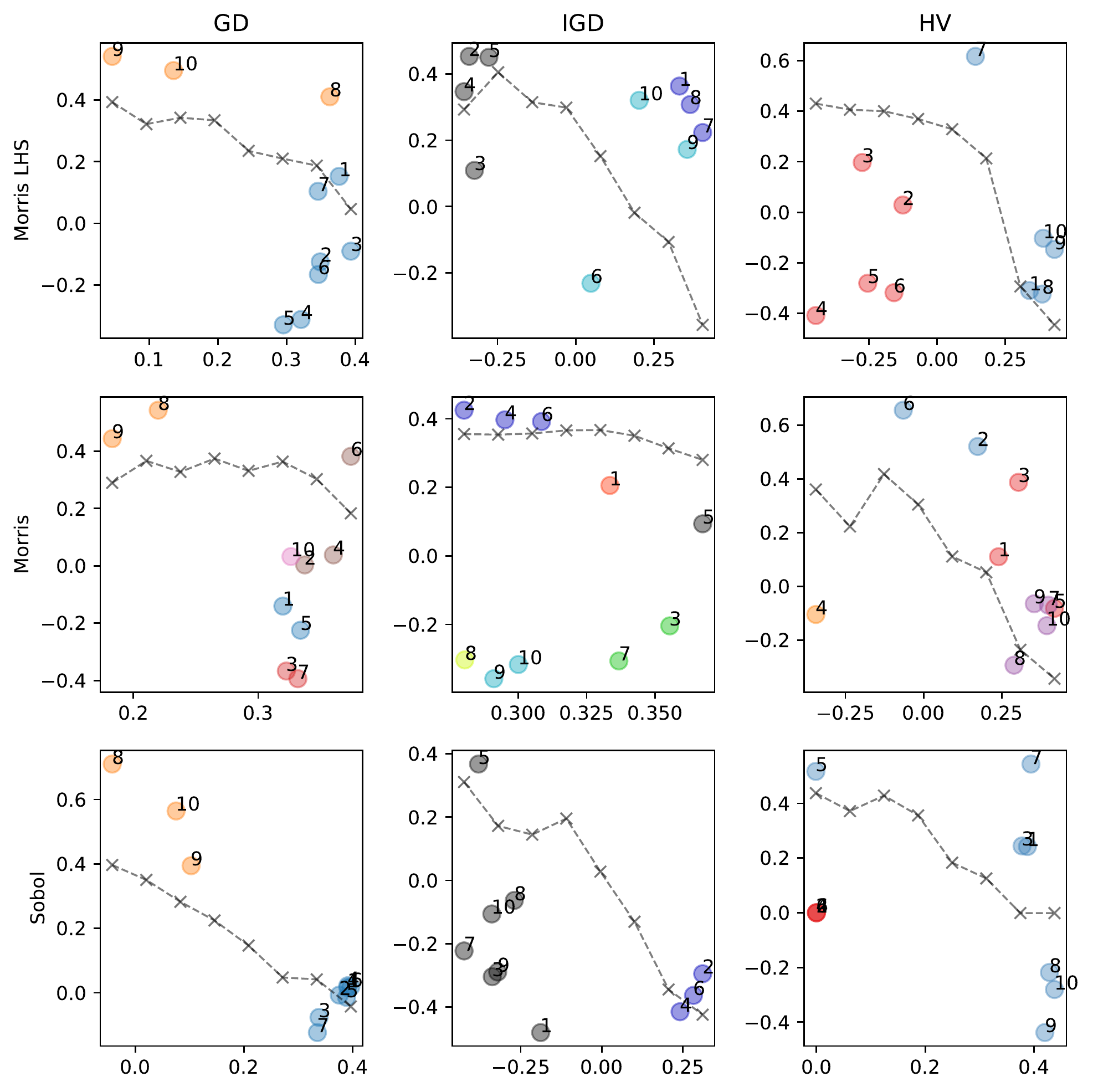}
		\caption{MOEA/D performance on multi-objective functions. Clustering based on the influence of the hyperparameters on the functions. The numbers are 1 to 10 represent optimization problems CDTLZ2, DTLZ1, DTLZ2, DTLZ3, DTLZ4, IDTLZ1, IDTLZ2, WFG3, WFG6, and WFG7, respectively. Different colors represent different clusters.
			\label{fig:MOEAD_cluster_func}}
	\end{figure}

	We hypothesized that the hyperparameters that have a similar influence on the performance of  algorithms tend to cluster together, and similarly, the functions that have similar characteristics may cluster together. Moreover, we may find the number of clusters and the trajectory of clustering based on the silhouette score also indicate differences in characteristics as a silhouette score defines how well the clusters are separated from each other and how good is the proximity of data points within a cluster. Figures~\ref{fig:DE_CMA_Cluster_param}, 
	\ref{fig:NSGAIII_cluster_param}, and 
	\ref{fig:MOEAD_cluster_param} are hyperparameters clustering. Whereas Figs.~\ref{fig:DE_CMA_Cluster_func}, 
	\ref{fig:NSGAIII_cluster_func}, and 
	\ref{fig:MOEAD_cluster_func} are functions clusters. The x and y axes are two principal components of the matrix formed by direct and interaction (total) influence values in these plots. 
	
	\textbf{Hyperparameters performance characteristics.} The hyperparameters clustering for single-objective algorithms in Fig.~\ref{fig:DE_CMA_Cluster_param} shows that hyperparameter $\mathbf{b}_{\mathrm{type}}$ of DE has clear, distinct influence characterization, whereas all other hyperparameters tend to cluster in a similar group. For CMA-ES, different sampling methods tend to cluster hyperparameters differently, where hyperparameter $\lambda$ shows more distinct behavior than others (see Fig.~\ref{fig:DE_CMA_Cluster_param}). We may also observe  the line of silhouette score that represents how the optimal clustering number varies as the number of clusters gradually tend to go to the maximum number of data points characterizing variation in the similarity and dissimilarity between hyperparameters due to different sampling methods and metrics.
	
	Multi-objective clustering results for MOEA/D in Fig.~\ref{fig:MOEAD_cluster_param} show that hyperparameter influence varies a lot (mostly on HV metric and less on IGD metric), and only a few clusters can be seen, such as $\lambda$ and the hyperparameters that are related also appear to have high proximity such as \{$P[\mathrm{X}]$ and $\mathrm{X}_{\mathrm{DI}}$\} and \{$P[\mathrm{PM}]$ and $\mathrm{PM}_{\mathrm{DI}}$\}. For NSGA-III, $\lambda$ on GD metric has a separate identity (see Fig.~\ref{fig:NSGAIII_cluster_param}). However, there are 2-3 groups of hyperparameters.
	
	\textbf{Functions performance characteristics.} When analyzing functions clustering characteristics, we may observe that (although difficult to read each individual number due to overlapping points) each hyperparameter's sampling method has different influences on the performance of the algorithms. Not only that, but different algorithms for different metrics also produce a variety of clustering of functions. However, one may be able to group functions into a few clusters, as evident from  Figs~\ref{fig:DE_CMA_Cluster_func}, 
	\ref{fig:MOEAD_cluster_func}, and 
	\ref{fig:NSGAIII_cluster_func}. 
	
\end{document}